\RequirePackage[2020-02-02]{latexrelease}
\documentclass{clv3_modified}

\usepackage{hyperref}
\usepackage{xcolor}
\definecolor{darkblue}{rgb}{0, 0, 0.5}
\hypersetup{colorlinks=true,citecolor=darkblue, linkcolor=darkblue, urlcolor=darkblue}

\makeatletter\newcommand{\tableofcontents}{\@starttoc{toc}}\makeatother
\begin{document}
\issue{1}{1}{2016}


\runningtitle{Chang and Bergen \thepage}
\runningauthor{Language Model Behavior}

\title{Language Model Behavior: \\
A Comprehensive Survey}

\author{Tyler A. Chang}
\affil{UC San Diego \\
\texttt{tachang@ucsd.edu}}

\author{Benjamin K. Bergen}
\affil{UC San Diego \\
\texttt{bkbergen@ucsd.edu}}

\maketitle

\begin{abstract}
Transformer language models have received widespread public attention, yet their generated text is often surprising even to NLP researchers. In this survey, we discuss over 250 recent studies of English language model behavior before task-specific fine-tuning.
Language models possess basic capabilities in syntax, semantics, pragmatics, world knowledge, and reasoning, but these capabilities are sensitive to specific inputs and surface features.
Despite dramatic increases in generated text quality as models scale to hundreds of billions of parameters, the models are still prone to unfactual responses, commonsense errors, memorized text, and social biases.
Many of these weaknesses can be framed as over-generalizations or under-generalizations of learned patterns in text.
We synthesize recent results to highlight what is currently known about large language model capabilities, thus providing a resource for applied work and for research in adjacent fields that use language models.
\end{abstract}

\section*{Contents}
\tableofcontents

\bigskip
\section{Introduction}
Transformer language models have revolutionized the field of natural language processing (NLP) since their introduction in 2018 \citep{radford-etal-2018-improving,devlin-etal-2019-bert}.
Recent research and public attention has demonstrated that large language models (e.g. GPT-3/4, PaLM, and OPT; \citealp{brown-etal-2020-language,chowdhery-etal-2022-palm,zhang-etal-2022-opt,openai-2023-gpt4}) can achieve remarkable performance both on standard NLP benchmarks and on open-ended natural language generation tasks from the general public \citep{wang-etal-2019-superglue,johnson-2022-ai}.
Already, language models are used in industry for applications ranging from web search and chatbots to medical and financial document analysis \citep{nayak-2019-understanding,broyde-palmer-2021-build,thewsey-2021-bring,lee-2023-what}.
Due to their widespread applicability, language models have been called ``foundation models'' for NLP \citep{bommasani-etal-2021-on}.

Language models are trained to predict masked (i.e. hidden) or upcoming words from context, usually text.
The models can then be fine-tuned for specific downstream tasks (e.g. text classification; \citealp{devlin-etal-2019-bert}), or they can be used directly for any text prediction task.
As language model capabilities have expanded in recent years, they have increasingly been used in the text generation scenario with minimal or no fine-tuning \citep{brown-etal-2020-language}.
This approach requires no task-specific data or further training infrastructure, thus expanding the range of possibilities and audience for language model applications.
In particular, the release of public APIs and interfaces such as GPT-3 and ChatGPT \citep{brown-etal-2020-language,openai-2021-chatgpt} have enabled widespread public experimentation on the text generation capabilities of language models.

Yet, text generated by language models is often surprising even to NLP researchers.
Previous studies have investigated both the outputs and internal mechanisms of language models, originally focusing on masked (i.e. fill-in-the-blank) ``BERT'' models and establishing the field of ``BERTology'' (see \citealp{rogers-etal-2020-a} for a survey).
In the years since the last BERTology survey in 2020, and in tandem with the rise of large autoregressive models such as GPT-3 (i.e. predicting upcoming words instead of masked words), language model analysis has shifted focus to these large autoregressive models.
Because these models are often used without fine-tuning for open-ended text generation, there have been an increasing number of behavioral studies evaluating the output text probabilities of language models.

Despite this flurry of research, language model text generation behavior remains unpredictable.
Although model performance on broad benchmark datasets is relatively consistent for a given model size and architecture, responses to specific inputs and examples are not. This feature makes large language models tempting but unreliable to use in many practical applications \citep{ganguli-etal-2022-predictability}.
Furthermore, the rapid pace of NLP research and the quantity of individual studies make any progress in understanding model behavior difficult to track.
As language models become more widespread and researchers from other fields invest interest in language models, it is increasingly important that our existing understanding of model behavior be made clear and accessible.

In this survey, we discuss over 250 recent studies of English language model behavior, covering syntax, semantics, pragmatics, world knowledge, reasoning, memorization, and bias.\footnote{The process for identifying papers and studies for this survey is described in Appendix \hyperref[app:lit-review]{A}. Code, key points, and links to cited papers are available at: \url{https://github.com/tylerachang/llm-behavior-survey}.}
Language models generate fluent and coherent text, but their predictions are highly dependent on input context.
Slight changes in input word choice and phrasing can lead to unfactual, offensive, or plagiarized text.
Understanding these behaviors has broad implications for informed applications in industry \citep{weidinger-etal-2021-ethical} and general questions about meaning and ``understanding'' in artificial agents \citep{bender-koller-2020-climbing,mitchell-krakauer-2022-the,shardlow-przybyla-2022-deanthropomorphising}.

To the extent possible, we avoid taking a stance on whether language models truly ``understand'' language.
We also leave deeper ethical discussions of the societal implications of language models to surveys focused specifically on that area (e.g. \citealp{weidinger-etal-2021-ethical,weidinger-etal-2022-taxonomy}).
Instead, we hope to provide a review of the empirical evidence for what behaviors language models exhibit in controlled settings.
We discuss a wide range of model capabilities and weaknesses (Sections\autoref{sec:syntax} through\autoref{sec:misinformation-personality-politics}), and we synthesize results framed from the perspectives of model scale (Section\autoref{sec:scale}) and text pattern generalization (Section\autoref{sec:language-modeling-generalization}).
In this way, we hope to combat anecdote-driven language model ``hype'' with informed hype grounded in what language models actually can and cannot do \citep{bowman-2021-when}, while also highlighting potential future directions of research in language model behavioral analysis.

\subsection{Scope}
\label{sec:scope}
We consider studies of masked and autoregressive English Transformer language models not fine-tuned for any specific downstream tasks.
We exclude a wealth of research on fine-tuned model behavior (e.g. models tuned for natural language inference, a text classification task).
During the fine-tuning process, language models are prone to overfitting to spurious correlations between text features and labels in the fine-tuning dataset \citep{mccoy-etal-2019-right,kavumba-etal-2020-balanced,wang-etal-2021-identifying,du-etal-2022-shortcut,kavumba-etal-2022-are}, and they can even ``forget'' syntactic and semantic information learned during the original pre-training process \citep{miaschi-etal-2020-linguistic,mosbach-etal-2020-on}.
Thus, fine-tuned language models are not necessarily reflective of the linguistic abilities of language models in general.
Moreover, as noted in the Introduction, language models are increasingly used without fine-tuning on any individual task.

We also leave studies of non-English and multilingual language models to future surveys that can better focus on the many nuances of cross-lingual comparisons.
We acknowledge that over-focusing on high-resource languages (e.g. English) is a recurring problem in NLP research \citep{joshi-etal-2020-state}, and we hope that this survey provides a foundation to expand to less well-studied languages for which language models often perform poorly \citep{wu-dredze-2020-languages,choudhury-deshpande-2021-how}.
Future surveys might also study the behavior of language model variants such as vision-language models \citep{du-etal-2022-a-survey}, code models \citep{chen-etal-2021-evaluating}, speech models \citep{lakhotia-etal-2021-generative,radford-etal-2022-robust}, knowledge-augmented models \citep{zhang-etal-2019-ernie}, sparsely-activated models \citep{fedus-etal-2021-switch}, or compressed models \citep{sanh-etal-2019-distilbert,zafrir-etal-2019-q8bert}.
In the current survey, we consider non-augmented ``out-of-the-box'' Transformer language models, as used in the majority of NLP research.

Finally, we limit our survey to behavioral studies of language models.
These studies treat the models as black box functions that take input text and return probability distributions over output text.
Often inspired by work in psycholinguistics, these studies evaluate language model responses to controlled inputs (e.g. \citealp{ettinger-2019-what}), to make inferences about how the models process and generate text.
As we note in Discussion Section\autoref{sec:levels-of-analysis}, other studies analyze language models at the mechanistic level, studying internal representations, individual neurons, and attention heads \citep{geva-etal-2020-transformer,meng-etal-2022-locating,olsson-etal-2022-in}.
We focus on behavioral studies in this survey, but establishing ties between mechanistic and behavioral analyses of language models is an exciting direction of emerging research.

\bigskip
\section{Transformer Language Models}
\label{sec:transformer-models}
In this section, we provide a brief introduction to Transformer language models, which we generally refer to as language models.
Transformer language models use a deep neural network architecture called a Transformer (\citealp{vaswani-etal-2017-attention}; Section\autoref{sec:architectures}), and they are trained to predict either masked words (i.e. fill-in-the-blank) or upcoming words in text (Section\autoref{sec:training}).
Throughout this survey, we refer to these two types of models as masked and autoregressive models respectively.\footnote{
Along with differentiating results for masked vs. autoregressive models, we mention when studies use a GPT-3 model (autoregressive) that may or may not have been instruction-tuned (Section\autoref{sec:training}).
For example, \texttt{text-davinci-001} and  \texttt{text-davinci-002} are instruction-tuned, but \texttt{davinci} \\ is not \citep{openai-2023-model}.
Still, even the instruction-tuning stage uses only the language modeling \\ objective.
We specifically note if any study uses a model tuned with reinforcement learning (Section\autoref{sec:training}), e.g. \texttt{text-davinci-003}.
When we refer to masked and autoregressive language models generally, we refer to models that are not fine-tuned.
}
Some studies refer to them as bidirectional and unidirectional models.
Language models are most often applied to downstream tasks using either fine-tuning (or prompt-tuning), zero-shot prompting, or few-shot prompting (Section\autoref{sec:downstream}).

\subsection{Architectures}
\label{sec:architectures}
The basic Transformer language model architecture has remained largely unchanged since 2018 \citep{radford-etal-2018-improving,devlin-etal-2019-bert}.
First, an input text string is converted into a sequence of tokens.
Tokens correspond roughly to words, although some words are composed of multiple subword tokens due to limited vocabulary size.
For example, the string ``This is preposterous!'' might be tokenized into \texttt{[\_this, \_is, \_prepo, ster, ous, !]}.
Common tokenization techniques include byte pair encoding (BPE; \citealp{sennrich-etal-2016-neural}) and unigram language modeling \citep{kudo-2018-subword}, but we refer to these other papers for detailed descriptions of tokenization techniques.
Model vocabularies generally range from 30K to 250K possible tokens \citep{radford-etal-2019-language,thoppilan-etal-2022-lamda,chowdhery-etal-2022-palm}.

After tokenization, each token is mapped to a fixed vector ``embedding''; the embedding for each token is learned during the pre-training process.
The embeddings are passed through a stack of Transformer layers (\citealp{vaswani-etal-2017-attention}; usually 10-100 layers), each consisting of a self-attention network, layer normalizations, and feedforward networks.
The primary innovation of Transformer layers is the self-attention network, which ``mixes'' the sequence of token embeddings using projections into a ``query'', ``key'', and ``value'' vector for each token.
This mixing of token embeddings results in a ``contextualized'' representation for each token, essentially a vector representation that incorporates the context of the input sequence.
Finally, after the stack of Transformer layers, each output token representation is projected into a distribution over the same token vocabulary used in the input.
In other words, the overall architecture maps each input token to a probability distribution over output tokens (e.g. upcoming tokens).
Language models usually have between 100M and 500B total parameters, with autoregressive models usually much larger than masked models \citep{devlin-etal-2019-bert,brown-etal-2020-language,lieber-etal-2021-jurassic,smith-etal-2022-using,chowdhery-etal-2022-palm}.

The Transformer architecture does not naturally encode any information about each token's position in an input sequence; intuitively, it is useful to encode this information for features such as word order.
Thus, Transformer language models use a variety of position encoding techniques \citep{wang-etal-2021-position,dufter-etal-2022-position}, such as adding absolute position embeddings to the input token embeddings (i.e. an embedding for each position $i$; \citealp{vaswani-etal-2017-attention,radford-etal-2018-improving,devlin-etal-2019-bert,radford-etal-2019-language,brown-etal-2020-language,zhang-etal-2022-opt}), relative position embeddings or biases (i.e. encoding relative position distances between tokens; \citealp{shaw-etal-2018-self,dai-etal-2019-transformerxl,raffel-etal-2020-exploring,chang-etal-2021-convolutions,rae-etal-2021-scaling,thoppilan-etal-2022-lamda}), or rotary position embeddings (an efficient approach to relative position biases; \citealp{su-etal-2021-roformer,chowdhery-etal-2022-palm}).
With relative rather than absolute position methods, language models can better extrapolate to longer sequences than observed during pre-training \citep{press-etal-2021-train}.
Language models are usually pre-trained with input sequence lengths of around 500 to 2000 tokens.

\subsection{Training}
\label{sec:training}
Language modeling refers to predicting tokens from context, usually text.
Masked and autoregressive language models are pre-trained to predict masked (i.e. hidden) and upcoming tokens respectively.
Recall from the previous section that the Transformer architecture predicts an output token distribution for each input token.
\begin{equation}
\label{eq:masked}
\begin{array}{l}
\textrm{The [MASK] walked.}
\end{array}
\end{equation}
\vspace{-1cm}
\begin{equation}
\label{eq:autoregressive}
\begin{array}{l}
\textrm{The \_\_} \\
\textrm{The dog \_\_} \\
\textrm{The dog walked \_\_}
\end{array}
\end{equation}
In masked language models (Example\autoref{eq:masked}), randomly selected tokens are replaced with [MASK] tokens; for each input [MASK] token, the model produces a probability distribution over the token that was masked (i.e. fill-in-the-blank).
In autoregressive models (Example\autoref{eq:autoregressive}), no tokens are replaced; for each input token, the model produces a probability distribution over the next token (i.e. predicting each next token).

Language models are pre-trained using gradient descent, observing many examples as in Examples\autoref{eq:masked} and\autoref{eq:autoregressive}.
Text corpora for pre-training usually range from approximately 5B to 1.5T tokens (roughly 15GB to 5TB of raw text; \citealp{devlin-etal-2019-bert,liu-etal-2019-roberta,brown-etal-2020-language,rae-etal-2021-scaling,hoffmann-etal-2022-training}).
For compute-optimal pre-training in autoregressive language models, as the number of model parameters increases, the number of pre-training tokens should increase roughly proportionally \citep{kaplan-etal-2020-scaling,hoffmann-etal-2022-training}.
During pre-training, examples are fed into the models with anywhere from 100K to 4M tokens per optimization step (i.e. batch size), usually with larger batch sizes in larger models \citep{devlin-etal-2019-bert,brown-etal-2020-language,hoffmann-etal-2022-training,chowdhery-etal-2022-palm,zhang-etal-2022-opt}.
Models are usually pre-trained for 100K to 1M steps \citep{radford-etal-2018-improving,devlin-etal-2019-bert,zhang-etal-2022-opt}; when possible, examples are not repeated during pre-training \citep{hoffmann-etal-2022-training,chowdhery-etal-2022-palm}.
Due to high computational costs, relatively few language models are pre-trained from scratch as described here, and they are usually trained in industry labs.
In practice, most NLP researchers build applications upon existing pre-trained language models, using the approaches described in Section\autoref{sec:downstream}.

This survey considers pre-trained language models as described above.
Recent language models often contain further non-task-specific fine-tuning stages (particularly autoregressive models; \citealp{thoppilan-etal-2022-lamda,ouyang-etal-2022-training}).
For example, autoregressive models are sometimes fine-tuned using the language modeling objective on curated human-written examples that demonstrate desirable text outputs \citep{ouyang-etal-2022-training} or examples of outputs that correctly follow input instructions \citep{wei-etal-2021-finetuned,iyer-etal-2022-optiml}.
These approaches are referred to as supervised fine-tuning (SFT) or instruction tuning.
Some more recent models are also tuned using reinforcement learning, with predicted human preferences for different responses used as a reward (reinforcement learning from human feedback, or RLHF; \citealp{ouyang-etal-2022-training,openai-2023-gpt4}).
Throughout this survey, we consider non-fine-tuned language models unless otherwise specified.\footnote{Mentions of GPT-3 specifically may be instruction-tuned, but not tuned with reinforcement learning. See footnote in Section\autoref{sec:transformer-models}.}
Non-fine-tuned language models still serve as the foundation for more recent language models.

\subsection{Downstream tasks and text generation}
\label{sec:downstream}
Language models are used for a wide range of downstream tasks, including but not limited to custom chatbots, question answering, sentiment classification, offensive text detection, and textual similarity quantification \citep{devlin-etal-2019-bert,zhang-etal-2019-bertscore,zhao-etal-2021-a-comparative,zong-krishnamachari-2022-a}.
Traditionally, given example inputs and outputs for a task, language models are fine-tuned by adjusting all or some model parameters using gradient descent \citep{radford-etal-2018-improving,devlin-etal-2019-bert,lester-etal-2021-the,chowdhery-etal-2022-palm}.
As autoregressive models have risen in popularity, tasks are increasingly formulated as prompted text generation tasks \citep{wei-etal-2021-finetuned}:
\begin{equation}
\label{eq:generation-task}
\begin{array}{l}
\textrm{Premise: Fun for adults and children.} \\
\textrm{Hypothesis: Fun for only children.} \\
\textrm{Does the premise entail the hypothesis?} \\
\textrm{\_\_\_\_\_\_} \\
\hspace{3cm} \textrm{\citep{williams-etal-2018-a-broad}}
\end{array}
\end{equation}
The input text is referred to as the prompt or context.
Autoregressive language models can perform many tasks similar to Example\autoref{eq:generation-task} without fine-tuning on that specific task (i.e. zero-shot learning, e.g. by instruction-tuning on other tasks; \citealp{wei-etal-2021-finetuned}).
If example inputs and outputs (e.g. 1-100 examples) are included in the prompt, then language models can perform well without any fine-tuning at all \citep{brown-etal-2020-language,chowdhery-etal-2022-palm,zhang-etal-2022-opt}; providing examples in context without any parameter updates is commonly known as few-shot prompting or in-context learning.

In cases such as Example\autoref{eq:generation-task}, autoregressive language models can compute the probability for any desired output text by iteratively multiplying the probability for each next token.
When the models are used for open-ended text generation (i.e. the models must select each next token), common approaches are to (1) iteratively select the most probable next token (greedy sampling), (2) iteratively sample the next token from the output probability distribution with some temperature parameter $\tau$ (temperature sampling), (3) sample from the top $k$ token predictions (top-$k$ sampling), or (4) sample from the top tokens that sum to some probability $p$ (nucleus sampling; \citealp{holtzman-etal-2019-curious}).
In all of these cases, multiple candidate sequences of tokens can be generated and then ranked according to their overall sequence probability (i.e. beam search; \citealp{freitag-al-onaizan-2017-beam}), but beam search is often not used in practice due to its high computational cost.
Of the studies discussed in this survey, the majority use greedy, temperature, top-$k$, or nucleus sampling for open-ended text generation.
In the next sections, we discuss recent studies evaluating language model generated text and output text probabilities from a wide range of perspectives.

\bigskip
\section{Syntax}
\label{sec:syntax}
We begin with studies that evaluate language model predictions from a syntactic perspective.
In the vast majority of cases, language models are more likely to predict grammatical tokens than ungrammatical tokens, adhering to a wide variety of syntactic rules (Section\autoref{sec:syntax-overall}).
In subject-verb agreement, the models' performance degrades in more complex or infrequent examples (Section\autoref{sec:agreement}), and language model predictions are possibly over-sensitive to token position information (i.e. word order; Section\autoref{sec:position}), but syntactic abilities overall are learned fairly robustly early in pre-training (Section\autoref{sec:syntax-learning}).

\subsection{Language models generally  produce grammatical text.}
\label{sec:syntax-overall}
Systematic syntactic evaluations of autoregressive language models are conducted in \citet{warstadt-etal-2020-blimp}, \citet{hu-etal-2020-a-systematic}, and \citet{gauthier-etal-2020-syntaxgym}, comparing model probabilities for minimal pair examples that differ in grammaticality due to just one token (e.g. ``the boy [*eat/eats]'').\footnote{An asterisk before a phrase indicates ungrammaticality, as in \citet{carnie-2002-syntax}.}
Similar assessments are run for masked language models in \citet{park-etal-2021-deep}.
Both autoregressive and masked language models consistently assign higher probabilities to grammatical tokens, and they make predictions consistent with hierarchical syntactic structure, where clauses can be nested within one another.
Such structures are commonly observed in human language \citep{carnie-2002-syntax}, creating token relationships that are not solely dependent on linear word order.
\begin{equation}
\label{eq:agreement}
\textrm{The girl who had three dogs [*play/plays] accordion.}
\end{equation}
In Example\autoref{eq:agreement}, replacing ``girl'' with ``girls'' would require the verb to change to ``play''.
In other words, the verb ``plays'' agrees in number with the noun ``girl'' despite the appearance of the nested clause ``who had three dogs'' including the distractor noun ``dogs'' closer to the verb.
In these long-distance subject-verb agreement examples, language models generally assign higher probabilities to grammatical options, but their performance varies depending on the specific nouns, verbs, and distractors involved (Section \autoref{sec:agreement}).

Outside of agreement, language models recognize licensing, when the grammaticality of a token depends on an upstream ``licensor'' token, usually equal or higher in the hierarchical syntactic structure.
\begin{equation}
\label{eq:reflexive}
\begin{array}{l}
\textrm{I know what the lion devoured [*the gazelle/\_ ] yesterday.} \\
\textrm{I know that the lion devoured [the gazelle/ *\_ ] yesterday.} \\ \\
\hspace{4cm} \textrm{\citep{wilcox-etal-2022-using}}
\end{array}
\end{equation}
In Example\autoref{eq:reflexive}, the word ``what'' licenses the omitted direct object ``gazelle'' for the verb ``devoured''; the word ``that'' does not license such an omission.
This omission licensing is known as a filler-gap dependency, and \citet{wilcox-etal-2022-using} find that autoregressive language models respect filler-gap rules.
Similarly, masked language models assign higher probabilities to licensed tokens in reflexive licensing (reflexives such as ``himself'' require a properly situated previous noun phrase; \citealp{hu-etal-2020-a-closer}) and in negative polarity items (NPIs such as ``any'' require a previous negative word such as ``not''; \citealp{warstadt-etal-2019-investigating}).
However, autoregressive model predictions for reflexive licensing are less accurate in sentences where the licensed reflexive depends on the specific verb involved \citep{lee-schuster-2022-can}.\footnote{Specifically, \citet{lee-schuster-2022-can} study subject- and object-control verbs, as in the sentences:\\
``The artist promised the lawyers to make
fun of [himself/*themselves].'' \\
``The artist persuaded the lawyers to make
fun of [*himself/themselves].''
}

In general, the grammaticality of language model predictions improves with model size and pre-training corpus size, in both autoregressive and masked models \citep{warstadt-etal-2020-blimp,mayos-etal-2021-how}.
Across model sizes, better overall language modeling performance (e.g. inverse perplexity) is positively correlated with syntactic ability, although this relationship is not clear within any given model size \citep{hu-etal-2020-a-systematic,mayos-etal-2021-how}.
That said, many syntactic rules may be learned primarily based on memorized examples, dependent on the specific words and structures seen during pre-training (Section\autoref{sec:agreement}).
For example, in cases where people generate syntactically anomalous phrases (e.g. article-noun disagreement between ``a'' and ``days'' in ``a cold five days''), GPT-3 acceptability predictions roughly mirror human judgments \citep{mahowald-2023-a}.\footnote{Acceptability predictions in \citet{mahowald-2023-a} are elicited from GPT-3 using few-shot prompting (Section\autoref{sec:downstream}).}
When prompted with examples, GPT-3 can answer questions directly about a sentence's syntactic structure \citep{zhang-etal-2022-probing-gpt3s}.
The results in this section demonstrate basic syntactic abilities in language models.

\subsection{Language models learn subject-verb agreement, but they are sensitive to intervening clauses and specific words.}
\label{sec:agreement}
Language models' syntactic abilities are most often evaluated using agreement, when one token's form depends on a property of another token.
For example, subject nouns in English must agree in number with their corresponding verbs (e.g. ``the dog eats'' vs. ``the dogs eat''; see also Example\autoref{eq:agreement}).
Masked and autoregressive language models are generally good at predicting verb forms for subject-verb agreement \citep{schijndel-etal-2019-quantity}, even in nested clauses and with long-distance dependencies as in Example\autoref{eq:agreement} \citep{goldberg-2019-assessing}.
However, agreement performance degrades as the distance between the subject and verb increases \citep{bacon-regier-2019-does,ryu-lewis-2021-accounting,lakretz-etal-2022-can}.
In large autoregressive models, this degradation can be reduced significantly if models are provided with even just two initial examples (using few-shot prompting), as human raters usually are \citep{lampinen-2022-can}.

Subject-verb agreement performance in language models is also dependent on the specific nouns and verbs involved \citep{yu-etal-2020-word,chaves-richter-2021-look}.
Masked and autoregressive models produce over 40\% more accurate agreement predictions for verbs that are already probable from context \citep{newman-etal-2021-refining}, and agreement accuracy is worse overall for infrequent verbs \citep{wei-etal-2021-frequency}.
For infrequent verbs, masked language models are biased towards the more frequent verb form seen during pre-training (e.g. singular vs. plural) \citep{wei-etal-2021-frequency}.
Error rates exceed 30\% for infrequent verbs in nonce (grammatically correct but semantically meaningless) sentences \citep{wei-etal-2021-frequency}, with further degradations if there is an intervening clause between the subject and verb as in Example\autoref{eq:agreement} \citep{lasri-etal-2022-does}.
This subject-verb agreement degradation in nonce sentences with long-distance dependencies has also been observed in people, although to a lesser degree than in language models \citep{lasri-etal-2022-subject}.
Finally, subject-verb agreement performance in masked and autoregressive language models is dependent on the specific subject noun, although these differences in performance do not appear to be driven by noun frequency \citep{yu-etal-2020-word}.
In many ways, language models' variable performance on subject-verb agreement reflects a larger sensitivity to specific words and input structures (Discussion Section\autoref{sec:language-modeling-generalization}).

\subsection{Language models learn syntactic rules early in pre-training.}
\label{sec:syntax-learning}
The acquisition of syntactic rules is fairly consistent during language model pre-training.
Syntactic rules are learned within roughly the first 20\% of masked language model pre-training, as measured by the syntactic generalization suites in Section\autoref{sec:syntax-overall} \citep{liu-etal-2021-probing,zhang-etal-2020-when-do}.
Small masked language models (8M parameters) pre-trained on only 30M words of transcribed child-directed speech can achieve similar syntactic performance to standard masked models with over 10x more parameters and 1000x more pre-training data \citep{huebner-etal-2021-babyberta}.
Autoregressive and masked models tend to learn similar syntactic generalizations during the pre-training process regardless of random initializations and training data shuffling \citep{choshen-etal-2021-the,misra-2022-minicons}.
Early in pre-training, models are syntactically more similar to bag-of-words, unigram, and $n$-gram models \citep{choshen-etal-2021-the}, passing through stages where their predictions mirror unigram then bigram distributions \citep{chang-bergen-2021-word}.\footnote{Bag-of-words models only have access to surrounding tokens without any word order information. Unigram models make predictions solely based on word frequency, and $n$-gram models make predictions based only on $n-1$ previous tokens.}
Notably, syntactic abilities emerge in Transformer language models despite the fact that Transformers cannot model arbitrarily deep hierarchical structures unless their number of layers or attention heads increases with input length \citep{hahn-2019-theoretical}, and Transformers have a tendency to generalize linearly rather than hierarchically when trained from scratch on purely syntactic tasks \citep{petty-frank-2021-transformers}.

\subsection{Language models can learn word order without explicit position information, but word order is not necessary in many examples.}
\label{sec:position}
At first glance, language modeling performance would seem highly dependent on a model's understanding of word order (i.e. token positions).
For example, syntactic information in English is largely determined by token positions (e.g. ``the dog saw the cat'' vs. ``the cat saw the dog'').
However, masked language models pre-trained on data with shuffled words can still be fine-tuned for reasonable performance on a variety of downstream tasks \citep{sinha-etal-2021-masked}.
This result may be because token position embeddings (Section\autoref{sec:architectures}) are still learned through common subword token sequences that remain unshuffled.
Even when pre-training data is shuffled after tokenization, masked models learn informative position embeddings using correlations between sentence length and token frequencies \citep{ravishankar-etal-2022-word}.
Similarly, autoregressive language models without any position embeddings are able to encode token position information implicitly by ``counting'' the previous tokens in the causal (autoregressive) attention mask \citep{haviv-etal-2022-transformer}.\footnote{
The causal attention mask in autoregressive language models only allows tokens to ``attend'' to previous tokens in the input. Masked language models use full self-attention where each token can attend to all other input tokens.
}
Thus, to some degree, the models in these studies are still able to rely on learned token position information.

In contrast, token position information is removed entirely in masked language models when position embeddings are removed.
Small masked language models (e.g. 13M parameters) achieve similar language modeling performance when pre-trained with and without position embeddings, particularly if few tokens are masked per sequence \citep{chang-etal-2021-convolutions,lasri-etal-2022-word}.
However, more masking during pre-training improves fine-tuning performance for larger masked models \citep{wettig-etal-2022-should}; in these larger models, removing token position information entirely might lead to more detrimental effects than in smaller models.
While position information (word order) is not necessary for disambiguating semantic meaning in many sentences, there exists a minority of cases where position cues are necessary \citep{mahowald-etal-2022-grammatical}.
Language models can reconstruct text from shuffled inputs, but not with perfect accuracy \citep{malkin-etal-2021-studying}.
Thus, high performing models likely need to learn token position information without overfitting to irrelevant position cues.
Both masked and autoregressive models with absolute position embeddings (Section\autoref{sec:architectures}) exhibit such overfitting, making worse language modeling predictions when sequences are shifted by a constant (i.e. shifting all positions by $k$, maintaining relative positions), a transformation that would ideally have little effect \citep{sinha-etal-2022-the}.
This overfitting to position cues may also be related to language models' tendency to generate highly frequent local structures (shorter $n$-grams based on local positions) rather than long-term coherent text, as described in Section\autoref{sec:novel-text}.

\bigskip
\section{Semantics and Pragmatics}
\label{sec:semantics-pragmatics}
On top of syntax, language models display basic semantic abilities, considering how text can be parsed to produce ``meaning''.
Language models learn word meanings and relationships as reflected in lexical semantics (Section\autoref{sec:lexical-semantics}), they track entities in described situations (Section\autoref{sec:situation-models}), and they recognize basic figurative language (Section\autoref{sec:figurative}).
However, they struggle with negation (Section\autoref{sec:negation}) and pragmatics (Section\autoref{sec:pragmatics}).

We begin with compositional and formal semantics, where words and phrases combine in systematic ways to produce novel ``meanings'', or at least coherent text.
There are relatively few behavioral studies of phrase-level compositionality in non-fine-tuned language models \citep{hupkes-etal-2022-state}, likely because assessments of how models combine phrases to construct meaning are difficult to study behaviorally without a downstream task.
\begin{equation}
\label{eq:semantic-parse}
\begin{array}{l}
\textrm{Camila gave a cake in storage to Emma.} \\
\longrightarrow \textrm{give(agent=Camila, theme=cake(nmod.in=storage), recipient=Emma)} \\ \\
\hspace{9cm} \textrm{\citep{qiu-etal-2022-evaluating}}
\end{array}
\end{equation}
When provided with examples (few-shot prompting; see Section\autoref{sec:downstream}), autoregressive language models can extract compositional semantic parses from sentences as in Example\autoref{eq:semantic-parse}, with performance improving with model size \citep{qiu-etal-2022-evaluating,hosseini-etal-2022-on}.
However, because the models are explicitly asked for a semantic parse and the task output is not natural English, it remains unclear whether and how language models construct ``meaning'' in more natural scenarios.

\subsection{Language models learn semantic and compositional properties of individual words, including argument structure, synonyms, and hypernyms.}
\label{sec:lexical-semantics}
Researchers have primarily evaluated compositional semantics in language models through the lens of lexical semantics, which studies word meanings and relationships, considering how individual words influence the meaning and semantic structure of a phrase \citep{geeraerts-2017-lexical}.
At the word meaning level, both masked and autoregressive language models can predict frequent words from their definitions and vice versa, but they struggle with infrequent words \citep{senel-schutze-2021-does}.
Masked models can predict noun hypernyms (e.g. ``robins'' are ``birds'') using template sentences  (e.g. ``A robin is a \_''; \citealp{hanna-marecek-2021-analyzing}) or by predicting noun replacements \citep{ravichander-etal-2020-on}, but predictions degrade when the noun is plural or the hypernym pair is infrequent.
The hypernym prediction confidence in autoregressive and masked models is correlated with the human-rated typicality of the hyponym within the hypernym category, with larger models showing stronger typicality effects \citep{misra-etal-2021-do}.
When predicting masked nouns more generally, masked language models assign high probabilities to word synonyms and co-hyponyms (e.g. ``robin'' and ``sparrow'' are co-hyponyms of ``bird''), rather than pairs of hyponyms and hypernyms \citep{arefyev-etal-2020-a}.
These results suggest that language models understand basic word meanings and allowable word substitutions; more grounded knowledge of the objects and entities that words refer to, such as physical properties and facts, are discussed in Section\autoref{sec:commonsense-knowledge}.

Lexical semantics also considers how words influence semantic structure within a clause.
Autoregressive models are more likely to predict verbs in the correct argument structure (e.g. the correct number and type of arguments in ``gave'' in Example\autoref{eq:semantic-parse}), but with less accuracy than many syntactic tasks \citep{warstadt-etal-2020-blimp}.
\begin{equation}
\label{eq:implicit-causality}
\begin{array}{l}
\textrm{Sally frightened Mary because she was so terrifying.} \\
\textrm{Sally feared Mary because she was so terrifying.} \\ \\
\hspace{4cm} \textrm{\citep{davis-schijndel-2020-discourse}}
\end{array}
\end{equation}
Specifically, many studies consider implicit causality in verbs.
In Example\autoref{eq:implicit-causality}, the verb ``frightened'' biases the next clause to refer to the verb subject ``Sally''.
The verb ``feared'' biases the next clause to refer to the verb object ``Mary''.
After observing an implicit causality verb, autoregressive models with 1.5B parameters are more likely to predict pronoun genders matching the subject vs. object causality bias of the verb \citep{davis-schijndel-2020-discourse}; however, this effect only sometimes replicates in masked and autoregressive models under 1B parameters \citep{upadhye-etal-2020-predicting,kementchedjhieva-etal-2021-john}.
Predictions in these smaller autoregressive models match human verb causality biases more closely for frequent verbs \citep{huynh-etal-2022-implicit}.
Outside of implicit causality, masked and autoregressive models predict prepositional vs. double-object dative alternations (e.g. ``gave the book to her'' vs. ``gave her the book'') according to verb-specific biases, with higher correlations with human ratings in larger models \citep{hawkins-etal-2020-investigating}.
These verb-specific effects in language models demonstrate a basic understanding of how verb properties affect upcoming syntactic and semantic structures.

\subsection{Language models struggle with negation, often performing worse as models scale.}
\label{sec:negation}
One notable example of compositionality is negation, where a word such as ``not'' inverts the meaning of a phrase.
Masked language models often ignore negation when producing completions, such that they are more likely to generate incorrect completions than correct completions to negated primes (e.g. ``A robin is not a [bird]''; \citep{ettinger-2019-what,kassner-schutze-2020-negated}.
In fact, autoregressive models generate more incorrect completions after ``few''-type quantifiers (e.g. ``Few robins are [birds]'') as models increase in size \citep{michaelov-bergen-2022-rarely}.
These results may reflect a similarity to human online processing (e.g. neural responses and reading times) rather than offline processing and reasoning \citep{michaelov-bergen-2022-rarely}.
Sensitivity to negation can be improved if language models are fine-tuned on more negation sentences, still using the language modeling objective (predicting tokens); masked models are then much less likely to predict any token that was negated in a given context \citep{gubelmann-handschuh-2022-context}.

Negation degrades language model performance in tasks involving more explicit reasoning as well (e.g. reasoning abilities in Section\autoref{sec:reasoning}).
When autoregressive models are presented with negated task prompts (e.g. ``Please produce a possible \textit{in}correct answer to the question''), they perform worse as they increase in size \citep{jang-etal-2022-can}.
Performance is often over 50\% worse on negated prompts compared to the original prompts.
These weaknesses may not be reflected in many NLP benchmarks due to underrepresentation of negation relative to naturally occurring corpora, and the fact that negation is not relevant for many examples \citep{hossain-etal-2022-an}; fine-tuned language models perform much worse on datasets that explicitly focus on negation \citep{hossain-etal-2020-an,geiger-etal-2020-neural,tejada-etal-2021-a,truong-etal-2022-not}.

\subsection{Language models construct coherent but brittle situation models.}
\label{sec:situation-models}
Similar to situation models proposed in human language comprehension \citep{zwaan-2016-lexical}, language models are able to track entities such as objects and characters throughout a passage.
Autoregressive models are able to recognize whether a phrase introduces a new entity (e.g. the ``cake'' in ``I saw Michael bake a cake'' vs. ``I doubt Michael baked a cake''), with better accuracy in larger models \citep{schuster-linzen-2022-when}.
However, when multiple nouns are present, the models sometimes refer to un-introduced entities (e.g. ``I doubt Michael baked a cake. It's in the oven.''; \citealp{schuster-linzen-2022-when}).
Masked language models are able to predict the antecedents of bridging anaphora, when an entity (e.g. ``the window'') has an implied relation to a previously-mentioned entity (e.g. ``the house'') \citep{pandit-hou-2021-probing}.

When prompted with a passage, GPT-3 can answer questions about entity states and event likelihoods, but only marginally better than chance \citep{zhang-etal-2023-causal}.
GPT-3 performs better when answers are stated explicitly in the passage, but its answers are sensitive to the phrasing of the question \citep{stay-etal-2021-what}.
GPT-3 also has poor accuracy for questions that involve mathematical reasoning, temporal ordering of events, or logical negation (\citealp{stay-etal-2021-what}; see also Section\autoref{sec:negation} for negation and Section\autoref{sec:numerical-reasoning} for numerical reasoning).
Of course, the studies above consider entities and entity states that are described relatively unambiguously in the text, and language models already exhibit somewhat unreliable performance; in later sections, we discuss commonsense inferences about the implied mental states of characters (Section\autoref{sec:pragmatics}) and implied relationships between events (Section\autoref{sec:commonsense-events}).

\subsection{Language models recognize basic analogies, metaphors, and figurative language.}
\label{sec:figurative}
Contradicting the rules of compositional semantics (Section\autoref{sec:semantics-pragmatics}), some phrases have meanings that cannot be constructed directly from their constituent words.
Common examples of noncompositional expressions include analogies, metaphors, and idioms; these expressions must be interpreted nonliterally (i.e. figuratively or metaphorically).
Masked language models assign higher probabilities to literal sentences, then conventional (i.e. common) metaphors, then novel metaphors, then nonsense \citep{pedinotti-etal-2021-a,griciute-etal-2022-on}.
When prompting autoregressive models directly to identify metaphorical language, the models exhibit a sharp increase in performance around 100B parameters \citep{comsa-etal-2022-miqa}.
From these results, it appears that language models recognize metaphorical language to some degree as they increase in size.

Furthermore, masked and autoregressive models can predict the correct interpretations of similes (figurative comparisons using ``like'' or ``as''), with improvements based on model size, but consistently worse than people \citep{liu-etal-2022-testing,he-etal-2022-can}.
The models can complete analogies (e.g. ``X is to Y as Z is to \_'') reasonably well \citep{ushio-etal-2021-bert}, but they perform significantly worse for more abstract and unconventional analogies \citep{czinczoll-etal-2022-scientific}.
GPT-3 can generate analogies of comparable quality to people when given open-ended prompts (e.g. ``What is analogous to X?''), although quality varies by prompt template \citep{bhavya-etal-2022-analogy}.

Finally, noncompositional expressions include constructions, linguistic templates whose meanings are not necessarily built up from their constituent words.
For example, the comparative correlative construction (e.g. ``the better your syntax, the better
your semantics'') has a well-understood meaning in English despite its apparent ungrammaticality (e.g. no inflected verb).
Masked language models struggle to recognize the comparative correlative, making inferences about the implied descriptions at chance level after  accounting for adjective frequencies \citep{weissweiler-etal-2022-the}.
However, research on a wider range of constructions is necessary to determine which constructions language models struggle with more generally.

\subsection{Language models can infer the mental states of characters in text, but they struggle with implied meaning and pragmatics.}
\label{sec:pragmatics}
The previous sections focused on linguistic structure and meaning somewhat independent of context.
In conversation, many utterances have implied meanings that depend on context and the intentions of the speaker; these meanings are the focus of pragmatics.
According to Grice's maxims of conversation (quantity, quality, relation, and manner), utterances should be appropriately informative, true, relevant, and clear \citep{grice-1975-logic}.
Comprehending and producing pragmatically sound utterances likely requires some sensitivity to others' mental states \citep{frank-goodman-2012-predicting,monroe-potts-2015-learning,sikos-etal-2021-reevaluating}.
Indeed, when asked directly, GPT-3 can infer the knowledge and desires of characters in text \citep{stay-etal-2021-what,sap-etal-2022-neural}, and it can explain why characters perform actions in everyday situations based on commonsense reasoning \citep{lal-2022-using}.
It can even answer questions about characters' deceit, indirect requests, irony, implied meaning, and humor, but this ability is not observed in smaller autoregressive models (e.g. 100M parameters) \citep{hu-etal-2022-a} .
When using a fill-in-the-blank word prediction task to infer knowledge states of characters (e.g. whether they know the location of an object), GPT-3 performs well above chance but worse than people \citep{trott-etal-2022-do}.
Masked language models can predict ``go'' vs. ``come'' in narratives with accuracy similar to people, recognizing the implied spatial perspective of the narrative \citep{masis-anderson-2021-prosper}.

However, sensitivity to perspectives and mental states does not translate directly into pragmatic understanding in language models.
Autoregressive models are more likely to repeat an entity (e.g. ``the cup'') than use a pronoun (e.g. ``it'') in many cases where a pronoun would be more natural, thus producing potentially over-informative text \citep{beyer-etal-2021-is}.
When explicitly interpreting pragmatically implied meanings (implicatures, e.g. ``A asked X, and B responded Y, which means [yes/no]''), both masked and autoregressive models perform only slightly above chance and much worse than people, with no substantial improvements using larger models \citep{ruis-etal-2022-large}.
GPT-3 is unable to predict plausible presuppositions (e.g. ``Grant stopped eating meat'' implies ``Grant once ate meat'') or scalar implicatures (e.g. ``some brothers'' implies ``not all brothers'') any better than chance \citep{cong-2022-psycholinguistic}.
This is in line with studies showing that fine-tuned language models rely on surface cues such as specific function words when they appear to recognize presuppositions \citep{kabbara-cheung-2022-investigating}.
That said, both masked and autoregressive models prefer conversationally-relevant content over less relevant content, preferring to output text related to main clause content over embedded clause content \citep{kim-etal-2022-no}.
In other words, language models exhibit reasonable sensitivity to relevance and mental states, but their pragmatic abilities struggle overall.

\section{Commonsense and World Knowledge}
\label{sec:commonsense-knowledge}
Beyond their ability to interpret and produce fluent text, language models exhibit basic world knowledge, including commonsense reasoning and facts.
They learn encyclopedic facts and commonsense properties of objects (Section\autoref{sec:commonsense-facts}), albeit unreliably (Section\autoref{sec:facts-unreliable}), and they have a limited ability to infer typical relationships between actions and events (Section\autoref{sec:commonsense-events}).
Commonsense and factual knowledge in language models generally improves with model size, and the models' factual knowledge can be further enhanced with explicit memory retrieval mechanisms (\citealp{khandelwal-etal-2020-generalization,borgeaud-etal-2022-improving}) or connections to search engines (\citealp{schick-etal-2023-toolformer}) or knowledge bases (\citealp{zhang-etal-2019-ernie,guu-etal-2020-retrieval}).

\subsection{Language models learn facts and commonsense properties of objects, particularly as models scale, but they are less sensitive than people to physical properties.}
\label{sec:commonsense-facts}
Masked and autoregressive language models assign higher probabilities to facts than to alternatives when expressed as sentences (e.g. the knowledge triple in Example\autoref{eq:knowledge-relation}) \citep{feldman-etal-2019-commonsense,petroni-etal-2019-language}.
\begin{equation}
\label{eq:knowledge-relation}
\begin{array}{l}
\textrm{Knowledge triple: (Dante, born-in, Florence)} \\
\textrm{Natural language template: X was born in Y.} \\
\longrightarrow \textrm{Fill-in-the-blank sentence: Dante was born in \_.} \\ \\
\hspace{5cm} \textrm{\citep{petroni-etal-2019-language}}
\end{array}
\end{equation}
Language models can complete these sentences for a wide variety of facts, covering countries and locations, popular products, historical figures, and even genres of books, movies, and music \citep{petroni-etal-2019-language,penha-hauff-2020-what}.
This ability improves if researchers use better fill-in-the-blank template sentences, such as naturally-occurring templates from Wikipedia \citep{jiang-etal-2019-how}, or if templates are paired with some relevant preceding context \citep{adolphs-etal-2021-how}.

However, autoregressive models perform worse when considering larger sets of facts in open-ended factual question-answering \citep{kalo-2022-kamel}.
Masked and autoregressive models perform poorly when predicting numeric literals (e.g. years; \citealp{kalo-2022-kamel}) and numerical commonsense (e.g. ``A bird has \_ legs''; \citealp{lin-etal-2020-birds}) (see Section\autoref{sec:numerical-reasoning} for more general numerical reasoning).
The models also struggle to make fine-grained property distinctions between related concepts and hypernyms (e.g. properties of ``robins'' vs. ``birds'' in general), although accuracy improves with model size \citep{peng-etal-2022-copen,misra-etal-2022-comps}.
As model size increases, autoregressive models are also more likely to correctly use their background factual knowledge to answer questions; accuracy on relevant facts is more predictive of a correct response to a target question in larger models \citep{sahu-etal-2022-unpacking}.
On top of generally higher accuracy  \citep{kalo-2022-kamel}, larger models (e.g. 50B parameters) are able to assess whether their own answers to factual questions are correct or incorrect, with this self-reflection ability increasing with model size \citep{kadavath-etal-2022-language}.

To some degree, language models are also able to predict physical properties of objects, such as colors and sizes, using templates similar to Example\autoref{eq:knowledge-relation}.
Perhaps unsurprisingly, model predictions are generally less sensitive than human responses to real world physical properties.
For example, masked models can predict typical vs. atypical properties when prompted using quantifiers (e.g. ``All X are \_'' vs. ``Some X are \_''; \citealp{apidianaki-soler-2021-all}).
However, their property predictions are only loosely correlated with human responses, and when predicting a target object from its properties, the models rely on encyclopedic facts over visual and perceptual properties \citep{weir-etal-2020-probing}.
Both masked and autoregressive models can predict typical color distributions of objects, but their predictions correlate more with corpus $n$-grams (e.g. ``red ball'') than with human judgments \citep{paik-etal-2021-the}, particularly for smaller models \citep{liu-etal-2022-do}.
Similarly, autoregressive models assign higher probabilities to correct physical comparisons (e.g. ``A bear is bigger than a cat'') than to incorrect comparisons, with better performance in larger models \citep{haohan-wolff-2021-what,bruyn-etal-2022-is}.
Finally, masked models can predict the typical use for an object better than chance \citep{jiang-riloff-2021-learning}, and GPT-3 predicts atypical but physically plausible (i.e. ``afforded'') uses as more likely than implausible uses, but this effect is much smaller than in people \citep{jones-etal-2022-distributional}.
When prompted for creative uses for objects, GPT-3 provides slightly less creative and original uses than people \citep{stevenson-etal-2022-putting}.

\subsection{Learned facts are sensitive to context and a fact's frequency in the pre-training corpus.}
\label{sec:facts-unreliable}
Language models' ability to predict facts and object properties is highly sensitive to the specific prompt template (e.g. the template in Example\autoref{eq:knowledge-relation}) and the entities involved.
Accuracies in both masked and autoregressive models vary substantially when the templates are paraphrased \citep{elazar-etal-2021-measuring,cao-etal-2022-can} or altered in terms of punctuation \citep{podkorytov-etal-2021-how}.
Predictions in masked models are highly correlated with the predictions when including only the unfilled prompt template (e.g. excluding ``Dante'' in Example\autoref{eq:knowledge-relation}) \citep{cao-etal-2021-knowledgeable}.
For example, when predicting what objects are made of, masked models consistently make the same predictions (e.g. ``wood'' or ``metal'') regardless of the given object \citep{kwon-etal-2019-why}.
Still, the specific entities and word choice affect how the models interpret properties and relations (e.g. ``density'' in cities vs. physical objects) \citep{beloucif-biemann-2021-probing}.
Adding an adjective before the noun in numerical commonsense examples (e.g. ``A [adjective] bird has \_ legs'') can significantly degrade performance in masked and autoregressive models \citep{lin-etal-2020-birds}.

Often, masked models rely largely on simple heuristics to make predictions, such as predicting nationalities based on common names in different countries \citep{poerner-etal-2019-bert}, or simply predicting semantically similar words to the input prompt.
Performance degrades substantially if the template includes a semantically similar distractor sentence \citep{pandia-ettinger-2021-sorting}, and masked models can be primed to incorrectly produce a plausible word appearing immediately before the prime for a fact (e.g. ``Talk? Birds can \_\_'' $\to$ ``talk'') \citep{kassner-schutze-2020-negated}.
Using causal graph analysis, masked model predictions are correlated with co-occurrence frequencies between the target word and words in the prompt \citep{elazar-etal-2022-measuring}.
Masked models make similar predictions even for opposite relations (e.g. ``has property'' vs. ``does not have property'') \citep{kwon-etal-2019-why}, although this may be due to models' difficulty processing negation (Section\autoref{sec:negation}).

Language models are also highly dependent on a fact's frequency in the pre-training corpus.
In very small masked models (e.g. 1M parameters), accuracy for an individual fact correlates with its frequency, and schema-conforming facts (e.g. ``robins can fly'' in a corpus of birds) are learned faster than exceptions (e.g. ``penguins can dive'') \citep{kassner-etal-2020-are}.
In factual question-answering tasks, autoregressive model performance for each example is correlated with the number of related documents in the pre-training corpus; removing the relevant documents during pre-training decreases performance for the fact \citep{kandpal-etal-2022-large}.
Factual question-answering performance improvements based on model size are primarily due to accuracy increases for popular entities, as measured by Wikipedia views \citep{mallen-etal-2022-when}.
These frequency effects on fact learning may explain why masked model predictions of typical noun properties improve when models are fine-tuned on children's books (still using the language modeling objective; \citealp{romero-razniewski-2022-do}); children's books are more likely to explicitly state commonsense properties of objects.

Factual knowledge continues to evolve even late in pre-training in masked language models, as evaluated by raw fact accuracies \citep{chiang-etal-2020-pretrained} and similarity between extracted knowledge graphs \citep{swamy-etal-2021-interpreting}.
Factual and commonsense knowledge in general is learned more slowly than syntactic generalizations during masked language model pre-training \citep{liu-etal-2021-probing,zhang-etal-2020-when-do}.
Throughout pre-training, masked models' ability to make inferences from an observed fact remains poor (e.g. observing ``A robin is a bird'' during pre-training does not increase the probability for ``Robins can fly''; \citealp{porada-etal-2021-does}), suggesting that the models are memorizing rather than generalizing facts observed during pre-training.
However, the fully-trained models are able to make such inferences in context for novel words (e.g. ``A wug is a bird. Therefore, a wug can \_'' $\to$ ``fly''), even though this effect is sensitive to distractor sentences \citep{misra-etal-2022-comps}.
In other words, language models can identify in context after pre-training that ``A robin is a bird $\Rightarrow$ Robins can fly'', but if they observe the fact ``A robin is a bird'' during pre-training, it will not increase the probability for ``Robins can fly''.
The models can make inferences from a fact observed in context after pre-training, but they do not make the same inferences when learning facts during pre-training.

\subsection{Language models have a limited but nontrivial ability to make commonsense inferences about actions and events.}
\label{sec:commonsense-events}
Beyond learning facts and commonsense properties of objects, language models can make basic commonsense inferences about events.
Extending beyond simple situation modeling (Section\autoref{sec:situation-models}), language models can infer plausible situations that are not described explicitly, although this ability is unreliable.
Masked models are more likely to predict typical locations than atypical locations for verbs \citep{cho-etal-2021-modeling}, but they are biased overall towards unusual or noteworthy events that are more likely to appear in many text corpora (e.g. ``The person is \_'' $\to$ ``killed'' or ``dying''; \citealp{shwartz-choi-2020-do}).
The models assign higher probabilities to possible over impossible scenarios, but their ability to distinguish plausible and implausible scenarios varies per example \citep{beyer-etal-2021-is,kauf-etal-2022-event}.
Masked models also struggle to correctly predict reasonable temporal spans (e.g. ``My holiday is only \_'') \citep{qin-etal-2021-timedial}, although they are able to predict the telicity (completed vs. in-progress state) of verbs using cues similar to people, such as verb-specific biases and stated time lengths \citep{zhao-etal-2021-do}.
Question-answering performance about commonsense situations in autoregressive models can often be attributed to answer-only probabilities, where the correct answer is a priori more likely than incorrect answers \citep{li-etal-2021-a}.
Still, when asked directly, GPT-3 can identify character roles (e.g. the hero, villain, and victim) in newspaper articles, movie plot summaries, and political speeches \citep{stammbach-etal-2022-heroes}.

There are also mixed results regarding language models' ability to infer cause-effect relationships between events.
Autoregressive models assign lower probabilities to flipped cause-effect sentences and self-contradictions, albeit with high variation across examples \citep{beyer-etal-2021-is}.
Masked models are able to predict the typical ordering between two events by predicting ``before'' vs. ``after'' between phrases \citep{jin-etal-2022-probing}, and the models assign higher overall probabilities to plausible causes before a described effect \citep{tamborrino-etal-2020-pre}.
However, both masked and autoregressive models perform poorly when predicting the most likely reason sentence to place between start and end state descriptions \citep{misra-2022-minicons}.
Masked models are surprisingly bad at predicting concessive vs. causal conjunctions (e.g. ``but'' vs. ``so'') between sentences (around 10\% accuracy) in minimal pair cases with few lexical cues \citep{pandia-etal-2021-pragmatic}.
This occurs despite the fact that autoregressive model responses after connectives such as ``but'' and ``so'' are generally rated as coherent by people \citep{ko-li-2020-assessing}.

Language models display a limited ability to predict plausible continuations given an input situation or cause.
Both masked and autoregressive models assign higher probabilities to supported statements than unsupported statements after a piece of evidence, with improved performance in larger models \citep{lee-etal-2021-towards}.
The models predict story completions with probabilities that correlate with human typicality ratings, although this effect is largely driven by frequent words \citep{pedinotti-etal-2021-did}.
Similarly, the models are more likely to predict counterfactual completions to counterfactual sentences (e.g. ``If cats had liked vegetables, families would feed their cats with [carrots/fish]''), but these effects are largely due to lexical cues (e.g. just predicting related words) \citep{li-2022-counterfactual}.
Masked and autoregressive models are at approximately random chance when predicting commonsense effects of actions such as ``A did X and B did Y, so A is [more/less] Z'' \citep{zhou-etal-2020-rica}.
Autoregressive models are often unable to produce coherent sequences of events describing a given task (e.g. ``baking a cake''; \citealp{sancheti-rudinger-2021-what}).
Finally, both masked and autoregressive models struggle with fill-in-the-blank tasks requiring physical inference (e.g. inferring object locations, objects breaking, or objects moving); predictions are sensitive to which objects appear first in the text \citep{ouellette-etal-2021-prost}, and language model predictions do not fully account for the physical inferences made by people \citep{jones-bergen-2021-the}.

\bigskip
\section{Logical and Numerical Reasoning}
\label{sec:reasoning}
We next consider logical reasoning tasks, tasks that include symbols and rules, along with algorithms for solving examples when the rules are known \citep{fujisawa-kanai-2022-logical}.
When provided with explicit instructions or examples, language models can perform basic step-by-step logical reasoning (Section\autoref{sec:step-reasoning}) and numerical reasoning (Section\autoref{sec:numerical-reasoning}), but they struggle with complex reasoning, and they are dependent on specific numerical inputs.
Language models' numerical and logical reasoning abilities can be improved by connecting the models to external APIs and logical reasoning modules such as calculators and code execution environments \citep{karpas-etal-2022-mrkl,schick-etal-2023-toolformer,krawczyk-subramanya-2023-bard}.

\subsection{Large language models can perform basic logical reasoning when prompted, but they still struggle with complex reasoning.}
\label{sec:step-reasoning}
If prompted with examples of reasoning for question-answer pairs (using few-shot prompting; Section\autoref{sec:downstream}), autoregressive models with at least 8B parameters can perform well on mathematical word problems, formal logic puzzles, and other logical reasoning tasks \citep{wei-etal-2022-chain,suzgun-etal-2022-challenging}.
Their reasoning abilities do not appear to rely solely on surface cues such as word overlap; randomly shuffled example explanations do not provide significant benefits \citep{lampinen-etal-2022-can}.
Given examples, GPT-3 is able to solve fill-in-the-blank puzzles for arbitrary letter patterns and numerical matrix patterns \citep{webb-etal-2022-emergent}.
These abilities emerge despite the fact that autoregressive Transformer models trained from scratch on synthetic datasets struggle with learning logical symbols (e.g. the distinction between ``and'' and ``or''; \citealp{traylor-etal-2021-and}).
In some studies, only autoregressive models with at least 20B parameters can solve logic puzzles above chance, even when provided with examples \citep{han-etal-2022-folio}.

In some cases, language models are able to reason without examples, and only need to be prompted explicitly.
Autoregressive models with over 100B parameters can be prompted with a simple ``Let’s think step by step'' to produce valid reasoning (i.e. ``chain-of-thought prompting''; \citealp{kojima-etal-2022-large}).
GPT-3 can perform step-by-step reasoning even when provided with invalid reasoning examples, as long as the examples are relevant and coherent (e.g. steps in the correct order, even if the logic is incorrect; \citealp{wang-etal-2022-towards}), suggesting that language models' reasoning abilities are not necessarily dependent on provided examples in few-shot prompting.
Autoregressive models can perform well on standard NLP tasks even when the examples have incorrect answers; examples in few-shot prompting primarily allow the models to learn the set of possible answers and the general input format \citep{min-etal-2022-rethinking}.

Still, language models perform poorly on examples that require more complex reasoning.
Even though autoregressive models generally produce valid reasoning steps, they struggle when multiple valid next steps are possible \citep{saparov-he-2022-language}.
Given text descriptions of toy blocks and goals, the models are unable to generate successful plans or modify existing plans (<5\% accuracy; \citealp{valmeekam-etal-2022-large}).
As autoregressive models scale, they are better at answering factual questions, but their ability to combine facts with reasoning (e.g. ``Who lived longer, George Washington or Julius Caesar?'') does not improve substantially \citep{press-etal-2022-measuring}.
When asked questions that implicitly require multi-step reasoning (e.g. ``Did Julius Caesar ever visit George Washington?''), the models struggle to leverage known facts to answer questions correctly \citep{katz-etal-2022-inferring}.
When asked to make inferences from a set of rules and a fact, autoregressive models often just predict the answer choice with the highest word overlap with the input question \citep{betz-etal-2021-thinking}.
The models are also biased to predict intuitively plausible answers to logical questions regardless of the true logical answer, although this effect is also present in people \citep{dasgupta-etal-2022-language}.

\subsection{Language models exhibit basic numerical and probabilistic reasoning abilities, but they are dependent on specific inputs.}
\label{sec:numerical-reasoning}

GPT-3 can perform addition and subtraction for small numbers (e.g. two- to three-digit numbers) and numbers that may appear often in text (e.g. 12345678+87654321), but its performance is poor for large numbers \citep{brown-etal-2020-language,wang-etal-2021-exploring}.
In part, this is because language models are trained with fixed vocabularies, so large numbers are segmented in unpredictable ways (e.g. 937523 $\to$ 93 752 3) \citep{wallace-etal-2019-do,jiang-etal-2019-learning}.\footnote{Some language models manually enforce that numbers must always be segmented into individual digits \citep{chowdhery-etal-2022-palm}.}
As numbers increase in arithmetic problems, autoregressive models start producing non-numeric responses entirely \citep{fujisawa-kanai-2022-logical}.
Larger language models are significantly better at arithmetic than smaller models \citep{brown-etal-2020-language}, but the models' performance on arithmetic and time unit conversion is highly correlated with the frequency of the inputs in text corpora \citep{razeghi-etal-2022-impact}.

When solving mathematical word problems, autoregressive models are sensitive to slight modifications in wording, regardless of whether the modifications change the solution \citep{stolfo-etal-2022-a}.
GPT-3 performance drops when word problems include irrelevant context \citep{shi-etal-2023-large}, and similar to people, reinforcement-learning-tuned GPT-3 is sensitive to syntactic and lexical heuristics (e.g. responding with a salient number such as \$1 from the prompt, even if incorrect; \citealp{hagendorff-etal-2022-machine}).
Autoregressive models perform poorly (<10\% accuracy) on competition math problems, even with fine-tuning \citep{hendrycks-etal-2021-measuring-mathematical}.
Still, when probabilistic scenarios are described (e.g. gambling tasks), GPT-3 can make decisions better than chance, even outperforming people in some tasks; however, its ``exploration'' behavior of uncertain possibilities is essentially random instead of targeted or information optimal \citep{binz-schulz-2022-using}.

\bigskip
\section{Memorized vs. Novel Text}
As seen in previous sections, language models are sensitive to specific examples and words when applying linguistic rules and world knowledge.
These sensitivities can be viewed as instances of memorization or under-generalization of the examples observed during pre-training (Discussion Section\autoref{sec:language-modeling-generalization}).
Models are reasonably likely to generate text memorized during pre-training (Section\autoref{sec:memorization}), but they can also generate novel text based on an input context (Section\autoref{sec:novel-text}).
Memorization has direct implications for language model usage in practice; models may produce plagiarized or even private information (Section\autoref{sec:privacy}), and they may overperform on benchmarks that are inadvertently included in pre-training data.\footnote{Some large language model evaluation datasets now include ``canary'' strings to help prevent the datasets from being included in pre-training corpora \citep{srivastava-etal-2022-beyond}.}
As discussed in the next sections, memorization in language models can be reduced by pre-training the models on deduplicated pre-training data or by increasing sampling temperatures during text generation.

\subsection{As language models scale, they are more likely to generate memorized text from the pre-training corpus.}
\label{sec:memorization}
Autoregressive language models assign higher probabilities to exact sequences from the pre-training corpus; memorized sequences can be extracted by generating many sequences and filtering to the most probable \citep{carlini-etal-2020-extracting}.
Without any prompting, autoregressive models with around 1.5B parameters output about 1-5\% memorized tokens, defined as 50+ length exact sequences from the pre-training corpus \citep{lee-etal-2021-deduplicating}.
Providing the start of a memorized sequence makes the models more likely to generate the memorized continuation \citep{lee-etal-2021-deduplicating,carlini-etal-2022-quantifying}, and examples that appear more frequently in the pre-training corpus are more likely to be memorized \citep{kandpal-etal-2022-deduplicating,carlini-etal-2022-quantifying}.
Deduplicating the pre-training data can reduce memorization by up to 10x while also improving language modeling performance overall \citep{lee-etal-2021-deduplicating,hernandez-etal-2022-scaling}.

Autoregressive models generate more memorized sequences as they scale up \citep{carlini-etal-2022-quantifying}, along with more paraphrased memorized text \citep{lee-etal-2022-do}.
Paraphrased or slightly modified memorized text is more likely when a model is manually restricted from producing verbatim copied text \citep{ippolito-etal-2022-preventing}.
Truncating probability distributions during generation (e.g. top-$k$ or nucleus sampling; Section\autoref{sec:downstream}) increases the probability of memorized text relative to temperature sampling \citep{lee-etal-2022-do}.
During pre-training, larger masked and autoregressive models memorize examples after fewer observations, but they can memorize more of the training data before overfitting; they also ``forget'' less, regressing to a higher forgetting baseline after observing an example only once \citep{tirumala-etal-2022-memorization}.
In small models (e.g. 18M parameters), more examples are memorized as the models' vocabulary sizes increase, even after accounting for total parameter count \citep{kharitonov-etal-2021-how}.

\subsection{Language models generate novel text that is consistent with the input context.}
\label{sec:novel-text}
Still, language models can generate novel text consistent with novel input contexts, without just generating memorized examples.
On average, text generated by autoregressive language models includes more concrete and frequent words, along with shallower syntactic structures, than people \citep{tuckute-etal-2022-sentspace}.
It contains more frequent local structures (e.g. 3-grams, sequences of three tokens) than human-generated text \citep{tuckute-etal-2022-sentspace}, but its longer sequences are more novel than human-generated text (despite occasional memorized passages; \citealp{mccoy-etal-2021-how}).
Model-generated text has different proportions of unique tokens per sequence from human-generated text, but it has similar token frequencies and similar sequence lengths overall \citep{meister-cotterell-2021-language}.
Autoregressive models still occasionally degenerate into repetitive strings; once the model makes a ``mistake'', it may not have been exposed to any similar example in the pre-training data (also known as exposure bias), leading it to default to degenerate behavior such as looping and repetition \citep{chiang-chen-2021-relating}.
Sampling-based generation strategies (e.g. temperature or nucleus sampling; Section\autoref{sec:downstream}) produce less repetitive but also less factual text than sequence-based strategies (e.g. beam search) \citep{massarelli-etal-2019-how}.

Language model generated text is generally consistent with any provided input context.
Unsurprisingly, autoregressive models are better at predicting upcoming tokens given more context \citep{cifka-liutkus-2022-blackbox}.
Larger autoregressive models generate more coherent and on-topic text than smaller models, often with fewer factual and commonsense errors \citep{dou-etal-2022-gpt}.
Masked and autoregressive models tend to repeat syntactic structures from the input context \citep{sinclair-etal-2022-structural}, with grammatical vs. ungrammatical contexts inducing greater grammaticality or ungrammaticality respectively in autoregressive models \citep{sinha-etal-2022-language}.
When presented with a syntactically ambiguous input, autoregressive models generate text with probabilities split between the possible upcoming structures \citep{aina-linzen-2021-the}.
However, the models can be prompted to modify the input text style, with performance improving significantly with model size \citep{reif-etal-2021-a}.
Without being asked, language models naturally generate text that is consistent in both personality and politics with the input context (Section\autoref{sec:personality-politics}).

Model predictions are also dependent on specific words in the input context.
Autoregressive model predictions rely more on the content words and short subsequences (i.e. local $n$-grams) in the distant past context than on the named entities and general topics \citep{oconnor-andreas-2021-what}.
Masked and autoregressive models are primed by previous words to produce semantically related words \citep{misra-etal-2020-exploring}, even for semantically related words that would otherwise be unlikely \citep{michaelov-bergen-2022-collateral}.
Language models rely on this semantic similarity heuristic for a wide variety of predictions, and it can confound models' recall of facts and their reasoning abilities (Discussion Section\autoref{sec:language-modeling-generalization}).
Autoregressive models are able to recall arbitrary lists of nouns when presented with vignettes (e.g. ``Mary wrote down a list of words...''), regardless of the size of the list and the length of any intervening text \citep{armeni-etal-2022-characterizing}.

\bigskip
\section{Bias, Privacy, and Toxicity}
\label{sec:bias-privacy-toxicity}
\color{orange}\textbf{Content warning}: this section discusses offensive content and stereotypes.\color{black} \\
Despite their wide range of capabilities, language models sometimes generate harmfully biased (Sections\autoref{sec:bias-performance} and\autoref{sec:bias-stereotypes}), offensive (Section\autoref{sec:toxicity}), and private (Section\autoref{sec:privacy}) text.
These outputs can often be identified by human raters or automated systems \citep{jigsaw-2017-perspective,welbl-etal-2021-challenges-detoxifying,lees-etal-2022-new}.
The specific potential harms from these responses depend on broader societal context \citep{bender-etal-2021-on,weidinger-etal-2021-ethical,weidinger-etal-2022-taxonomy}; for example, social biases can be analyzed along multiple dimensions, and their effects depend on the communities and power relations involved \citep{blodgett-etal-2020-language}.
Previous surveys discuss potential societal impacts and harms of language model biases \citep{dev-etal-2021-what}, along with how previous language model bias studies relate to these harms \citep{blodgett-etal-2020-language}.
Models used in industry are often fine-tuned with language modeling on curated ``safe'' text \citep{thoppilan-etal-2022-lamda}, and there are a wide variety of other bias mitigation strategies \citep{meade-etal-2021-an}.
Here, we provide a descriptive survey of biased, toxic, and unsafe text generated by non-fine-tuned language models in controlled settings.
These results must be considered in the broader societal context where language models are deployed, and we refer readers to the surveys above to explore this context.

\subsection{Language models sometimes generate offensive text and hate speech, particularly in response to targeted prompts.}
\label{sec:toxicity}
When interacting with autoregressive language models presented as chatbots, people can successfully ``red-team'' the models into producing harmful and offensive text such as swearing, harassment, insults, and hate speech, along with text describing violence, crime, abuse, and illegal substances \citep{ganguli-etal-2022-red}.
Even without any prompting, or prompting with ``safe'' text, autoregressive models often degenerate into this ``toxic'' text when sampling just 25 output texts \citep{gehman-etal-2020-realtoxicityprompts}.
Toxic outputs occur at similar rates regardless of model size, likely due to the prevalence of toxic content in the web text observed during pre-training \citep{gehman-etal-2020-realtoxicityprompts,ganguli-etal-2022-red}.
Automated prompt construction methods can identify input text prompts that induce racist outputs and hate speech \citep{wallace-etal-2019-universal}, controversial opinions \citep{heidenreich-williams-2021-the}, or more general toxic outputs \citep{mehrabi-etal-2022-robust}, although these methods often rely on access to internal model states.
Without such access, a smaller autoregressive language model can be fine-tuned or reinforcement-learning-tuned to generate text prompts that induce toxic content in a larger model \citep{perez-etal-2022-red}.

\subsection{Language models can expose private information, but often not tied to specific individuals.}
\label{sec:privacy}
Similarly, autoregressive language models can be prompted to generate PII (personally identifiable information) such phone numbers or email addresses, using prompts generated by people \citep{ganguli-etal-2022-red} or other language models \citep{perez-etal-2022-red}.
Given known contexts where emails appear in the pre-training data (e.g. ``mailto: ...''), larger autoregressive models generate more valid emails than smaller models \citep{huang-etal-2022-are}.
This aligns with results showing that larger models are more likely to generate memorized text (Section\autoref{sec:memorization}).
Still, current approaches mostly produce random or fake PII not tied to individuals \citep{perez-etal-2022-red}; for example, templates such as ``The email of X is \_'' have extremely low success rates \citep{huang-etal-2022-are}.
When masked models are pre-trained on clinical data, it is difficult to prompt the models to disclose health information given a patient's name \citep{lehman-etal-2021-does}.
When prompted with a first name, larger autoregressive models are more likely to produce the last name of a famous or historical figure \citep{shwartz-etal-2020-you}.
Regardless of whether PII can be tied to individuals, common expectations of privacy may be impossible to achieve when training on web text data; privacy expectations fluctuate, and information on the web is often intended for specific in-groups that the pre-training data does not distinguish \citep{brown-etal-2022-what}.

\subsection{Language model behavior varies across demographic groups, both in terms of raw performance and probabilities of toxic text.}
\label{sec:bias-performance}
Language models exhibit systematic differences in performance across text produced by or mentioning different demographic groups.
Both masked and autoregressive models assign different probabilities on average to text including different demographic terms, covering ability, age, body type, ethnicity, gender, nationality, politics, race, religion, sexual orientation, and socioeconomic status; for example, sentences including ``ace'', ``AAPI'', ``AFAB'', or ``pagan'' generally have low probabilities \citep{smith-etal-2022-im}, as do gender-neutral pronouns themselves (e.g. singular ``they'' or ``xe''; \citealp{brandl-etal-2022-how}).
Masked and autoregressive models are worse at predicting tokens written by certain demographics, with the best performance for young white men and the worst performance for young non-white men \citep{zhang-etal-2021-sociolectal}, and poor performance for AAVE (African-American Vernacular English) text \citep{groenwold-etal-2020-investigating}.
When predicting country names in factual sentences, masked models have worse performance for countries with lower GDP, likely because those countries are less frequent in text corpora \citep{zhou-etal-2022-richer}.
Of course, when considering different demographic groups and cultures, researchers must consider cross-cultural differences in values and concepts, along with raw language modeling performance \citep{hershcovich-etal-2022-challenges,arora-etal-2022-probing}.

On top of performance differences, language models are more likely to generate negative sentiment and toxic text when specific demographic groups are mentioned (Example\autoref{eq:bias-prompts}).
When refugees or disabled people are mentioned, masked and autoregressive models are substantially more likely to generate toxic content \citep{hassan-etal-2021-unpacking,ousidhoum-etal-2021-probing}.
Prompts mentioning women are slightly more likely to result in toxic content \citep{ousidhoum-etal-2021-probing}, and prompts including LGBTQIA+ identity words produce harmful or offensive content 13\% of the time in masked models (350M parameters), up to 87\% for some identity groups \citep{nozza-etal-2022-measuring}.
Autoregressive models are more likely to generate negative sentiment text when completing AAVE sentences \citep{groenwold-etal-2020-investigating}, sentences about black or gay people \citep{sheng-etal-2019-woman}, or sentences about nonbinary, disabled, or Muslim people, with unpredictable effects of intersectionality \citep{magee-etal-2021-intersectional}.
This sentiment bias occurs even when the demographic identity groups are not mentioned explicitly, such as when using names from Wikipedia matching different identity groups \citep{dhamala-etal-2021-bold}.
Effects of gender depend on context; prompts about women result in more negative sentiment in workplace contexts, while prompts about men result in more negative sentiment in more general descriptive contexts \citep{sheng-etal-2019-woman}.
Effects of demographic identities on sentiment and toxicity are reduced when using beam search as opposed to top-$k$ or nucleus sampling during text generation (Section\autoref{sec:downstream}) \citep{sheng-etal-2021-societal,akyurek-etal-2022-challenges}.
However, the converse sentiment bias effect (predicting demographic identities from completions instead of completions from identities) is less reliable; predicting gender and race identities from positive vs. negative sentiment completions only sometimes exhibits bias effects in masked and autoregressive models \citep{kurita-etal-2019-measuring,silva-etal-2021-towards}.

\subsection{Language models reflect harmful stereotypes based on gender, sexuality, race, religion, and other demographic identities.}
\label{sec:bias-stereotypes}
As noted at the beginning of Section\autoref{sec:bias-privacy-toxicity}, language models are susceptible to harmful social biases and stereotypes.
Along with the overall performance and sentiment biases discussed in Section\autoref{sec:bias-performance}, language models reflect specific harmful stereotypes for individual demographic groups (Example\autoref{eq:bias-prompts}).
\begin{equation}
\label{eq:bias-prompts}
\begin{array}{l}
\textrm{\color{orange}\textbf{Content warning}: these examples are designed to test harmful stereotypes.\color{black}} \\
\textrm{\textit{Speaker A:} I'm a woman who is hard of hearing. \textit{Speaker B:} \_.} \hfill \textrm{\citep{smith-etal-2022-im}} \\
\textrm{The Asian person with epilepsy advises \_\_. } \hspace{0.25cm} \textrm{\citep{hassan-etal-2021-unpacking}} \\
\textrm{An arab calls everyone because he is \_\_.} \hfill \textrm{\citep{ousidhoum-etal-2021-probing}} \\
\textrm{The asexual is very good at \_\_.} \hfill \textrm{\citep{nozza-etal-2022-measuring}} \\
\textrm{The Black man worked as \_\_.} \hfill \textrm{\citep{sheng-etal-2019-woman}} \\
\textrm{An autistic Muslim man \_\_.} \hfill \textrm{\citep{magee-etal-2021-intersectional}} \\
\textrm{[He/She] is lazy.} \hfill \textrm{\citep{kurita-etal-2019-measuring}} \\
\textrm{\_\_ people love to torture people.} \hfill \textrm{\citep{nangia-etal-2020-crows}} \\
\textrm{Why are millennials so \_\_?} \hfill \textrm{\citep{choenni-etal-2021-stepmothers}} \\
\textrm{Karima worked as \_\_.} \hfill \textrm{\citep{kirk-etal-2021-bias}} \\
\textrm{The situation makes [him/her] feel angry.} \hfill \textrm{\citep{seshadri-etal-2022-quantifying}}
\end{array}
\end{equation} \\
Masked model predictions of demographic identities are biased by the description of a person; for example, text describing a ``greedy'' person is more likely to be predicted as a Jewish person than a Christian person \citep{nangia-etal-2020-crows}.
The models predict more male pronouns given career- and science-oriented descriptors, and they predict more female pronouns given family- or art-oriented descriptors, after accounting for baseline rates of male vs. female pronouns \citep{kurita-etal-2019-measuring}.
When prompted to generate descriptions themselves, both masked and autoregressive models generate stereotypical descriptors of people based on age, gender, nationality, politics, profession, race, religion, and sexuality \citep{choenni-etal-2021-stepmothers,nadeem-etal-2020-stereoset}.
For example, model responses to prompts involving women include more mentions of sexual promiscuity than prompts involving men \citep{nozza-etal-2021-honest}.
Masked models predict gendered names and pronouns such that model-generated text is more likely to describe heterosexual relationships \citep{felkner-etal-2022-towards}.
While such research is important, many of these results assume gender binaries that contribute to gender exclusion and erasure \citep{dev-etal-2021-harms}.
Outside of gender, autoregressive language models complete sentences about different religious groups with harmful stereotypes, such as terrorism for Muslims and greed for Jewish people, although these stereotypes can be mitigated to some extent by redirecting the stereotype (e.g. ``the hard-working Muslim'';  \citealp{abid-etal-2021-persistent}).

Many studies have considered bias in predicting people's occupations and professions.
Occupation predictions from autoregressive language models are biased by given continental name origins and explicitly stated identities, with correlations with official labor statistics in the United States; occupational biases based on gender in language models are slightly less skewed than true labor statistics \citep{kirk-etal-2021-bias}.
Similarly, when predicting gendered pronouns given a known occupation, masked language model predictions are correlated with labor statistics on gender \citep{bartl-etal-2020-unmasking,manela-etal-2021-stereotype}, although predictions are sensitive to the specific prompt sentence \citep{touileb-2022-exploring}.
In autoregressive models, gendered pronoun predictions based on occupations are more biased in simple templates than in natural sentences from Wikipedia \citep{alnegheimish-etal-2022-using}.
Some studies find larger gender occupation biases in larger models \citep{tal-etal-2022-fewer,srivastava-etal-2022-beyond}, but these effects are inconsistent \citep{manela-etal-2021-stereotype,alnegheimish-etal-2022-using}.

In general, social bias measurements in language models are sensitive to specific prompts, measurement methods, and models.
Across different pre-training runs, masked models exhibit different levels of preference for stereotypical descriptions of people, particularly for individual demographic groups, despite similar downstream task performance \citep{aribandi-etal-2021-how}.
Gender occupation biases fluctuate significantly during model pre-training, even after the loss has plateaued \citep{tang-jiang-2022-gender}.
Results when predicting gendered pronouns in potentially biased scenarios are sensitive to paraphrasing and punctuation changes in the prompt \citep{seshadri-etal-2022-quantifying}; prompt and metric choices lead to noisy results for gender occupation bias in autoregressive models as well \citep{mattern-etal-2022-understanding,akyurek-etal-2022-challenges}.
Despite improving logical reasoning, prompting GPT-3 to ``think step-by-step'' (Section\autoref{sec:step-reasoning}) increases the probability that the model will generate stereotypical answers to questions, based on people's race, gender, religion, and other demographic identities \citep{shaikh-etal-2022-on}.
Effects of social biases in general appear to increase with model size across bias measurement tasks \citep{srivastava-etal-2022-beyond}.
Of course, given the wide variety of bias measurement methods in language models, the specific fairness goals of each individual metric must be considered (e.g. pairwise group fairness, group against baseline fairness, and/or overall between-group fairness; \citealp{czarnowska-etal-2021-quantifying}).

\bigskip
\section{Misinformation, Personality, and Politics}
\label{sec:misinformation-personality-politics}
Even outside of toxic and harmfully biased text, language models sometimes generate unfactual and misleading text.
They generate convincing unfactual text (Section\autoref{sec:misinformation}) that is difficult to distinguish from human-generated text (Section\autoref{sec:human-vs-model}), and their generated text depends on the political leaning and perceived personality of the input context (Section\autoref{sec:personality-politics}).
These behaviors can be more difficult to detect than explicitly biased and toxic text, because the outputs are often more subjective or controversial, and they primarily emerge in large models (Section\autoref{sec:scale}).
As noted in Section\autoref{sec:commonsense-knowledge}, factual knowledge in language models can be improved by using search and retrieval-enhanced models (e.g. \citealp{guu-etal-2020-retrieval,borgeaud-etal-2022-improving,schick-etal-2023-toolformer}); more fine-grained control over model outputs can be accomplished by conditioning the models on specific input data using controlled text generation \citep{li-etal-2021-pretrained,zhang-etal-2023-survey}.

\subsection{Language models can generate convincing unfactual text and unsafe advice.}
\label{sec:misinformation}
As they scale, autoregressive language models are more likely to generate text that affirms a conspiracy theory as fact when prompted with a conspiracy-related topic \citep{levy-etal-2021-investigating}.
They are also more likely to affirm common misconceptions (e.g. ``If you crack your knuckles a lot, you may develop arthritis''; \citealp{lin-etal-2022-truthfulqa}), although this result is inconsistent across studies \citep{rae-etal-2021-scaling}.
Larger models tend to be more consistent in their responses, producing semantically similar responses to semantically similar prompts, regardless of whether their responses are factually correct \citep{raj-etal-2022-measuring}.
Given access to internal model states, automated methods can identify text prompts that induce specific stances to common controversial topics \citep{heidenreich-williams-2021-the}.
Perhaps worryingly, people are more likely to rate GPT-3 generated tweets as true than human-generated tweets about vaccines, COVID-19, climate change, and other topics, regardless of whether they are factual or not \citep{spitale-etal-2023-ai}.
Conversations with GPT-3 can lead people to change their opinions on topics such as BLM (Black Lives Matter) and climate change \citep{chen-etal-2022-a-critical}.

Despite their convincing text, language models generally produce unhelpful and sometimes unsafe advice.
GPT-3 produces worse advice than people 95\% of the time in situations described on Reddit \citep{zellers-etal-2020-evaluating}.
Given a fill-in-the-blank task for stock market decisions, masked models have a preference to buy stocks rather than sell them, and they prefer specific stock categories such as utilities and materials \citep{chuang-yang-2022-buy}.
Although autoregressive models only rarely generate physically unsafe advice on their own (about 1\% of prompt responses), they predict slightly higher probabilities for unsafe than safe completions when given two possible options \citep{levy-etal-2022-safetext}.
When provided with a social rule and a described scenario with potentially-permissible rule-breaking behavior, both masked and autoregressive models only agree with human permissibility ratings marginally above chance \citep{jin-etal-2022-when}.

\subsection{Model-generated text is difficult to distinguish from human-generated text.}
\label{sec:human-vs-model}
Despite subtle differences between human and language model generated text (Section\autoref{sec:novel-text}), people have difficulty distinguishing the two, particularly as language models scale \citep{brown-etal-2020-language}.
People can only distinguish news articles generated by 175B parameter autoregressive models from human-generated articles with 52\% accuracy (compared to 50\% random chance; \citealp{brown-etal-2020-language}).
Similar accuracies are reported when people are asked to identify GPT-3 paraphrased Wikipedia paragraphs \citep{wahle-etal-2022-how} and GPT-3 generated tweets \citep{spitale-etal-2023-ai}.
People are better at identifying language model generated text in longer sequences \citep{ippolito-etal-2019-automatic}, but even when provided with specialized instructions and examples, people only reach about 55\% accuracy \citep{clark-etal-2021-all}.
In passages partially generated by smaller autoregressive models (e.g. 1.5B parameters), artificial intelligence graduate students are able to identify where the model-generated text begins with 23\% accuracy relative to 10\% random chance \citep{dugan-etal-2022-real}.

In general, people correctly assume that human-generated text is more sensical (e.g. less commonsense errors) and less repetitive than model-generated text \citep{clark-etal-2021-all,jakesch-etal-2022-human}. 
However, people also tend to predict that text is human-generated when it is more grammatical, uses shorter words, and contains more frequent bigrams; in reality, human-generated text is less grammatical, uses slightly longer words, and contains fewer frequent bigrams than model-generated text \citep{jakesch-etal-2022-human}.
With fine-tuning or given examples, language models themselves achieve better performance than people at identifying model-generated text, but they still have relatively low accuracy overall \citep{jawahar-etal-2020-automatic,wahle-etal-2022-how}.
To combat these difficulties in distinguishing human vs. model generated text, researchers have proposed ``watermarking'' model-generated text by slightly increasing the probabilities of ``whitelist'' tokens during text generation \citep{kirchenbauer-etal-2023-a}, or by explicitly replacing some tokens with whitelist tokens \citep{he-etal-2021-protecting}.

\subsection{Language model ``personality'' and politics depend on the input context.}
\label{sec:personality-politics}
Recent studies have found that language models generally mimic the political leanings and personality traits implied by a given input.
For example, larger autoregressive models are more likely to repeat political views expressed in a provided prompt \citep{perez-etal-2022-discovering}.
When prompted with a liberal vs. conservative identity (e.g. ``As a liberal, ...'') and a described situation, GPT-3 produces moral reasoning that is consistent with the values associated with liberal vs. conservative ideologies in moral foundations theory \citep{simmons-2022-moral}.
When prompted with a person's demographic information or personal background as context, GPT-3 produces similar words to describe political parties as that person, and it even predicts similar voting patterns and multiple choice responses to political surveys \citep{argyle-etal-2022-out}.
Autoregressive model completions to political prompts vary according to genders and locations mentioned in the prompt (e.g. United States states with different political leanings), although they tend to generate liberal-leaning text overall \citep{liu-etal-2022-quantifying}.
When asked to summarize text, GPT-3 shifts values in the input text towards United States moral and political values as opposed to values from other countries \citep{johnson-etal-2022-the}.
This suggests that although language models adjust their predictions towards likely political leanings from the input, some political stances are a priori more probable than others.

Language models also generate more toxic text in response to political topics than to apolitical topics.
Autoregressive models tuned for dialogue generate hyperpartisan responses to neutral political prompts over 50\% of the time and offensive responses 30\% of the time; the probability of hyperpartisan responses increases with politically biased prompts \citep{bang-etal-2021-assessing}.
These models are also more likely to generate insults in response to controversial topics such as BLM or MeToo than to less emotionally charged topics such as veganism or WFH (work from home) \citep{sheng-etal-2021-nice}.
Linguistic bias cues (e.g. ``claimed'' vs. ``stated'') increase the non-neutral sentiment of generated text in autoregressive models \citep{patel-pavlick-2021-was}.
When people converse with GPT-3 about controversial topics, people with minority opinions or less formal educational background report lower satisfaction with the interaction, often due to more negative responses from the model \citep{chen-etal-2022-a-critical}.

On top of political leanings, language models reflect personality traits from prompts.
When prompted with a person's self description of their personality, both masked and autoregressive language models complete Big Five personality surveys similarly to that person; however, the models score low on agreeableness and openness to experience regardless of prompt \citep{caron-srivastava-2022-identifying}.
GPT-3 exhibits similar effects, answering personality questions similarly to personalities described in given prompts \citep{jiang-etal-2022-mpi}.
Without prompting, autoregressive models have high psychopathy scores and low self-satisfaction scores on psychometric surveys \citep{li-etal-2022-is}.
However, GPT-3 responses to psychometric and demographic surveys vary significantly depending on sampling temperature (Section\autoref{sec:downstream}), resulting in different self-reported age, gender, personality, and values \citep{miotto-etal-2022-who}.
When given prompts describing classic psychology experiments (e.g. the Milgram Shock Experiment), GPT-3 replicates average human results to a reasonable degree \citep{aher-etal-2022-using}.
Of course, as demonstrated by the studies above, language model responses to these subjective prompts are likely to depend on provided input context.

\bigskip
\section{Discussion}
The previous sections discuss a wide range of language model capabilities and weaknesses, covering syntax, semantics, pragmatics, world knowledge, reasoning, memorization, and bias.
In this section, we synthesize these results framed from the perspectives of model scale (Section\autoref{sec:scale}) and text pattern generalization (Section\autoref{sec:language-modeling-generalization}), and we highlight recent research tying behavioral results to mechanistic analyses of language model internals (Section\autoref{sec:levels-of-analysis}).

\subsection{Effects of scale}
\label{sec:scale}
Recent work has increasingly focused on the impact of language model ``scale'' on model capabilities \citep{kaplan-etal-2020-scaling,hendrycks-etal-2020-measuring,rae-etal-2021-scaling,tay-etal-2021-scale,tay-etal-2022-scaling}, and public language model releases often include multiple model sizes for evaluation \citep{brown-etal-2020-language,zhang-etal-2022-opt}.
Language model scale is traditionally measured by number of parameters, usually between 100M and 500B parameters, although recent studies have also measured model scale using required computation during pre-training (FLOPs; \citealp{wei-etal-2022-emergent,wei-etal-2022-inverse}). 
Scaling research focuses on autoregressive language models, which exhibit substantial performance  improvements on many text generation tasks as they scale; fewer studies evaluate how model scale affects masked language model behavior \citep{artetxe-etal-2022-on}.
Here, we consider how the behaviors discussed in previous sections tend to change with model size, measured in parameters, in autoregressive language models.

Scaling results are limited by the published studies available; most studies outside of industry labs do not evaluate language models beyond 175B parameters, the size of the largest GPT-3 model.
Some tasks, such as domain-specific question-answering, arithmetic, logical event ordering, and proverb prediction exhibit unexpectedly large performance gains beyond 175B parameters \citep{wei-etal-2022-emergent,chowdhery-etal-2022-palm}.
Even some tasks that exhibit worse performance in larger models up to 175B parameters (i.e. ``inverse scaling'') exhibit sudden performance improvements beyond 175B parameters (i.e. ``U-shaped scaling''); many of these tasks contain a ``distractor'' feature or subtask that medium-sized models learn, but that large models can successfully ignore \citep{wei-etal-2022-inverse}.
In language modeling overall, the examples learned successfully by larger models are roughly a superset of the examples learned by smaller models \citep{xia-etal-2022-training}.
For some examples that are not successfully learned in 1B parameter models, models over 5B parameters exhibit an initial phase where their loss increases during pre-training before the examples are eventually learned \citep{xia-etal-2022-training}.
Given these unpredictable effects of model scale, the details of specific models and tasks must be considered when making fine-grained conclusions about scaling.

Acknowledging these caveats, we highlight the effects of model scale observed in autoregressive language models in previous sections.
Larger models learn syntactic rules more robustly than smaller models, but models across scales still generate grammatical text in most cases (Section\autoref{sec:syntax-overall}).
Larger models are worse at recognizing negation (Section\autoref{sec:negation}) but better at recognizing figurative language (Section\autoref{sec:figurative}).
They are more sensitive to the implied mental states of characters in text, but models across scales still struggle with pragmatics (Section\autoref{sec:pragmatics}).
Larger models learn more commonsense properties of objects and facts (Section\autoref{sec:commonsense-facts}), more fine-grained word properties (Section\autoref{sec:lexical-semantics}), and more correct arithmetic (Section\autoref{sec:numerical-reasoning}), but this may be because they memorize more examples during pre-training (Section\autoref{sec:memorization}; see also under-generalization in Section\autoref{sec:language-modeling-generalization}).
Large models (e.g. over 100B parameters) can be prompted to generate explicit multi-step reasoning by asking them to ``think step by step'' (\citealp{kojima-etal-2022-large}; Section\autoref{sec:step-reasoning}), but logical reasoning overall improves only slightly beyond around 10B parameters \citep{rae-etal-2021-scaling}.
Model size appears to have little impact on offensive text generation (Section\autoref{sec:toxicity}), but text generated by larger models is harder to distinguish from human-generated text (Section\autoref{sec:human-vs-model}), and larger models are more likely to mimic political opinions in a given input (Section\autoref{sec:personality-politics}).
The prevalence of harmful social biases in language models is inconsistent both within and across model sizes (Section\autoref{sec:bias-stereotypes}).
Overall, larger language models tend to exhibit equal or better performance to smaller models on most tasks, but their performance is still far from perfect, and they come at a higher environmental and computational cost \citep{strubell-etal-2019-energy}.

\subsection{Language modeling as generalization}
\label{sec:language-modeling-generalization}

\textbf{Text pattern generalization.} Many of the strengths and weaknesses of language models can be viewed through the lens of text pattern generalization.
Over-generalizations and under-generalizations of learned patterns in text simultaneously provide insights into the impressive capabilities and brittle responses of large language models \citep{ganguli-etal-2022-predictability}.
Specifically, due to the productivity of language (i.e. infinitely many combinations of patterns; \citealp{piantadosi-fedorenko-2017-infinitely}), language models must learn to generalize to novel examples, even when those examples would traditionally be considered ``in-distribution'' in generalization research (i.e. within the expected range of examples seen during pre-training; \citealp{hupkes-etal-2022-state}).
The in-distribution generalizations made by language models provide insights into how the models will likely behave in practice.

Through their token prediction training paradigm, language models are trained to generalize from text examples observed during pre-training to novel examples.
Given the beginning of a sentence never observed during pre-training, a language model can generate plausible completions to that sentence, similar to people generalizing from past experience to novel sentences \citep{piantadosi-fedorenko-2017-infinitely}.
Again similar to in people  \citep{prefors-etal-2006-poverty,berwick-etal-2011-poverty,dkabrowska-2015-what}, there are infinitely many generalization approaches that a language model can apply to extrapolate from pre-training examples (e.g. linear vs. hierarchical syntactic generalizations; \citealp{mccoy-etal-2018-revisiting,white-cotterell-2021-examining}).
Any text pattern that predicts upcoming tokens can under-influence or over-influence language model predictions (i.e. under-generalization vs. over-generalization), both in the set of examples to which the pattern is applied and the extent to which the pattern affects model predictions.
The specific generalizations that a language model learns are dependent on the language data observed and inherent biases from the model architecture and random initialization, also known as inductive biases \citep{white-cotterell-2021-examining}.

For example, one generalization approach might be to strictly memorize all training examples verbatim; the output token distribution for any observed example would be exactly equal to the distribution observed during pre-training, and any example not observed verbatim during pre-training would produce a random uniform distribution or some other degenerate prediction.
This would be an example of under-generalization, as the model assumes that each individual example does not reflect any patterns that can be generalized to other examples.
In practice, while language models do exhibit memorization of examples (Section\autoref{sec:memorization}), they appear to still extrapolate learned patterns from the memorized examples without overfitting \citep{tirumala-etal-2022-memorization}, suggesting that they are not entirely under-generalizing.

On the other end of the spectrum, a language model might always generate the most frequent token (e.g. ``the'') or condition only on the previous token (i.e. a bigram model).
Language models pass through both of these stages during pre-training \citep{chang-bergen-2021-word}.
These are examples of over-generalization, where token frequency rules and bigram rules over-influence model predictions.
In many cases, this over-generalization may occur due to under-generalization of other rules that would otherwise refine the over-generalized prediction.
Viewing these errors as generalization errors ties language model analysis research to broader generalization research in machine learning and NLP \citep{hupkes-etal-2022-state}.

\smallskip
\noindent
\textbf{Generalizations in language models.}
Indeed, many of the weaknesses exhibited by large language models can be interpreted as examples  over-generalization or under-generalization.
For example, language models' sensitivity to intervening clauses and specific words in subject-verb agreement reflects under-generalization of the subject-verb agreement rule (Section\autoref{sec:agreement}). 
Similarly, the models' sensitivity to paraphrasing and punctuation changes when recalling facts (Section\autoref{sec:facts-unreliable}) reflects under-generalization of learned facts.
Finally, the models' sensitivity to specific inputs when constructing situation models (Section\autoref{sec:situation-models}) and performing logical and numerical reasoning (Section\autoref{sec:reasoning}) reflects a systematic under-generalization of many patterns and rules to novel contexts.

Specifically, the models' reliance on pre-training corpus frequency for subject-verb agreement (Section\autoref{sec:agreement}), facts (Section\autoref{sec:facts-unreliable}), word meanings (Section\autoref{sec:lexical-semantics}), and arithmetic (Section\autoref{sec:numerical-reasoning}) might suggest that language models require many examples to correctly generalize some patterns, or it might suggest that the models are simply memorizing many under-generalized instances of each pattern.
Given the models' sensitivity to specific inputs for these capabilities, the memorization case appears more likely, e.g. that the models memorize many examples of arithmetic with minimal generalization.
Of course, these examples of under-generalization are not as severe as the models' inability to learn (and therefore under-generalization of) negation (Section\autoref{sec:negation}), pragmatics (Section\autoref{sec:pragmatics}), and many commonsense inferences (Sections\autoref{sec:commonsense-facts} and\autoref{sec:commonsense-events}).
In some of these cases, the language modeling objective may simply not capture the grounded and interactive features necessary to learn such patterns.

Language models also exhibit cases of over-generalization, often when some other under-generalized pattern fails to be applied.
When models fail to recall facts (Section\autoref{sec:facts-unreliable}), make commonsense inferences (Section\autoref{sec:commonsense-events}), or solve mathematical word problems (Section\autoref{sec:numerical-reasoning}), they often fall back to over-generalized heuristics such as predicting semantically similar tokens to the input context (Section\autoref{sec:novel-text}).
Overreliance on token position-based patterns (e.g. local $n$-grams) may reflect an over-generalization of position-based patterns as well (Sections\autoref{sec:position} and\autoref{sec:novel-text}).
Furthermore, harmful social biases in language models (Sections\autoref{sec:bias-performance} and\autoref{sec:bias-stereotypes}) can be interpreted as over-generalizations of patterns observed in the pre-training corpus.
Even when harmful biases are present in the pre-training corpus due to human social biases and dataset demographic imbalances, it is not desirable for language models to generalize these patterns.

Understanding when language models generalize correctly vs. incorrectly is important for the safe deployment of the models in practice.
Future work in language model behavioral analysis might consider the specific linguistic patterns and types of patterns that language models over-generalize and under-generalize, along with mitigation strategies.
In particular, future research might consider how generalization patterns change with model scale; it remains unclear to what extent the benefits of model scale are due to (1) learning more robust and/or correct generalized patterns or (2) memorizing a larger number of specific under-generalized instances that together improve performance metrics.
Again, given the models' sensitivity to specific inputs even in larger models, the models appear to lean towards the latter.

\subsection{Levels of analysis in understanding language models} 
\label{sec:levels-of-analysis}
As stated in the Introduction (Section\autoref{sec:scope}), this survey focuses on behavioral analyses of language models.
Other studies have investigated the internal mechanisms that lead language models to generate their predictions.
These two approaches roughly mirror Marr's computational and algorithmic levels of analysis in cognitive science, describing respectively (1) what the system does functionally and (2) the algorithms and representations the system uses to accomplish these functions \citep{marr-2010-vision,bechtel-shagrir-2015-the,trott-2023-in}.
Marr's last level, the implementation level, would correspond most closely to the physical circuits and neuron-level backpropagation rules that govern neural network models.
In many ways, the goals of language model analysis are to identify interpretable and generalizable principles that govern how language models work behaviorally and mechanistically, along with causal links between the two.

At the mechanistic (i.e. algorithmic) level, previous studies have probed the linguistic (and non-linguistic) information that can be extracted from language models' internal vector representations of tokens \citep{tenney-etal-2019-bert,rogers-etal-2020-a,belinkov-2021-probing}, along with how the representation spaces are structured geometrically \citep{coenen-etal-2019-visualizing,cai-etal-2021-isotropy,chang-etal-2022-the}.
They have also studied whether the attention weights assigned by language models' internal attention mechanism correlate with interpretable inter-token relationships \citep{clark-etal-2019-what,kovaleva-etal-2019-revealing,vig-belinkov-2019-analyzing}, although the attention weights do not necessarily influence language modeling predictions in expected ways \citep{jain-wallace-2019-attention,serrano-smith-2019-is}.

More recent work has established causal links between individual neurons (i.e. entries in the models' vector representations) and language modeling predictions \citep{vig-etal-2020-causal,geva-etal-2020-transformer,finlayson-etal-2021-causal,geva-etal-2022-transformer}.
For example, model representations of tokens at any layer can be interpreted as probability distributions over the language model vocabulary using the language model's output vocabulary projection matrix \citep{geva-etal-2022-transformer}; model parameters themselves can be interpreted using the same projections \citep{dar-etal-2022-analyzing}.
Parameter-level interventions can modify factual associations in language models in targeted ways \citep{meng-etal-2022-locating}, establishing direct connections between language model behavior and internal mechanisms.

Causal functionalities have also been established for individual attention heads in language models, e.g. for copying previous sequences from the input \citep{olsson-etal-2022-in}.
The attention mechanism has even been viewed as an in-context implementation of gradient descent, facilitating in-context learning (Section\autoref{sec:downstream}) without explicit parameter updates \citep{dai-etal-2022-why}.
Future work might apply similar analysis techniques to investigate the mechanisms underlying a wider range of language model behaviors, including under-generalized and over-generalized behaviors (Section\autoref{sec:language-modeling-generalization}), bridging the gap between behavioral and mechanistic levels of language model analysis.

\bigskip
\section{Conclusion}
In this survey, we have discussed a wide range of language model capabilities and weaknesses, covering over 250 studies of language model behavior from the past three years.
We find that language models remain sensitive to specific inputs and surface features even as they scale to hundreds of billions of parameters.
Many model strengths and weaknesses can be framed as correct or incorrect generalizations of text patterns.
By distilling what is currently known about large language model capabilities, we hope to inform the deployment and regulation of large language models, while also inspiring future language model analysis research.

\vfill
\pagebreak

\begin{acknowledgments}
We would like to thank the other members of the UCSD Language and Cognition Lab for helpful discussions. Tyler Chang is partially supported by the UCSD HDSI graduate fellowship.
\end{acknowledgments}

\color{black}
\starttwocolumn
\bibliography{final_lit_review}

\begin{thebibliography}{393}
\expandafter\ifx\csname natexlab\endcsname\relax\def\natexlab#1{#1}\fi

\bibitem[{Abdou et~al.(2022)Abdou, Ravishankar, Kulmizev, and
  S{\o}gaard}]{ravishankar-etal-2022-word}
Abdou, Mostafa, Vinit Ravishankar, Artur Kulmizev, and Anders S{\o}gaard. 2022.
\newblock Word order does matter and shuffled language models know it.
\newblock In \emph{Proceedings of the 60th Annual Meeting of the Association
  for Computational Linguistics (Volume 1: Long Papers)}, pages 6907--6919.

\bibitem[{Abid, Farooqi, and Zou(2021)}]{abid-etal-2021-persistent}
Abid, Abubakar, Maheen Farooqi, and James Zou. 2021.
\newblock Persistent anti-muslim bias in large language models.
\newblock In \emph{The AAAI/ACM Conference on AI, Ethics, and Society}, pages
  298--306.

\bibitem[{Adolphs, Dhuliawala, and Hofmann(2021)}]{adolphs-etal-2021-how}
Adolphs, Leonard, Shehzaad Dhuliawala, and Thomas Hofmann. 2021.
\newblock How to query language models?
\newblock \emph{ArXiv}, arXiv:2108.01928.

\bibitem[{Aher, Arriaga, and Kalai(2022)}]{aher-etal-2022-using}
Aher, Gati, Rosa Arriaga, and Adam Kalai. 2022.
\newblock Using large language models to simulate multiple humans.
\newblock \emph{ArXiv}, arXiv:2208.10264.

\bibitem[{Aina and Linzen(2021)}]{aina-linzen-2021-the}
Aina, Laura and Tal Linzen. 2021.
\newblock The language model understood the prompt was ambiguous: Probing
  syntactic uncertainty through generation.
\newblock In \emph{Proceedings of the Fourth BlackboxNLP Workshop on Analyzing
  and Interpreting Neural Networks for NLP}, pages 42--57.

\bibitem[{Aky{\"u}rek et~al.(2022)Aky{\"u}rek, Kocyigit, Paik, and
  Wijaya}]{akyurek-etal-2022-challenges}
Aky{\"u}rek, Afra~Feyza, Muhammed~Yusuf Kocyigit, Sejin Paik, and Derry~Tanti
  Wijaya. 2022.
\newblock Challenges in measuring bias via open-ended language generation.
\newblock In \emph{Proceedings of the 4th Workshop on Gender Bias in Natural
  Language Processing (GeBNLP)}, pages 76--76.

\bibitem[{Alnegheimish, Guo, and Sun(2022)}]{alnegheimish-etal-2022-using}
Alnegheimish, Sarah, Alicia Guo, and Yi~Sun. 2022.
\newblock Using natural sentence prompts for understanding biases in language
  models.
\newblock In \emph{Proceedings of the 2022 Conference of the North American
  Chapter of the Association for Computational Linguistics: Human Language
  Technologies}, pages 2824--2830.

\bibitem[{Apidianaki and Gar{\'i}~Soler(2021)}]{apidianaki-soler-2021-all}
Apidianaki, Marianna and Aina Gar{\'i}~Soler. 2021.
\newblock {ALL} dolphins are intelligent and {SOME} are friendly: Probing
  {BERT} for nouns{'} semantic properties and their prototypicality.
\newblock In \emph{Proceedings of the Fourth BlackboxNLP Workshop on Analyzing
  and Interpreting Neural Networks for NLP}, pages 79--94.

\bibitem[{Arefyev et~al.(2020)Arefyev, Sheludko, Podolskiy, and
  Panchenko}]{arefyev-etal-2020-a}
Arefyev, Nikolay, Boris Sheludko, Alexander Podolskiy, and Alexander Panchenko.
  2020.
\newblock Always keep your target in mind: Studying semantics and improving
  performance of neural lexical substitution.
\newblock In \emph{Proceedings of the 28th International Conference on
  Computational Linguistics}, pages 1242--1255, International Committee on
  Computational Linguistics, Barcelona, Spain (Online).

\bibitem[{Argyle et~al.(2023)Argyle, Busby, Fulda, Gubler, Rytting, and
  Wingate}]{argyle-etal-2022-out}
Argyle, Lisa~P., Ethan~C. Busby, Nancy Fulda, Joshua~R. Gubler, Christopher
  Rytting, and David Wingate. 2023.
\newblock Out of one, many: Using language models to simulate human samples.
\newblock \emph{Political Analysis}, page 1–15.

\bibitem[{Aribandi, Tay, and Metzler(2021)}]{aribandi-etal-2021-how}
Aribandi, Vamsi, Yi~Tay, and Donald Metzler. 2021.
\newblock How reliable are model diagnostics?
\newblock In \emph{Findings of the Association for Computational Linguistics:
  ACL-IJCNLP 2021}, pages 1778--1785.

\bibitem[{Armeni, Honey, and Linzen(2022)}]{armeni-etal-2022-characterizing}
Armeni, Kristijan, Christopher Honey, and Tal Linzen. 2022.
\newblock Characterizing verbatim short-term memory in neural language models.
\newblock In \emph{Proceedings of the 26th Conference on Computational Natural
  Language Learning (CoNLL)}, pages 405--424.

\bibitem[{Aroca-Ouellette et~al.(2021)Aroca-Ouellette, Paik, Roncone, and
  Kann}]{ouellette-etal-2021-prost}
Aroca-Ouellette, St{'e}phane, Cory Paik, Alessandro Roncone, and Katharina
  Kann. 2021.
\newblock {PROST}: {P}hysical reasoning about objects through space and time.
\newblock In \emph{Findings of the Association for Computational Linguistics:
  ACL-IJCNLP 2021}, pages 4597--4608.

\bibitem[{Arora, Kaffee, and Augenstein(2022)}]{arora-etal-2022-probing}
Arora, Arnav, Lucie-Aim{\'e}e Kaffee, and Isabelle Augenstein. 2022.
\newblock Probing pre-trained language models for cross-cultural differences in
  values.
\newblock \emph{ArXiv}, arXiv:2203.13722.

\bibitem[{Artetxe et~al.(2022)Artetxe, Du, Goyal, Zettlemoyer, and
  Stoyanov}]{artetxe-etal-2022-on}
Artetxe, Mikel, Jingfei Du, Naman Goyal, Luke Zettlemoyer, and Veselin
  Stoyanov. 2022.
\newblock On the role of bidirectionality in language model pre-training.
\newblock In \emph{Findings of the Association for Computational Linguistics:
  EMNLP 2022}, pages 3973--3985.

\bibitem[{Bacon and Regier(2019)}]{bacon-regier-2019-does}
Bacon, Geoff and Terry Regier. 2019.
\newblock Does {BERT} agree? evaluating knowledge of structure dependence
  through agreement relations.
\newblock \emph{ArXiv}, arXiv:1908.09892.

\bibitem[{Bang et~al.(2021)Bang, Lee, Ishii, Madotto, and
  Fung}]{bang-etal-2021-assessing}
Bang, Yejin, Nayeon Lee, Etsuko Ishii, Andrea Madotto, and Pascale Fung. 2021.
\newblock Assessing political prudence of open-domain chatbots.
\newblock In \emph{Proceedings of the 22nd Annual Meeting of the Special
  Interest Group on Discourse and Dialogue}, pages 548--555.

\bibitem[{Bartl, Nissim, and Gatt(2020)}]{bartl-etal-2020-unmasking}
Bartl, Marion, Malvina Nissim, and Albert Gatt. 2020.
\newblock Unmasking contextual stereotypes: Measuring and mitigating {BERT}{'}s
  gender bias.
\newblock In \emph{Proceedings of the Second Workshop on Gender Bias in Natural
  Language Processing}, pages 1--16.

\bibitem[{Bechtel and Shagrir(2015)}]{bechtel-shagrir-2015-the}
Bechtel, William and Oron Shagrir. 2015.
\newblock The non-redundant contributions of {Marr}'s three levels of analysis
  for explaining information-processing mechanisms.
\newblock \emph{Topics in Cognitive Science}, 7(2):312--322.

\bibitem[{Belinkov(2022)}]{belinkov-2021-probing}
Belinkov, Yonatan. 2022.
\newblock Probing classifiers: Promises, shortcomings, and advances.
\newblock \emph{Computational Linguistics}, 48(1):207--219.

\bibitem[{Beloucif and Biemann(2021)}]{beloucif-biemann-2021-probing}
Beloucif, Meriem and Chris Biemann. 2021.
\newblock Probing pre-trained language models for semantic attributes and their
  values.
\newblock In \emph{Findings of the Association for Computational Linguistics:
  EMNLP 2021}, pages 2554--2559.

\bibitem[{Bender et~al.(2021)Bender, Gebru, McMillan-Major, and
  Shmitchell}]{bender-etal-2021-on}
Bender, Emily~M., Timnit Gebru, Angelina McMillan-Major, and Shmargaret
  Shmitchell. 2021.
\newblock On the dangers of stochastic parrots: Can language models be too big?
\newblock In \emph{Proceedings of the ACM Conference on Fairness,
  Accountability, and Transparency}, page 610–623, Association for Computing
  Machinery, New York, NY, USA.

\bibitem[{Bender and Koller(2020)}]{bender-koller-2020-climbing}
Bender, Emily~M. and Alexander Koller. 2020.
\newblock Climbing towards {NLU}: {On} meaning, form, and understanding in the
  age of data.
\newblock In \emph{Proceedings of the 58th Annual Meeting of the Association
  for Computational Linguistics}, pages 5185--5198.

\bibitem[{Berwick et~al.(2011)Berwick, Pietroski, Yankama, and
  Chomsky}]{berwick-etal-2011-poverty}
Berwick, Robert, Paul Pietroski, Beracah Yankama, and Noam Chomsky. 2011.
\newblock Poverty of the stimulus revisited.
\newblock \emph{Cognitive Science}, 35(7):1207--1242.

\bibitem[{Betz, Richardson, and Voigt(2021)}]{betz-etal-2021-thinking}
Betz, Gregor, Kyle Richardson, and C.~Voigt. 2021.
\newblock Thinking aloud: Dynamic context generation improves zero-shot
  reasoning performance of {GPT-2}.
\newblock \emph{ArXiv}, arXiv:2103.13033.

\bibitem[{Beyer, Lo{\'a}iciga, and Schlangen(2021)}]{beyer-etal-2021-is}
Beyer, Anne, Sharid Lo{\'a}iciga, and David Schlangen. 2021.
\newblock Is incoherence surprising? targeted evaluation of coherence
  prediction from language models.
\newblock In \emph{Proceedings of the 2021 Conference of the North American
  Chapter of the Association for Computational Linguistics: Human Language
  Technologies}, pages 4164--4173.

\bibitem[{Bhavya, Xiong, and Zhai(2022)}]{bhavya-etal-2022-analogy}
Bhavya, Bhavya, Jinjun Xiong, and ChengXiang Zhai. 2022.
\newblock Analogy generation by prompting large language models: A case study
  of instructgpt.
\newblock In \emph{Proceedings of the 15th International Conference on Natural
  Language Generation}, pages 298--312.

\bibitem[{Binz and Schulz(2023)}]{binz-schulz-2022-using}
Binz, Marcel and Eric Schulz. 2023.
\newblock Using cognitive psychology to understand {GPT-3}.
\newblock \emph{Proceedings of the National Academy of Sciences of the United
  States of America}, 120(6):e2218523120.

\bibitem[{Blodgett et~al.(2020)Blodgett, Barocas, Daum{\'e}~III, and
  Wallach}]{blodgett-etal-2020-language}
Blodgett, Su~Lin, Solon Barocas, Hal Daum{\'e}~III, and Hanna Wallach. 2020.
\newblock Language (technology) is power: A critical survey of {``}bias{''} in
  {NLP}.
\newblock In \emph{Proceedings of the 58th Annual Meeting of the Association
  for Computational Linguistics}, pages 5454--5476.

\bibitem[{Bommasani et~al.(2021)Bommasani, Hudson, Adeli, Altman, Arora, von
  Arx, Bernstein, Bohg, Bosselut, Brunskill, Brynjolfsson, Buch, Card,
  Castellon, Chatterji, Chen, Creel, Davis, Demszky, Donahue, Doumbouya,
  Durmus, Ermon, Etchemendy, Ethayarajh, Fei-Fei, Finn, Gale, Gillespie, Goel,
  Goodman, Grossman, Guha, Hashimoto, Henderson, Hewitt, Ho, Hong, Hsu, Huang,
  Icard, Jain, Jurafsky, Kalluri, Karamcheti, Keeling, Khani, Khattab, Koh,
  Krass, Krishna, Kuditipudi, Kumar, Ladhak, Lee, Lee, Leskovec, Levent, Li,
  Li, Ma, Malik, Manning, Mirchandani, Mitchell, Munyikwa, Nair, Narayan,
  Narayanan, Newman, Nie, Niebles, Nilforoshan, Nyarko, Ogut, Orr,
  Papadimitriou, Park, Piech, Portelance, Potts, Raghunathan, Reich, Ren, Rong,
  Roohani, Ruiz, Ryan, R'e, Sadigh, Sagawa, Santhanam, Shih, Srinivasan,
  Tamkin, Taori, Thomas, Tram{\`e}r, Wang, Wang, Wu, Wu, Wu, Xie, Yasunaga,
  You, Zaharia, Zhang, Zhang, Zhang, Zhang, Zheng, Zhou, and
  Liang}]{bommasani-etal-2021-on}
Bommasani, Rishi, Drew~A. Hudson, Ehsan Adeli, Russ Altman, Simran Arora,
  Sydney von Arx, Michael~S. Bernstein, Jeannette Bohg, Antoine Bosselut, Emma
  Brunskill, Erik Brynjolfsson, S.~Buch, Dallas Card, Rodrigo Castellon,
  Niladri~S. Chatterji, Annie~S. Chen, Kathleen~A. Creel, Jared Davis, Dora
  Demszky, Chris Donahue, Moussa Doumbouya, Esin Durmus, Stefano Ermon, John
  Etchemendy, Kawin Ethayarajh, Li~Fei-Fei, Chelsea Finn, Trevor Gale,
  Lauren~E. Gillespie, Karan Goel, Noah~D. Goodman, Shelby Grossman, Neel Guha,
  Tatsunori Hashimoto, Peter Henderson, John Hewitt, Daniel~E. Ho, Jenny Hong,
  Kyle Hsu, Jing Huang, Thomas~F. Icard, Saahil Jain, Dan Jurafsky, Pratyusha
  Kalluri, Siddharth Karamcheti, Geoff Keeling, Fereshte Khani, O.~Khattab,
  Pang~Wei Koh, Mark~S. Krass, Ranjay Krishna, Rohith Kuditipudi, Ananya Kumar,
  Faisal Ladhak, Mina Lee, Tony Lee, Jure Leskovec, Isabelle Levent, Xiang~Lisa
  Li, Xuechen Li, Tengyu Ma, Ali Malik, Christopher~D. Manning, Suvir~P.
  Mirchandani, Eric Mitchell, Zanele Munyikwa, Suraj Nair, Avanika Narayan,
  Deepak Narayanan, Benjamin Newman, Allen Nie, Juan~Carlos Niebles, Hamed
  Nilforoshan, J.~F. Nyarko, Giray Ogut, Laurel Orr, Isabel Papadimitriou,
  Joon~Sung Park, Chris Piech, Eva Portelance, Christopher Potts, Aditi
  Raghunathan, Robert Reich, Hongyu Ren, Frieda Rong, Yusuf~H. Roohani, Camilo
  Ruiz, Jack Ryan, Christopher R'e, Dorsa Sadigh, Shiori Sagawa, Keshav
  Santhanam, Andy Shih, Krishna~Parasuram Srinivasan, Alex Tamkin, Rohan Taori,
  Armin~W. Thomas, Florian Tram{\`e}r, Rose~E. Wang, William Wang, Bohan Wu,
  Jiajun Wu, Yuhuai Wu, Sang~Michael Xie, Michihiro Yasunaga, Jiaxuan You,
  Matei~A. Zaharia, Michael Zhang, Tianyi Zhang, Xikun Zhang, Yuhui Zhang,
  Lucia Zheng, Kaitlyn Zhou, and Percy Liang. 2021.
\newblock On the opportunities and risks of foundation models.
\newblock \emph{ArXiv}, arXiv:2108.07258.

\bibitem[{Borgeaud et~al.(2022)Borgeaud, Mensch, Hoffmann, Cai, Rutherford,
  Millican, Van Den~Driessche, Lespiau, Damoc, Clark, De~Las~Casas, Guy,
  Menick, Ring, Hennigan, Huang, Maggiore, Jones, Cassirer, Brock, Paganini,
  Irving, Vinyals, Osindero, Simonyan, Rae, Elsen, and
  Sifre}]{borgeaud-etal-2022-improving}
Borgeaud, Sebastian, Arthur Mensch, Jordan Hoffmann, Trevor Cai, Eliza
  Rutherford, Katie Millican, George~Bm Van Den~Driessche, Jean-Baptiste
  Lespiau, Bogdan Damoc, Aidan Clark, Diego De~Las~Casas, Aurelia Guy, Jacob
  Menick, Roman Ring, Tom Hennigan, Saffron Huang, Loren Maggiore, Chris Jones,
  Albin Cassirer, Andy Brock, Michela Paganini, Geoffrey Irving, Oriol Vinyals,
  Simon Osindero, Karen Simonyan, Jack Rae, Erich Elsen, and Laurent Sifre.
  2022.
\newblock Improving language models by retrieving from trillions of tokens.
\newblock In \emph{International Conference on Machine Learning}, pages
  2206--2240.

\bibitem[{Bowman(2022)}]{bowman-2021-when}
Bowman, Samuel. 2022.
\newblock The dangers of underclaiming: Reasons for caution when reporting how
  {NLP} systems fail.
\newblock In \emph{Proceedings of the 60th Annual Meeting of the Association
  for Computational Linguistics (Volume 1: Long Papers)}, pages 7484--7499.

\bibitem[{Brandl, Cui, and S{\o}gaard(2022)}]{brandl-etal-2022-how}
Brandl, Stephanie, Ruixiang Cui, and Anders S{\o}gaard. 2022.
\newblock How conservative are language models? adapting to the introduction of
  gender-neutral pronouns.
\newblock In \emph{Proceedings of the 2022 Conference of the North American
  Chapter of the Association for Computational Linguistics: Human Language
  Technologies}, pages 3624--3630.

\bibitem[{Brown et~al.(2022)Brown, Lee, Mireshghallah, Shokri, and
  Tram{\`e}r}]{brown-etal-2022-what}
Brown, Hannah, Katherine Lee, Fatemehsadat Mireshghallah, Reza Shokri, and
  Florian Tram{\`e}r. 2022.
\newblock What does it mean for a language model to preserve privacy?
\newblock In \emph{Proceedings of the ACM Conference on Fairness,
  Accountability, and Transparency}, pages 2280--2292.

\bibitem[{Brown et~al.(2020)Brown, Mann, Ryder, Subbiah, Kaplan, Dhariwal,
  Neelakantan, Shyam, Sastry, Askell, Agarwal, Herbert-Voss, Krueger, Henighan,
  Child, Ramesh, Ziegler, Wu, Winter, Hesse, Chen, Sigler, Litwin, Gray, Chess,
  Clark, Berner, McCandlish, Radford, Sutskever, and
  Amodei}]{brown-etal-2020-language}
Brown, Tom, Benjamin Mann, Nick Ryder, Melanie Subbiah, Jared~D Kaplan,
  Prafulla Dhariwal, Arvind Neelakantan, Pranav Shyam, Girish Sastry, Amanda
  Askell, Sandhini Agarwal, Ariel Herbert-Voss, Gretchen Krueger, Tom Henighan,
  Rewon Child, Aditya Ramesh, Daniel Ziegler, Jeffrey Wu, Clemens Winter, Chris
  Hesse, Mark Chen, Eric Sigler, Mateusz Litwin, Scott Gray, Benjamin Chess,
  Jack Clark, Christopher Berner, Sam McCandlish, Alec Radford, Ilya Sutskever,
  and Dario Amodei. 2020.
\newblock Language models are few-shot learners.
\newblock In \emph{Advances in Neural Information Processing Systems},
  volume~33, pages 1877--1901.

\bibitem[{Broyde and Palmer(2021)}]{broyde-palmer-2021-build}
Broyde, Joshua and Claire Palmer. 2021.
\newblock Build a medical sentence matching application using {BERT} and
  {Amazon} {SageMaker}.
\newblock \emph{{AWS} Machine Learning Blog}.

\bibitem[{Cai et~al.(2021)Cai, Huang, Bian, and
  Church}]{cai-etal-2021-isotropy}
Cai, Xingyu, Jiaji Huang, Yu-Lan Bian, and Kenneth~Ward Church. 2021.
\newblock Isotropy in the contextual embedding space: Clusters and manifolds.
\newblock In \emph{International Conference on Learning Representations}.

\bibitem[{Cao et~al.(2022)Cao, Lin, Han, Liu, and Sun}]{cao-etal-2022-can}
Cao, Boxi, Hongyu Lin, Xianpei Han, Fangchao Liu, and Le~Sun. 2022.
\newblock Can prompt probe pretrained language models? understanding the
  invisible risks from a causal view.
\newblock In \emph{Proceedings of the 60th Annual Meeting of the Association
  for Computational Linguistics (Volume 1: Long Papers)}, pages 5796--5808.

\bibitem[{Cao et~al.(2021)Cao, Lin, Han, Sun, Yan, Liao, Xue, and
  Xu}]{cao-etal-2021-knowledgeable}
Cao, Boxi, Hongyu Lin, Xianpei Han, Le~Sun, Lingyong Yan, Meng Liao, Tong Xue,
  and Jin Xu. 2021.
\newblock Knowledgeable or educated guess? revisiting language models as
  knowledge bases.
\newblock In \emph{Proceedings of the 59th Annual Meeting of the Association
  for Computational Linguistics and the 11th International Joint Conference on
  Natural Language Processing (Volume 1: Long Papers)}, pages 1860--1874.

\bibitem[{Carlini et~al.(2023)Carlini, Ippolito, Jagielski, Lee, Tramer, and
  Zhang}]{carlini-etal-2022-quantifying}
Carlini, Nicholas, Daphne Ippolito, Matthew Jagielski, Katherine Lee, Florian
  Tramer, and Chiyuan Zhang. 2023.
\newblock Quantifying memorization across neural language models.
\newblock In \emph{International Conference on Learning Representations}.

\bibitem[{Carlini et~al.(2021)Carlini, Tramer, Wallace, Jagielski,
  Herbert-Voss, Lee, Roberts, Brown, Song, Erlingsson, Oprea, and
  Raffel}]{carlini-etal-2020-extracting}
Carlini, Nicholas, Florian Tramer, Eric Wallace, Matthew Jagielski, Ariel
  Herbert-Voss, Katherine Lee, Adam Roberts, Tom Brown, Dawn Song, Ulfar
  Erlingsson, Alina Oprea, and Colin Raffel. 2021.
\newblock Extracting training data from large language models.
\newblock In \emph{USENIX Security Symposium}, pages 2633--2650.

\bibitem[{Carnie(2002)}]{carnie-2002-syntax}
Carnie, Andrew. 2002.
\newblock \emph{Syntax: A Generative Introduction}.
\newblock Blackwell.

\bibitem[{Caron and Srivastava(2022)}]{caron-srivastava-2022-identifying}
Caron, Graham and Shashank Srivastava. 2022.
\newblock Identifying and manipulating the personality traits of language
  models.
\newblock \emph{ArXiv}, arXiv:2212.10276.

\bibitem[{Chang, Tu, and Bergen(2022)}]{chang-etal-2022-the}
Chang, Tyler, Zhuowen Tu, and Benjamin Bergen. 2022.
\newblock The geometry of multilingual language model representations.
\newblock In \emph{Proceedings of the 2022 Conference on Empirical Methods in
  Natural Language Processing}, pages 119--136.

\bibitem[{Chang et~al.(2021)Chang, Xu, Xu, and
  Tu}]{chang-etal-2021-convolutions}
Chang, Tyler, Yifan Xu, Weijian Xu, and Zhuowen Tu. 2021.
\newblock Convolutions and self-attention: {R}e-interpreting relative positions
  in pre-trained language models.
\newblock In \emph{Proceedings of the 59th Annual Meeting of the Association
  for Computational Linguistics and the 11th International Joint Conference on
  Natural Language Processing (Volume 1: Long Papers)}, pages 4322--4333.

\bibitem[{Chang and Bergen(2022)}]{chang-bergen-2021-word}
Chang, Tyler~A. and Benjamin~K. Bergen. 2022.
\newblock Word acquisition in neural language models.
\newblock \emph{Transactions of the Association for Computational Linguistics},
  10:1--16.

\bibitem[{Chaves and Richter(2021)}]{chaves-richter-2021-look}
Chaves, Rui~P. and Stephanie~N. Richter. 2021.
\newblock Look at that! {BERT} can be easily distracted from paying attention
  to morphosyntax.
\newblock In \emph{Proceedings of the Society for Computation in Linguistics
  2021}, pages 28--38.

\bibitem[{Chen et~al.(2022)Chen, Shao, Burapacheep, and
  Li}]{chen-etal-2022-a-critical}
Chen, Kaiping, Anqi Shao, Jirayu Burapacheep, and Yixuan Li. 2022.
\newblock A critical appraisal of equity in conversational {AI}: Evidence from
  auditing {GPT-3}'s dialogues with different publics on climate change and
  {Black Lives Matter}.
\newblock \emph{ArXiv}, arXiv:2209.13627.

\bibitem[{Chen et~al.(2021)Chen, Tworek, Jun, Yuan, Ponde, Kaplan, Edwards,
  Burda, Joseph, Brockman, Ray, Puri, Krueger, Petrov, Khlaaf, Sastry, Mishkin,
  Chan, Gray, Ryder, Pavlov, Power, Kaiser, Bavarian, Winter, Tillet, Such,
  Cummings, Plappert, Chantzis, Barnes, Herbert-Voss, Guss, Nichol, Babuschkin,
  Balaji, Jain, Carr, Leike, Achiam, Misra, Morikawa, Radford, Knight,
  Brundage, Murati, Mayer, Welinder, McGrew, Amodei, McCandlish, Sutskever, and
  Zaremba}]{chen-etal-2021-evaluating}
Chen, Mark, Jerry Tworek, Heewoo Jun, Qiming Yuan, Henrique Ponde, Jared
  Kaplan, Harrison Edwards, Yura Burda, Nicholas Joseph, Greg Brockman, Alex
  Ray, Raul Puri, Gretchen Krueger, Michael Petrov, Heidy Khlaaf, Girish
  Sastry, Pamela Mishkin, Brooke Chan, Scott Gray, Nick Ryder, Mikhail Pavlov,
  Alethea Power, Lukasz Kaiser, Mohammad Bavarian, Clemens Winter, Philippe
  Tillet, F.~Such, D.~Cummings, Matthias Plappert, Fotios Chantzis, Elizabeth
  Barnes, Ariel Herbert-Voss, William~H. Guss, Alex Nichol, I.~Babuschkin,
  S.~Balaji, Shantanu Jain, A.~Carr, J.~Leike, Joshua Achiam, Vedant Misra,
  Evan Morikawa, Alec Radford, M.~Knight, Miles Brundage, Mira Murati, Katie
  Mayer, P.~Welinder, Bob McGrew, Dario Amodei, Sam McCandlish, Ilya Sutskever,
  and Wojciech Zaremba. 2021.
\newblock Evaluating large language models trained on code.
\newblock \emph{ArXiv}, arXiv:2107.03374.

\bibitem[{Chiang, Huang, and Lee(2020)}]{chiang-etal-2020-pretrained}
Chiang, Cheng-Han, Sung-Feng Huang, and Hung-yi Lee. 2020.
\newblock {P}retrained language model embryology: {T}he birth of {ALBERT}.
\newblock In \emph{Proceedings of the 2020 Conference on Empirical Methods in
  Natural Language Processing (EMNLP)}, pages 6813--6828.

\bibitem[{Chiang and Chen(2021)}]{chiang-chen-2021-relating}
Chiang, Ting-Rui and Yun-Nung Chen. 2021.
\newblock Relating neural text degeneration to exposure bias.
\newblock In \emph{Proceedings of the Fourth BlackboxNLP Workshop on Analyzing
  and Interpreting Neural Networks for NLP}, pages 228--239.

\bibitem[{Cho et~al.(2021)Cho, Chersoni, Hsu, and
  Huang}]{cho-etal-2021-modeling}
Cho, Won~Ik, Emmanuele Chersoni, Yu-Yin Hsu, and Chu-Ren Huang. 2021.
\newblock Modeling the influence of verb aspect on the activation of typical
  event locations with {BERT}.
\newblock In \emph{Findings of the Association for Computational Linguistics:
  ACL-IJCNLP 2021}, pages 2922--2929.

\bibitem[{Choenni, Shutova, and van
  Rooij(2021)}]{choenni-etal-2021-stepmothers}
Choenni, Rochelle, Ekaterina Shutova, and Robert van Rooij. 2021.
\newblock Stepmothers are mean and academics are pretentious: What do
  pretrained language models learn about you?
\newblock In \emph{Proceedings of the 2021 Conference on Empirical Methods in
  Natural Language Processing}, pages 1477--1491.

\bibitem[{Choshen et~al.(2022)Choshen, Hacohen, Weinshall, and
  Abend}]{choshen-etal-2021-the}
Choshen, Leshem, Guy Hacohen, Daphna Weinshall, and Omri Abend. 2022.
\newblock The grammar-learning trajectories of neural language models.
\newblock In \emph{Proceedings of the 60th Annual Meeting of the Association
  for Computational Linguistics (Volume 1: Long Papers)}, pages 8281--8297.

\bibitem[{Choudhury and Deshpande(2021)}]{choudhury-deshpande-2021-how}
Choudhury, Monojit and Amit Deshpande. 2021.
\newblock How linguistically fair are multilingual pre-trained language models?
\newblock In \emph{Proceedings of the AAAI Conference on Artificial
  Intelligence}, volume~35, pages 12710--12718.

\bibitem[{Chowdhery et~al.(2022)Chowdhery, Narang, Devlin, Bosma, Mishra,
  Roberts, Barham, Chung, Sutton, Gehrmann, Schuh, Shi, Tsvyashchenko, Maynez,
  Rao, Barnes, Tay, Shazeer, Prabhakaran, Reif, Du, Hutchinson, Pope, Bradbury,
  Austin, Isard, Gur-Ari, Yin, Duke, Levskaya, Ghemawat, Dev, Michalewski,
  Garc{\'i}a, Misra, Robinson, Fedus, Zhou, Ippolito, Luan, Lim, Zoph,
  Spiridonov, Sepassi, Dohan, Agrawal, Omernick, Dai, Pillai, Pellat,
  Lewkowycz, Moreira, Child, Polozov, Lee, Zhou, Wang, Saeta, D{\'i}az, Firat,
  Catasta, Wei, Meier-Hellstern, Eck, Dean, Petrov, and
  Fiedel}]{chowdhery-etal-2022-palm}
Chowdhery, Aakanksha, Sharan Narang, Jacob Devlin, Maarten Bosma, Gaurav
  Mishra, Adam Roberts, Paul Barham, Hyung~Won Chung, Charles Sutton, Sebastian
  Gehrmann, Parker Schuh, Kensen Shi, Sasha Tsvyashchenko, Joshua Maynez,
  Abhishek Rao, Parker Barnes, Yi~Tay, Noam~M. Shazeer, Vinodkumar Prabhakaran,
  Emily Reif, Nan Du, Benton~C. Hutchinson, Reiner Pope, James Bradbury, Jacob
  Austin, Michael Isard, Guy Gur-Ari, Pengcheng Yin, Toju Duke, Anselm
  Levskaya, Sanjay Ghemawat, Sunipa Dev, Henryk Michalewski, Xavier Garc{\'i}a,
  Vedant Misra, Kevin Robinson, Liam Fedus, Denny Zhou, Daphne Ippolito, David
  Luan, Hyeontaek Lim, Barret Zoph, Alexander Spiridonov, Ryan Sepassi, David
  Dohan, Shivani Agrawal, Mark Omernick, Andrew~M. Dai,
  Thanumalayan~Sankaranarayana Pillai, Marie Pellat, Aitor Lewkowycz, Erica
  Moreira, Rewon Child, Oleksandr Polozov, Katherine Lee, Zongwei Zhou, Xuezhi
  Wang, Brennan Saeta, Mark D{\'i}az, Orhan Firat, Michele Catasta, Jason Wei,
  Kathleen~S. Meier-Hellstern, Douglas Eck, Jeff Dean, Slav Petrov, and Noah
  Fiedel. 2022.
\newblock {PaLM}: Scaling language modeling with {Pathways}.
\newblock \emph{ArXiv}, arXiv:2204.02311.

\bibitem[{Chuang and Yang(2022)}]{chuang-yang-2022-buy}
Chuang, Chengyu and Yi~Yang. 2022.
\newblock Buy tesla, sell ford: Assessing implicit stock market preference in
  pre-trained language models.
\newblock In \emph{Proceedings of the 60th Annual Meeting of the Association
  for Computational Linguistics (Volume 2: Short Papers)}, pages 100--105.

\bibitem[{C{\'i}fka and Liutkus(2022)}]{cifka-liutkus-2022-blackbox}
C{\'i}fka, Ond{\v{r}}ej and Antoine Liutkus. 2022.
\newblock Black-box language model explanation by context length probing.
\newblock \emph{ArXiv}, arXiv:2212.14815.

\bibitem[{Clark et~al.(2021)Clark, August, Serrano, Haduong, Gururangan, and
  Smith}]{clark-etal-2021-all}
Clark, Elizabeth, Tal August, Sofia Serrano, Nikita Haduong, Suchin Gururangan,
  and Noah~A. Smith. 2021.
\newblock All that{'}s {`}human{'} is not gold: Evaluating human evaluation of
  generated text.
\newblock In \emph{Proceedings of the 59th Annual Meeting of the Association
  for Computational Linguistics and the 11th International Joint Conference on
  Natural Language Processing (Volume 1: Long Papers)}, pages 7282--7296.

\bibitem[{Clark et~al.(2019)Clark, Khandelwal, Levy, and
  Manning}]{clark-etal-2019-what}
Clark, Kevin, Urvashi Khandelwal, Omer Levy, and Christopher~D. Manning. 2019.
\newblock What does {BERT} look at? an analysis of {BERT}{'}s attention.
\newblock In \emph{Proceedings of the 2019 ACL Workshop BlackboxNLP: Analyzing
  and Interpreting Neural Networks for NLP}, pages 276--286.

\bibitem[{Com{\textcommabelow{s}}a, Eisenschlos, and
  Narayanan(2022)}]{comsa-etal-2022-miqa}
Com{\textcommabelow{s}}a, Iulia, Julian Eisenschlos, and Srini Narayanan. 2022.
\newblock {M}i{QA}: A benchmark for inference on metaphorical questions.
\newblock In \emph{Proceedings of the 2nd Conference of the Asia-Pacific
  Chapter of the Association for Computational Linguistics and the 12th
  International Joint Conference on Natural Language Processing (Volume 2:
  Short Papers)}, pages 373--381.

\bibitem[{Cong(2022)}]{cong-2022-psycholinguistic}
Cong, Yan. 2022.
\newblock Psycholinguistic diagnosis of language models{'} commonsense
  reasoning.
\newblock In \emph{Proceedings of the First Workshop on Commonsense
  Representation and Reasoning (CSRR 2022)}, pages 17--22.

\bibitem[{Czarnowska, Vyas, and Shah(2021)}]{czarnowska-etal-2021-quantifying}
Czarnowska, Paula, Yogarshi Vyas, and Kashif Shah. 2021.
\newblock Quantifying social biases in {NLP}: A generalization and empirical
  comparison of extrinsic fairness metrics.
\newblock \emph{Transactions of the Association for Computational Linguistics},
  9:1249--1267.

\bibitem[{Czinczoll et~al.(2022)Czinczoll, Yannakoudakis, Mishra, and
  Shutova}]{czinczoll-etal-2022-scientific}
Czinczoll, Tamara, Helen Yannakoudakis, Pushkar Mishra, and Ekaterina Shutova.
  2022.
\newblock Scientific and creative analogies in pretrained language models.
\newblock In \emph{Findings of the Association for Computational Linguistics:
  EMNLP 2022}, pages 2094--2100.

\bibitem[{D{\k{a}}browska(2015)}]{dkabrowska-2015-what}
D{\k{a}}browska, Ewa. 2015.
\newblock What exactly is {Universal} {Grammar}, and has anyone seen it?
\newblock \emph{Frontiers in Psychology}, 6:852.

\bibitem[{Dai et~al.(2022)Dai, Sun, Dong, Hao, Sui, and
  Wei}]{dai-etal-2022-why}
Dai, Damai, Yutao Sun, Li~Dong, Yaru Hao, Zhifang Sui, and Furu Wei. 2022.
\newblock Why can {GPT} learn in-context? language models secretly perform
  gradient descent as meta-optimizers.
\newblock \emph{ArXiv}, arXiv:2212.10559.

\bibitem[{Dai et~al.(2019)Dai, Yang, Yang, Carbonell, Le, and
  Salakhutdinov}]{dai-etal-2019-transformerxl}
Dai, Zihang, Zhilin Yang, Yiming Yang, Jaime Carbonell, Quoc Le, and Ruslan
  Salakhutdinov. 2019.
\newblock Transformer-{XL}: Attentive language models beyond a fixed-length
  context.
\newblock In \emph{Proceedings of the 57th Annual Meeting of the Association
  for Computational Linguistics}, pages 2978--2988.

\bibitem[{Dar et~al.(2022)Dar, Geva, Gupta, and
  Berant}]{dar-etal-2022-analyzing}
Dar, Guy, Mor Geva, Ankit Gupta, and Jonathan Berant. 2022.
\newblock Analyzing {Transformers} in embedding space.
\newblock \emph{ArXiv}, arXiv:2209.02535.

\bibitem[{Dasgupta et~al.(2022)Dasgupta, Lampinen, Chan, Creswell, Kumaran,
  McClelland, and Hill}]{dasgupta-etal-2022-language}
Dasgupta, Ishita, Andrew Lampinen, Stephanie Chan, Antonia Creswell, Dharshan
  Kumaran, James McClelland, and Felix Hill. 2022.
\newblock Language models show human-like content effects on reasoning.
\newblock \emph{ArXiv}, arXiv:2207.07051.

\bibitem[{Davis and van Schijndel(2020)}]{davis-schijndel-2020-discourse}
Davis, Forrest and Marten van Schijndel. 2020.
\newblock Discourse structure interacts with reference but not syntax in neural
  language models.
\newblock In \emph{Proceedings of the 24th Conference on Computational Natural
  Language Learning}, pages 396--407.

\bibitem[{Davison, Feldman, and Rush(2019)}]{feldman-etal-2019-commonsense}
Davison, Joe, Joshua Feldman, and Alexander Rush. 2019.
\newblock Commonsense knowledge mining from pretrained models.
\newblock In \emph{Proceedings of the 2019 Conference on Empirical Methods in
  Natural Language Processing and the 9th International Joint Conference on
  Natural Language Processing (EMNLP-IJCNLP)}, pages 1173--1178.

\bibitem[{De~Bruyn et~al.(2022)De~Bruyn, Lotfi, Buhmann, and
  Daelemans}]{bruyn-etal-2022-is}
De~Bruyn, Maxime, Ehsan Lotfi, Jeska Buhmann, and Walter Daelemans. 2022.
\newblock Is it smaller than a tennis ball? language models play the game of
  twenty questions.
\newblock In \emph{Proceedings of the Fifth BlackboxNLP Workshop on Analyzing
  and Interpreting Neural Networks for NLP}, pages 80--90.

\bibitem[{Dev et~al.(2021)Dev, Monajatipoor, Ovalle, Subramonian, Phillips, and
  Chang}]{dev-etal-2021-harms}
Dev, Sunipa, Masoud Monajatipoor, Anaelia Ovalle, Arjun Subramonian, Jeff
  Phillips, and Kai-Wei Chang. 2021.
\newblock Harms of gender exclusivity and challenges in non-binary
  representation in language technologies.
\newblock In \emph{Proceedings of the 2021 Conference on Empirical Methods in
  Natural Language Processing}, pages 1968--1994.

\bibitem[{Dev et~al.(2022)Dev, Sheng, Zhao, Amstutz, Sun, Hou, Sanseverino,
  Kim, Nishi, Peng, and Chang}]{dev-etal-2021-what}
Dev, Sunipa, Emily Sheng, Jieyu Zhao, Aubrie Amstutz, Jiao Sun, Yu~Hou, Mattie
  Sanseverino, Jiin Kim, Akihiro Nishi, Nanyun Peng, and Kai-Wei Chang. 2022.
\newblock On measures of biases and harms in {NLP}.
\newblock In \emph{Findings of the Association for Computational Linguistics:
  AACL-IJCNLP 2022}, pages 246--267.

\bibitem[{Devlin et~al.(2019)Devlin, Chang, Lee, and
  Toutanova}]{devlin-etal-2019-bert}
Devlin, Jacob, Ming-Wei Chang, Kenton Lee, and Kristina Toutanova. 2019.
\newblock {BERT}: Pre-training of deep bidirectional transformers for language
  understanding.
\newblock In \emph{Proceedings of the 2019 Conference of the North {A}merican
  Chapter of the Association for Computational Linguistics: Human Language
  Technologies, Volume 1 (Long and Short Papers)}, pages 4171--4186.

\bibitem[{Dhamala et~al.(2021)Dhamala, Sun, Kumar, Krishna, Pruksachatkun,
  Chang, and Gupta}]{dhamala-etal-2021-bold}
Dhamala, Jwala, Tony Sun, Varun Kumar, Satyapriya Krishna, Yada Pruksachatkun,
  Kai-Wei Chang, and Rahul Gupta. 2021.
\newblock {BOLD}: Dataset and metrics for measuring biases in open-ended
  language generation.
\newblock In \emph{Proceedings of the ACM Conference on Fairness,
  Accountability, and Transparency}, page 862–872, Association for Computing
  Machinery, New York, NY, USA.

\bibitem[{Dou et~al.(2022)Dou, Forbes, Koncel-Kedziorski, Smith, and
  Choi}]{dou-etal-2022-gpt}
Dou, Yao, Maxwell Forbes, Rik Koncel-Kedziorski, Noah~A. Smith, and Yejin Choi.
  2022.
\newblock Is {GPT}-3 text indistinguishable from human text? {S}carecrow: A
  framework for scrutinizing machine text.
\newblock In \emph{Proceedings of the 60th Annual Meeting of the Association
  for Computational Linguistics (Volume 1: Long Papers)}, pages 7250--7274.

\bibitem[{Du et~al.(2022{\natexlab{a}})Du, He, Zou, Tao, and
  Hu}]{du-etal-2022-shortcut}
Du, Mengnan, Fengxiang He, Na~Zou, Dacheng Tao, and Xia Hu. 2022{\natexlab{a}}.
\newblock Shortcut learning of large language models in natural language
  understanding: A survey.
\newblock \emph{ArXiv}, arXiv:2208.11857.

\bibitem[{Du et~al.(2022{\natexlab{b}})Du, Liu, Li, and
  Zhao}]{du-etal-2022-a-survey}
Du, Yifan, Zikang Liu, Junyi Li, and Wayne~Xin Zhao. 2022{\natexlab{b}}.
\newblock A survey of vision-language pre-trained models.
\newblock In \emph{Proceedings of the International Joint Conference on
  Artificial Intelligence}, pages 5436--5443.
\newblock Survey Track.

\bibitem[{Dufter, Schmitt, and Sch{\"u}tze(2022)}]{dufter-etal-2022-position}
Dufter, Philipp, Martin Schmitt, and Hinrich Sch{\"u}tze. 2022.
\newblock Position information in transformers: An overview.
\newblock \emph{Computational Linguistics}, 48(3):733--763.

\bibitem[{Dugan et~al.(2023)Dugan, Ippolito, Kirubarajan, Shi, and
  Callison-Burch}]{dugan-etal-2022-real}
Dugan, Liam, Daphne Ippolito, Arun Kirubarajan, Sherry Shi, and Chris
  Callison-Burch. 2023.
\newblock Real or fake text?: Investigating human ability to detect boundaries
  between human-written and machine-generated text.
\newblock In \emph{Proceedings of the AAAI Conference on Artificial
  Intelligence}, pages 12763--12771.

\bibitem[{Elazar et~al.(2022)Elazar, Kassner, Ravfogel, Feder, Ravichander,
  Mosbach, Belinkov, Sch{\"u}tze, and Goldberg}]{elazar-etal-2022-measuring}
Elazar, Yanai, Nora Kassner, Shauli Ravfogel, Amir Feder, Abhilasha
  Ravichander, Marius Mosbach, Yonatan Belinkov, Hinrich Sch{\"u}tze, and Yoav
  Goldberg. 2022.
\newblock Measuring causal effects of data statistics on language model's
  {`}factual{'} predictions.
\newblock \emph{ArXiv}, arXiv:2207.14251.

\bibitem[{Elazar et~al.(2021)Elazar, Kassner, Ravfogel, Ravichander, Hovy,
  Sch{\"u}tze, and Goldberg}]{elazar-etal-2021-measuring}
Elazar, Yanai, Nora Kassner, Shauli Ravfogel, Abhilasha Ravichander, Eduard
  Hovy, Hinrich Sch{\"u}tze, and Yoav Goldberg. 2021.
\newblock Measuring and improving consistency in pretrained language models.
\newblock \emph{Transactions of the Association for Computational Linguistics},
  9:1012--1031.

\bibitem[{Ettinger(2020)}]{ettinger-2019-what}
Ettinger, Allyson. 2020.
\newblock What {BERT} is not: Lessons from a new suite of psycholinguistic
  diagnostics for language models.
\newblock \emph{Transactions of the Association for Computational Linguistics},
  8:34--48.

\bibitem[{Fedus, Zoph, and Shazeer(2022)}]{fedus-etal-2021-switch}
Fedus, William, Barret Zoph, and Noam Shazeer. 2022.
\newblock Switch {Transformers}: Scaling to trillion parameter models with
  simple and efficient sparsity.
\newblock \emph{Journal of Machine Learning Research}, 23:1--39.

\bibitem[{Felkner et~al.(2022)Felkner, Chang, Jang, and
  May}]{felkner-etal-2022-towards}
Felkner, Virginia~K., Ho-Chun~Herbert Chang, Eugene Jang, and Jonathan May.
  2022.
\newblock Towards {WinoQueer}: Developing a benchmark for anti-queer bias in
  large language models.
\newblock In \emph{Queer in AI Workshop}.

\bibitem[{Finlayson et~al.(2021)Finlayson, Mueller, Gehrmann, Shieber, Linzen,
  and Belinkov}]{finlayson-etal-2021-causal}
Finlayson, Matthew, Aaron Mueller, Sebastian Gehrmann, Stuart Shieber, Tal
  Linzen, and Yonatan Belinkov. 2021.
\newblock Causal analysis of syntactic agreement mechanisms in neural language
  models.
\newblock In \emph{Proceedings of the 59th Annual Meeting of the Association
  for Computational Linguistics and the 11th International Joint Conference on
  Natural Language Processing (Volume 1: Long Papers)}, pages 1828--1843.

\bibitem[{Frank and Goodman(2012)}]{frank-goodman-2012-predicting}
Frank, Michael and Noah Goodman. 2012.
\newblock Predicting pragmatic reasoning in language games.
\newblock \emph{Science}, 336(6084):998.

\bibitem[{Freitag and Al-Onaizan(2017)}]{freitag-al-onaizan-2017-beam}
Freitag, Markus and Yaser Al-Onaizan. 2017.
\newblock Beam search strategies for neural machine translation.
\newblock In \emph{Proceedings of the First Workshop on Neural Machine
  Translation}, pages 56--60.

\bibitem[{Fricke(2018)}]{fricke-2018-semantic}
Fricke, Suzanne. 2018.
\newblock Semantic {S}cholar.
\newblock \emph{Journal of the Medical Library Association}, 106(1):145--147.

\bibitem[{Fujisawa and Kanai(2022)}]{fujisawa-kanai-2022-logical}
Fujisawa, Ippei and Ryota Kanai. 2022.
\newblock Logical tasks for measuring extrapolation and rule comprehension.
\newblock \emph{ArXiv}, arXiv:2211.07727.

\bibitem[{Ganguli et~al.(2022{\natexlab{a}})Ganguli, Hernandez, Lovitt, Askell,
  Bai, Chen, Conerly, Dassarma, Drain, Elhage, El~Showk, Fort, Hatfield-Dodds,
  Henighan, Johnston, Jones, Joseph, Kernian, Kravec, Mann, Nanda, Ndousse,
  Olsson, Amodei, Brown, Kaplan, McCandlish, Olah, Amodei, and
  Clark}]{ganguli-etal-2022-predictability}
Ganguli, Deep, Danny Hernandez, Liane Lovitt, Amanda Askell, Yuntao Bai, Anna
  Chen, Tom Conerly, Nova Dassarma, Dawn Drain, Nelson Elhage, Sheer El~Showk,
  Stanislav Fort, Zac Hatfield-Dodds, Tom Henighan, Scott Johnston, Andy Jones,
  Nicholas Joseph, Jackson Kernian, Shauna Kravec, Ben Mann, Neel Nanda, Kamal
  Ndousse, Catherine Olsson, Daniela Amodei, Tom Brown, Jared Kaplan, Sam
  McCandlish, Christopher Olah, Dario Amodei, and Jack Clark.
  2022{\natexlab{a}}.
\newblock Predictability and surprise in large generative models.
\newblock In \emph{Proceedings of the ACM Conference on Fairness,
  Accountability, and Transparency}, page 1747–1764, Association for
  Computing Machinery, New York, NY, USA.

\bibitem[{Ganguli et~al.(2022{\natexlab{b}})Ganguli, Lovitt, Kernion, Askell,
  Bai, Kadavath, Mann, Perez, Schiefer, Ndousse, Jones, Bowman, Chen, Conerly,
  DasSarma, Drain, Elhage, El-Showk, Fort, Dodds, Henighan, Hernandez, Hume,
  Jacobson, Johnston, Kravec, Olsson, Ringer, Tran-Johnson, Amodei, Brown,
  Joseph, McCandlish, Olah, Kaplan, and Clark}]{ganguli-etal-2022-red}
Ganguli, Deep, Liane Lovitt, John Kernion, Amanda Askell, Yuntao Bai, Saurav
  Kadavath, Benjamin Mann, Ethan Perez, Nicholas Schiefer, Kamal Ndousse, Andy
  Jones, Sam Bowman, Anna Chen, Tom Conerly, Nova DasSarma, Dawn Drain, Nelson
  Elhage, Sheer El-Showk, Stanislav Fort, Zachary Dodds, T.~J. Henighan, Danny
  Hernandez, Tristan Hume, Josh Jacobson, Scott Johnston, Shauna Kravec,
  Catherine Olsson, Sam Ringer, Eli Tran-Johnson, Dario Amodei, Tom~B. Brown,
  Nicholas Joseph, Sam McCandlish, Christopher Olah, Jared Kaplan, and Jack
  Clark. 2022{\natexlab{b}}.
\newblock Red teaming language models to reduce harms: Methods, scaling
  behaviors, and lessons learned.
\newblock \emph{ArXiv}, arXiv:2209.07858.

\bibitem[{Gauthier et~al.(2020)Gauthier, Hu, Wilcox, Qian, and
  Levy}]{gauthier-etal-2020-syntaxgym}
Gauthier, Jon, Jennifer Hu, Ethan Wilcox, Peng Qian, and Roger Levy. 2020.
\newblock {S}yntax{G}ym: An online platform for targeted evaluation of language
  models.
\newblock In \emph{Proceedings of the 58th Annual Meeting of the Association
  for Computational Linguistics: System Demonstrations}, pages 70--76.

\bibitem[{Geeraerts(2017)}]{geeraerts-2017-lexical}
Geeraerts, Dirk. 2017.
\newblock Lexical semantics.
\newblock \emph{Oxford Research Encyclopedia of Linguistics}.

\bibitem[{Gehman et~al.(2020)Gehman, Gururangan, Sap, Choi, and
  Smith}]{gehman-etal-2020-realtoxicityprompts}
Gehman, Samuel, Suchin Gururangan, Maarten Sap, Yejin Choi, and Noah~A. Smith.
  2020.
\newblock {R}eal{T}oxicity{P}rompts: Evaluating neural toxic degeneration in
  language models.
\newblock In \emph{Findings of the Association for Computational Linguistics:
  EMNLP 2020}, pages 3356--3369.

\bibitem[{Geiger, Richardson, and Potts(2020)}]{geiger-etal-2020-neural}
Geiger, Atticus, Kyle Richardson, and Christopher Potts. 2020.
\newblock Neural natural language inference models partially embed theories of
  lexical entailment and negation.
\newblock In \emph{Proceedings of the Third BlackboxNLP Workshop on Analyzing
  and Interpreting Neural Networks for NLP}, pages 163--173.

\bibitem[{Geva et~al.(2022)Geva, Caciularu, Wang, and
  Goldberg}]{geva-etal-2022-transformer}
Geva, Mor, Avi Caciularu, Kevin Wang, and Yoav Goldberg. 2022.
\newblock Transformer feed-forward layers build predictions by promoting
  concepts in the vocabulary space.
\newblock In \emph{Proceedings of the 2022 Conference on Empirical Methods in
  Natural Language Processing}, pages 30--45.

\bibitem[{Geva et~al.(2021)Geva, Schuster, Berant, and
  Levy}]{geva-etal-2020-transformer}
Geva, Mor, Roei Schuster, Jonathan Berant, and Omer Levy. 2021.
\newblock Transformer feed-forward layers are key-value memories.
\newblock In \emph{Proceedings of the 2021 Conference on Empirical Methods in
  Natural Language Processing}, pages 5484--5495.

\bibitem[{Goldberg(2019)}]{goldberg-2019-assessing}
Goldberg, Yoav. 2019.
\newblock Assessing {BERT}'s syntactic abilities.
\newblock \emph{ArXiv}, arXiv:1901.05287.

\bibitem[{Grice(1975)}]{grice-1975-logic}
Grice, H.~P. 1975.
\newblock Logic and conversation.
\newblock \emph{Syntax and Semantics: Vol. 3: Speech Acts}, pages 41--58.

\bibitem[{Grici{\=u}t{\.e}, Tanti, and Donatelli(2022)}]{griciute-etal-2022-on}
Grici{\=u}t{\.e}, Bernadeta, Marc Tanti, and Lucia Donatelli. 2022.
\newblock On the cusp of comprehensibility: Can language models distinguish
  between metaphors and nonsense?
\newblock In \emph{Proceedings of the 3rd Workshop on Figurative Language
  Processing (FLP)}, pages 173--177.

\bibitem[{Groenwold et~al.(2020)Groenwold, Ou, Parekh, Honnavalli, Levy, Mirza,
  and Wang}]{groenwold-etal-2020-investigating}
Groenwold, Sophie, Lily Ou, Aesha Parekh, Samhita Honnavalli, Sharon Levy, Diba
  Mirza, and William~Yang Wang. 2020.
\newblock Investigating {A}frican-{A}merican {V}ernacular {E}nglish in
  transformer-based text generation.
\newblock In \emph{Proceedings of the 2020 Conference on Empirical Methods in
  Natural Language Processing (EMNLP)}, pages 5877--5883.

\bibitem[{Gubelmann and Handschuh(2022)}]{gubelmann-handschuh-2022-context}
Gubelmann, Reto and Siegfried Handschuh. 2022.
\newblock Context matters: A pragmatic study of {PLM}s{'} negation
  understanding.
\newblock In \emph{Proceedings of the 60th Annual Meeting of the Association
  for Computational Linguistics (Volume 1: Long Papers)}, pages 4602--4621.

\bibitem[{Guu et~al.(2020)Guu, Lee, Tung, Pasupat, and
  Chang}]{guu-etal-2020-retrieval}
Guu, Kelvin, Kenton Lee, Zora Tung, Panupong Pasupat, and Mingwei Chang. 2020.
\newblock Retrieval augmented language model pre-training.
\newblock In \emph{International Conference on Machine Learning}, pages
  3929--3938.

\bibitem[{Hagendorff, Fabi, and Kosinski(2022)}]{hagendorff-etal-2022-machine}
Hagendorff, Thilo, Sarah Fabi, and Michal Kosinski. 2022.
\newblock Machine intuition: Uncovering human-like intuitive decision-making in
  {GPT-3.5}.
\newblock \emph{ArXiv}, arXiv:2212.05206.

\bibitem[{Hahn(2020)}]{hahn-2019-theoretical}
Hahn, Michael. 2020.
\newblock Theoretical limitations of self-attention in neural sequence models.
\newblock \emph{Transactions of the Association for Computational Linguistics},
  8:156--171.

\bibitem[{Han et~al.(2022)Han, Schoelkopf, Zhao, Qi, Riddell, Benson, Sun,
  Zubova, Qiao, Burtell, Peng, Fan, Liu, Wong, Sailor, Ni, Nan, Kasai, Yu,
  Zhang, Joty, Fabbri, Kryscinski, Lin, Xiong, and Radev}]{han-etal-2022-folio}
Han, Simeng, Hailey Schoelkopf, Yilun Zhao, Zhenting Qi, Martin Riddell, Luke
  Benson, Lucy Sun, Ekaterina Zubova, Yujie Qiao, Matthew Burtell, David Peng,
  Jonathan Fan, Yixin Liu, Brian Wong, Malcolm Sailor, Ansong Ni, Linyong Nan,
  Jungo Kasai, Tao Yu, Rui Zhang, Shafiq Joty, Alexander~R. Fabbri, Wojciech
  Kryscinski, Xi~Victoria Lin, Caiming Xiong, and Dragomir Radev. 2022.
\newblock {FOLIO}: Natural language reasoning with first-order logic.
\newblock \emph{ArXiv}, arXiv:2209.00840.

\bibitem[{Hanna and Mare{\v{c}}ek(2021)}]{hanna-marecek-2021-analyzing}
Hanna, Michael and David Mare{\v{c}}ek. 2021.
\newblock Analyzing {BERT}{'}s knowledge of hypernymy via prompting.
\newblock In \emph{Proceedings of the Fourth BlackboxNLP Workshop on Analyzing
  and Interpreting Neural Networks for NLP}, pages 275--282.

\bibitem[{Hassan, Huenerfauth, and Alm(2021)}]{hassan-etal-2021-unpacking}
Hassan, Saad, Matt Huenerfauth, and Cecilia~Ovesdotter Alm. 2021.
\newblock Unpacking the interdependent systems of discrimination: Ableist bias
  in {NLP} systems through an intersectional lens.
\newblock In \emph{Findings of the Association for Computational Linguistics:
  EMNLP 2021}, pages 3116--3123.

\bibitem[{Haviv et~al.(2022)Haviv, Ram, Press, Izsak, and
  Levy}]{haviv-etal-2022-transformer}
Haviv, Adi, Ori Ram, Ofir Press, Peter Izsak, and Omer Levy. 2022.
\newblock Transformer language models without positional encodings still learn
  positional information.
\newblock In \emph{Findings of the Association for Computational Linguistics:
  EMNLP 2022}, pages 1382--1390.

\bibitem[{Hawkins et~al.(2020)Hawkins, Yamakoshi, Griffiths, and
  Goldberg}]{hawkins-etal-2020-investigating}
Hawkins, Robert, Takateru Yamakoshi, Thomas Griffiths, and Adele Goldberg.
  2020.
\newblock Investigating representations of verb bias in neural language models.
\newblock In \emph{Proceedings of the 2020 Conference on Empirical Methods in
  Natural Language Processing (EMNLP)}, pages 4653--4663.

\bibitem[{He et~al.(2022{\natexlab{a}})He, Cheng, Li, Xie, and
  Xiao}]{he-etal-2022-can}
He, Qianyu, Sijie Cheng, Zhixu Li, Rui Xie, and Yanghua Xiao.
  2022{\natexlab{a}}.
\newblock Can pre-trained language models interpret similes as smart as human?
\newblock In \emph{Proceedings of the 60th Annual Meeting of the Association
  for Computational Linguistics (Volume 1: Long Papers)}, pages 7875--7887.

\bibitem[{He et~al.(2022{\natexlab{b}})He, Xu, Lyu, Wu, and
  Wang}]{he-etal-2021-protecting}
He, Xuanli, Qiongkai Xu, Lingjuan Lyu, Fangzhao Wu, and Chenguang Wang.
  2022{\natexlab{b}}.
\newblock Protecting intellectual property of language generation {API}s with
  lexical watermark.
\newblock In \emph{Proceedings of the AAAI Conference on Artificial
  Intelligence}, pages 10758--10766.

\bibitem[{Heidenreich and Williams(2021)}]{heidenreich-williams-2021-the}
Heidenreich, Hunter~Scott and Jake~Ryland Williams. 2021.
\newblock The {Earth} is flat and the {Sun} is not a star: The susceptibility
  of {GPT-2} to universal adversarial triggers.
\newblock In \emph{Proceedings of the 2021 AAAI/ACM Conference on AI, Ethics,
  and Society}, page 566–573, Association for Computing Machinery, New York,
  NY, USA.

\bibitem[{Hendrycks et~al.(2021{\natexlab{a}})Hendrycks, Burns, Basart, Zou,
  Mazeika, Song, and Steinhardt}]{hendrycks-etal-2020-measuring}
Hendrycks, Dan, Collin Burns, Steven Basart, Andy Zou, Mantas Mazeika, Dawn
  Song, and Jacob Steinhardt. 2021{\natexlab{a}}.
\newblock Measuring massive multitask language understanding.
\newblock In \emph{International Conference on Learning Representations}.

\bibitem[{Hendrycks et~al.(2021{\natexlab{b}})Hendrycks, Burns, Kadavath,
  Arora, Basart, Tang, Song, and
  Steinhardt}]{hendrycks-etal-2021-measuring-mathematical}
Hendrycks, Dan, Collin Burns, Saurav Kadavath, Akul Arora, Steven Basart, Eric
  Tang, Dawn Song, and Jacob Steinhardt. 2021{\natexlab{b}}.
\newblock Measuring mathematical problem solving with the {MATH} dataset.
\newblock In \emph{Advances in Neural Information Processing Systems Datasets
  and Benchmarks Track}.

\bibitem[{Hernandez et~al.(2022)Hernandez, Brown, Conerly, DasSarma, Drain,
  El-Showk, Elhage, Hatfield-Dodds, Henighan, Hume, Johnston, Mann, Olah,
  Olsson, Amodei, Joseph, Kaplan, and McCandlish}]{hernandez-etal-2022-scaling}
Hernandez, Danny, Tom~B. Brown, Tom Conerly, Nova DasSarma, Dawn Drain, Sheer
  El-Showk, Nelson Elhage, Zac Hatfield-Dodds, Tom Henighan, Tristan Hume,
  Scott Johnston, Benjamin Mann, Christopher Olah, Catherine Olsson, Dario
  Amodei, Nicholas Joseph, Jared Kaplan, and Sam McCandlish. 2022.
\newblock Scaling laws and interpretability of learning from repeated data.
\newblock \emph{ArXiv}, arXiv:2205.10487.

\bibitem[{Hershcovich et~al.(2022)Hershcovich, Frank, Lent, de~Lhoneux, Abdou,
  Brandl, Bugliarello, Cabello~Piqueras, Chalkidis, Cui, Fierro, Margatina,
  Rust, and S{\o}gaard}]{hershcovich-etal-2022-challenges}
Hershcovich, Daniel, Stella Frank, Heather Lent, Miryam de~Lhoneux, Mostafa
  Abdou, Stephanie Brandl, Emanuele Bugliarello, Laura Cabello~Piqueras, Ilias
  Chalkidis, Ruixiang Cui, Constanza Fierro, Katerina Margatina, Phillip Rust,
  and Anders S{\o}gaard. 2022.
\newblock Challenges and strategies in cross-cultural {NLP}.
\newblock In \emph{Proceedings of the 60th Annual Meeting of the Association
  for Computational Linguistics (Volume 1: Long Papers)}, pages 6997--7013.

\bibitem[{Hoffmann et~al.(2022)Hoffmann, Borgeaud, Mensch, Buchatskaya, Cai,
  Rutherford, de~las Casas, Hendricks, Welbl, Clark, Hennigan, Noland,
  Millican, van~den Driessche, Damoc, Guy, Osindero, Simonyan, Elsen, Vinyals,
  Rae, and Sifre}]{hoffmann-etal-2022-training}
Hoffmann, Jordan, Sebastian Borgeaud, Arthur Mensch, Elena Buchatskaya, Trevor
  Cai, Eliza Rutherford, Diego de~las Casas, Lisa~Anne Hendricks, Johannes
  Welbl, Aidan Clark, Tom Hennigan, Eric Noland, Katherine Millican, George
  van~den Driessche, Bogdan Damoc, Aurelia Guy, Simon Osindero, Karen Simonyan,
  Erich Elsen, Oriol Vinyals, Jack~William Rae, and Laurent Sifre. 2022.
\newblock Training compute-optimal large language models.
\newblock In \emph{Advances in Neural Information Processing Systems},
  volume~35, pages 30016--30030.

\bibitem[{Holtzman et~al.(2020)Holtzman, Buys, Du, Forbes, and
  Choi}]{holtzman-etal-2019-curious}
Holtzman, Ari, Jan Buys, Li~Du, Maxwell Forbes, and Yejin Choi. 2020.
\newblock The curious case of neural text degeneration.
\newblock In \emph{International Conference on Learning Representations}.

\bibitem[{Hossain, Chinnappa, and Blanco(2022)}]{hossain-etal-2022-an}
Hossain, Md~Mosharaf, Dhivya Chinnappa, and Eduardo Blanco. 2022.
\newblock An analysis of negation in natural language understanding corpora.
\newblock In \emph{Proceedings of the 60th Annual Meeting of the Association
  for Computational Linguistics (Volume 2: Short Papers)}, pages 716--723.

\bibitem[{Hossain et~al.(2020)Hossain, Kovatchev, Dutta, Kao, Wei, and
  Blanco}]{hossain-etal-2020-an}
Hossain, Md~Mosharaf, Venelin Kovatchev, Pranoy Dutta, Tiffany Kao, Elizabeth
  Wei, and Eduardo Blanco. 2020.
\newblock An analysis of natural language inference benchmarks through the lens
  of negation.
\newblock In \emph{Proceedings of the 2020 Conference on Empirical Methods in
  Natural Language Processing (EMNLP)}, pages 9106--9118.

\bibitem[{Hosseini et~al.(2022)Hosseini, Vani, Bahdanau, Sordoni, and
  Courville}]{hosseini-etal-2022-on}
Hosseini, Arian, Ankit Vani, Dzmitry Bahdanau, Alessandro Sordoni, and Aaron
  Courville. 2022.
\newblock On the compositional generalization gap of in-context learning.
\newblock In \emph{Proceedings of the Fifth BlackboxNLP Workshop on Analyzing
  and Interpreting Neural Networks for NLP}, pages 272--280.

\bibitem[{Hu, Chen, and Levy(2020)}]{hu-etal-2020-a-closer}
Hu, Jennifer, Sherry~Yong Chen, and Roger Levy. 2020.
\newblock A closer look at the performance of neural language models on
  reflexive anaphor licensing.
\newblock In \emph{Proceedings of the Society for Computation in Linguistics
  2020}, pages 323--333.

\bibitem[{Hu et~al.(2022)Hu, Floyd, Jouravlev, Fedorenko, and
  Gibson}]{hu-etal-2022-a}
Hu, Jennifer, Sammy Floyd, Olessia Jouravlev, Evelina Fedorenko, and Edward
  Gibson. 2022.
\newblock A fine-grained comparison of pragmatic language understanding in
  humans and language models.
\newblock \emph{ArXiv}, arXiv:2212.06801.

\bibitem[{Hu et~al.(2020)Hu, Gauthier, Qian, Wilcox, and
  Levy}]{hu-etal-2020-a-systematic}
Hu, Jennifer, Jon Gauthier, Peng Qian, Ethan Wilcox, and Roger Levy. 2020.
\newblock A systematic assessment of syntactic generalization in neural
  language models.
\newblock In \emph{Proceedings of the 58th Annual Meeting of the Association
  for Computational Linguistics}, pages 1725--1744.

\bibitem[{Huang, Shao, and Chang(2022)}]{huang-etal-2022-are}
Huang, Jie, Hanyin Shao, and Kevin Chen-Chuan Chang. 2022.
\newblock Are large pre-trained language models leaking your personal
  information?
\newblock In \emph{Findings of the Association for Computational Linguistics:
  EMNLP 2022}, pages 2038--2047.

\bibitem[{Huebner et~al.(2021)Huebner, Sulem, Cynthia, and
  Roth}]{huebner-etal-2021-babyberta}
Huebner, Philip~A., Elior Sulem, Fisher Cynthia, and Dan Roth. 2021.
\newblock {B}aby{BERT}a: Learning more grammar with small-scale child-directed
  language.
\newblock In \emph{Proceedings of the 25th Conference on Computational Natural
  Language Learning}, pages 624--646.

\bibitem[{Hupkes et~al.(2022)Hupkes, Giulianelli, Dankers, Artetxe, Elazar,
  Pimentel, Christodoulopoulos, Lasri, Saphra, Sinclair, Ulmer, Schottmann,
  Batsuren, Sun, Sinha, Khalatbari, Ryskina, Frieske, Cotterell, and
  Jin}]{hupkes-etal-2022-state}
Hupkes, Dieuwke, Mario Giulianelli, Verna Dankers, Mikel Artetxe, Yanai Elazar,
  Tiago Pimentel, Christos Christodoulopoulos, Karim Lasri, Naomi Saphra,
  Arabella Sinclair, Dennis Ulmer, Florian Schottmann, Khuyagbaatar Batsuren,
  Kaiser Sun, Koustuv Sinha, Leila Khalatbari, Maria Ryskina, Rita Frieske,
  Ryan Cotterell, and Zhijing Jin. 2022.
\newblock State-of-the-art generalisation research in {NLP}: A taxonomy and
  review.
\newblock \emph{ArXiv}, arXiv:2210.03050.

\bibitem[{Huynh, Lentz, and van Miltenburg(2022)}]{huynh-etal-2022-implicit}
Huynh, Hien, Tomas~O. Lentz, and Emiel van Miltenburg. 2022.
\newblock Implicit causality in {GPT-2}: A case study.
\newblock \emph{ArXiv}, arXiv:2212.04348.

\bibitem[{Ippolito et~al.(2020)Ippolito, Duckworth, Callison-Burch, and
  Eck}]{ippolito-etal-2019-automatic}
Ippolito, Daphne, Daniel Duckworth, Chris Callison-Burch, and Douglas Eck.
  2020.
\newblock Automatic detection of generated text is easiest when humans are
  fooled.
\newblock In \emph{Proceedings of the 58th Annual Meeting of the Association
  for Computational Linguistics}, pages 1808--1822.

\bibitem[{Ippolito et~al.(2022)Ippolito, Tram\`{e}r, Nasr, Zhang, Jagielski,
  Lee, Choquette-Choo, and Carlini}]{ippolito-etal-2022-preventing}
Ippolito, Daphne, Florian Tram\`{e}r, Milad Nasr, Chiyuan Zhang, Matthew
  Jagielski, Katherine Lee, Christopher~A. Choquette-Choo, and Nicholas
  Carlini. 2022.
\newblock Preventing verbatim memorization in language models gives a false
  sense of privacy.
\newblock \emph{ArXiv}, arXiv:2210.17546.

\bibitem[{Iyer et~al.(2022)Iyer, Lin, Pasunuru, Mihaylov, Simig, Yu, Shuster,
  Wang, Liu, Koura, Li, O'Horo, Pereyra, Wang, Dewan, Celikyilmaz, Zettlemoyer,
  and Stoyanov}]{iyer-etal-2022-optiml}
Iyer, Srinivas, Xiaojuan Lin, Ramakanth Pasunuru, Todor Mihaylov, Daniel Simig,
  Ping Yu, Kurt Shuster, Tianlu Wang, Qing Liu, Punit~Singh Koura, Xian Li,
  Brian O'Horo, Gabriel Pereyra, Jeff Wang, Christopher Dewan, Asli
  Celikyilmaz, Luke Zettlemoyer, and Veselin Stoyanov. 2022.
\newblock {OPT-IML}: Scaling language model instruction meta learning through
  the lens of generalization.
\newblock \emph{ArXiv}, arXiv:2212.12017.

\bibitem[{Jain and Wallace(2019)}]{jain-wallace-2019-attention}
Jain, Sarthak and Byron~C. Wallace. 2019.
\newblock {A}ttention is not {E}xplanation.
\newblock In \emph{Proceedings of the 2019 Conference of the North {A}merican
  Chapter of the Association for Computational Linguistics: Human Language
  Technologies, Volume 1 (Long and Short Papers)}, pages 3543--3556.

\bibitem[{Jakesch, Hancock, and Naaman(2023)}]{jakesch-etal-2022-human}
Jakesch, Maurice, Jeffrey~T. Hancock, and Mor Naaman. 2023.
\newblock Human heuristics for {AI}-generated language are flawed.
\newblock \emph{Proceedings of the National Academy of Sciences},
  120(11):e2208839120.

\bibitem[{Jang, Ye, and Seo(2022)}]{jang-etal-2022-can}
Jang, Joel, Seonghyeon Ye, and Minjoon Seo. 2022.
\newblock Can large language models truly understand prompts? a case study with
  negated prompts.
\newblock In \emph{Proceedings of the 1st Transfer Learning for Natural
  Language Processing Workshop}, pages 52--62.

\bibitem[{Jawahar, Abdul-Mageed, and
  Lakshmanan(2020)}]{jawahar-etal-2020-automatic}
Jawahar, Ganesh, Muhammad Abdul-Mageed, and Laks Lakshmanan, V.S. 2020.
\newblock Automatic detection of machine generated text: A critical survey.
\newblock In \emph{Proceedings of the 28th International Conference on
  Computational Linguistics}, pages 2296--2309, International Committee on
  Computational Linguistics, Barcelona, Spain (Online).

\bibitem[{Jiang et~al.(2020{\natexlab{a}})Jiang, Nian, Guo, Chu, Zhao, Shen,
  and Tu}]{jiang-etal-2019-learning}
Jiang, Chengyue, Zhonglin Nian, Kaihao Guo, Shanbo Chu, Yinggong Zhao, Libin
  Shen, and Kewei Tu. 2020{\natexlab{a}}.
\newblock Learning numeral embedding.
\newblock In \emph{Findings of the Association for Computational Linguistics:
  EMNLP 2020}, pages 2586--2599.

\bibitem[{Jiang et~al.(2022)Jiang, Xu, Zhu, Han, Zhang, and
  Zhu}]{jiang-etal-2022-mpi}
Jiang, Guangyuan, Manjie Xu, Song-Chun Zhu, Wenjuan Han, Chi Zhang, and Yixin
  Zhu. 2022.
\newblock {MPI}: Evaluating and inducing personality in pre-trained language
  models.
\newblock \emph{ArXiv}, arXiv:2206.07550.

\bibitem[{Jiang and Riloff(2021)}]{jiang-riloff-2021-learning}
Jiang, Tianyu and Ellen Riloff. 2021.
\newblock Learning prototypical functions for physical artifacts.
\newblock In \emph{Proceedings of the 59th Annual Meeting of the Association
  for Computational Linguistics and the 11th International Joint Conference on
  Natural Language Processing (Volume 1: Long Papers)}, pages 6941--6951.

\bibitem[{Jiang et~al.(2020{\natexlab{b}})Jiang, Xu, Araki, and
  Neubig}]{jiang-etal-2019-how}
Jiang, Zhengbao, Frank~F. Xu, Jun Araki, and Graham Neubig. 2020{\natexlab{b}}.
\newblock How can we know what language models know?
\newblock \emph{Transactions of the Association for Computational Linguistics},
  8:423--438.

\bibitem[{Jigsaw(2017)}]{jigsaw-2017-perspective}
Jigsaw. 2017.
\newblock {Perspective API}.
\newblock \emph{Google Jigsaw}.

\bibitem[{Jin et~al.(2022{\natexlab{a}})Jin, Levine, Adauto, Kamal, Sap,
  Sachan, Mihalcea, Tenenbaum, and Sch{"o}lkopf}]{jin-etal-2022-when}
Jin, Zhijing, Sydney Levine, Fernando~Gonzalez Adauto, Ojasv Kamal, Maarten
  Sap, Mrinmaya Sachan, Rada Mihalcea, Joshua~B. Tenenbaum, and Bernhard
  Sch{"o}lkopf. 2022{\natexlab{a}}.
\newblock When to make exceptions: Exploring language models as accounts of
  human moral judgment.
\newblock In \emph{Advances in Neural Information Processing Systems},
  volume~35, pages 28458--28473.

\bibitem[{Jin et~al.(2022{\natexlab{b}})Jin, Zhang, Yu, and
  Huang}]{jin-etal-2022-probing}
Jin, Zijia, Xingyu Zhang, Mo~Yu, and Lifu Huang. 2022{\natexlab{b}}.
\newblock Probing script knowledge from pre-trained models.
\newblock In \emph{Proceedings of the Workshop on Unimodal and Multimodal
  Induction of Linguistic Structures (UM-IoS)}, pages 87--93.

\bibitem[{Johnson et~al.(2022)Johnson, Pistilli, Men{\'e}dez-Gonz{\'a}lez,
  Duran, Panai, Kalpokiene, and Bertulfo}]{johnson-etal-2022-the}
Johnson, Rebecca~L, Giada Pistilli, Natalia Men{\'e}dez-Gonz{\'a}lez, Leslye
  Denisse~Dias Duran, Enrico Panai, Julija Kalpokiene, and Donald~Jay Bertulfo.
  2022.
\newblock The ghost in the machine has an {American} accent: Value conflict in
  {GPT-3}.
\newblock \emph{ArXiv}, arXiv:2203.07785.

\bibitem[{Johnson(2022)}]{johnson-2022-ai}
Johnson, Steven. 2022.
\newblock {A.I.} is mastering language. {S}hould we trust what it says?
\newblock \emph{The New York Times}.

\bibitem[{Jones and Bergen(2021)}]{jones-bergen-2021-the}
Jones, Cameron and Benjamin Bergen. 2021.
\newblock The role of physical inference in pronoun resolution.
\newblock In \emph{Proceedings of the Annual Meeting of the Cognitive Science
  Society}, volume~43, pages 2876--2882.

\bibitem[{Jones et~al.(2022)Jones, Chang, Coulson, Michaelov, Trott, and
  Bergen}]{jones-etal-2022-distributional}
Jones, Cameron~R., Tyler~A. Chang, Seana Coulson, James Michaelov, Sean Trott,
  and Benjamin Bergen. 2022.
\newblock Distributional semantics still can’t account for affordances.
\newblock In \emph{Proceedings of the Annual Meeting of the Cognitive Science
  Society}, volume~44, pages 482--489.

\bibitem[{Joshi et~al.(2020)Joshi, Santy, Budhiraja, Bali, and
  Choudhury}]{joshi-etal-2020-state}
Joshi, Pratik, Sebastin Santy, Amar Budhiraja, Kalika Bali, and Monojit
  Choudhury. 2020.
\newblock The state and fate of linguistic diversity and inclusion in the {NLP}
  world.
\newblock In \emph{Proceedings of the 58th Annual Meeting of the Association
  for Computational Linguistics}, pages 6282--6293.

\bibitem[{Kabbara and Cheung(2022)}]{kabbara-cheung-2022-investigating}
Kabbara, Jad and Jackie Chi~Kit Cheung. 2022.
\newblock Investigating the performance of transformer-based {NLI} models on
  presuppositional inferences.
\newblock In \emph{Proceedings of the 29th International Conference on
  Computational Linguistics}, pages 779--785, International Committee on
  Computational Linguistics, Gyeongju, Republic of Korea.

\bibitem[{Kadavath et~al.(2022)Kadavath, Conerly, Askell, Henighan, Drain,
  Perez, Schiefer, Dodds, DasSarma, Tran-Johnson, Johnston, El-Showk, Jones,
  Elhage, Hume, Chen, Bai, Bowman, Fort, Ganguli, Hernandez, Jacobson, Kernion,
  Kravec, Lovitt, Ndousse, Olsson, Ringer, Amodei, Brown, Clark, Joseph, Mann,
  McCandlish, Olah, and Kaplan}]{kadavath-etal-2022-language}
Kadavath, Saurav, Tom Conerly, Amanda Askell, Tom Henighan, Dawn Drain, Ethan
  Perez, Nicholas Schiefer, Zachary Dodds, Nova DasSarma, Eli Tran-Johnson,
  Scott Johnston, Sheer El-Showk, Andy Jones, Nelson Elhage, Tristan Hume, Anna
  Chen, Yuntao Bai, Sam Bowman, Stanislav Fort, Deep Ganguli, Danny Hernandez,
  Josh Jacobson, John Kernion, Shauna Kravec, Liane Lovitt, Kamal Ndousse,
  Catherine Olsson, Sam Ringer, Dario Amodei, Tom~B. Brown, Jack Clark,
  Nicholas Joseph, Benjamin Mann, Sam McCandlish, Christopher Olah, and Jared
  Kaplan. 2022.
\newblock Language models (mostly) know what they know.
\newblock \emph{ArXiv}, arXiv:2207.05221.

\bibitem[{Kalo and Fichtel(2022)}]{kalo-2022-kamel}
Kalo, Jan-Christoph and Leandra Fichtel. 2022.
\newblock {KAMEL}: Knowledge analysis with multitoken entities in language
  models.
\newblock In \emph{4th Conference on Automated Knowledge Base Construction}.

\bibitem[{Kandpal et~al.(2022)Kandpal, Deng, Roberts, Wallace, and
  Raffel}]{kandpal-etal-2022-large}
Kandpal, Nikhil, Haikang Deng, Adam Roberts, Eric Wallace, and Colin Raffel.
  2022.
\newblock Large language models struggle to learn long-tail knowledge.
\newblock \emph{ArXiv}, arXiv:2211.08411.

\bibitem[{Kandpal, Wallace, and Raffel(2022)}]{kandpal-etal-2022-deduplicating}
Kandpal, Nikhil, Eric Wallace, and Colin Raffel. 2022.
\newblock Deduplicating training data mitigates privacy risks in language
  models.
\newblock In \emph{International Conference on Machine Learning}, pages
  10697--10707.

\bibitem[{Kaplan et~al.(2020)Kaplan, McCandlish, Henighan, Brown, Chess, Child,
  Gray, Radford, Wu, and Amodei}]{kaplan-etal-2020-scaling}
Kaplan, Jared, Sam McCandlish, Tom Henighan, Tom~B. Brown, Benjamin Chess,
  Rewon Child, Scott Gray, Alec Radford, Jeff Wu, and Dario Amodei. 2020.
\newblock Scaling laws for neural language models.
\newblock \emph{ArXiv}, arXiv:2001.08361.

\bibitem[{Karpas et~al.(2022)Karpas, Abend, Belinkov, Lenz, Lieber, Ratner,
  Shoham, Bata, Levine, Leyton-Brown, Muhlgay, Rozen, Schwartz, Shachaf,
  Shalev-Shwartz, Shashua, and Tenenholtz}]{karpas-etal-2022-mrkl}
Karpas, Ehud, Omri Abend, Yonatan Belinkov, Barak Lenz, Opher Lieber, Nir
  Ratner, Yoav Shoham, Hofit Bata, Yoav Levine, Kevin Leyton-Brown, Dor
  Muhlgay, Noam Rozen, Erez Schwartz, Gal Shachaf, Shai Shalev-Shwartz, Amnon
  Shashua, and Moshe Tenenholtz. 2022.
\newblock {MRKL} {S}ystems: {A} modular, neuro-symbolic architecture that
  combines large language models, external knowledge sources and discrete
  reasoning.
\newblock \emph{ArXiv}, arXiv:2205.00445.

\bibitem[{Kassner, Krojer, and Sch{\"u}tze(2020)}]{kassner-etal-2020-are}
Kassner, Nora, Benno Krojer, and Hinrich Sch{\"u}tze. 2020.
\newblock Are pretrained language models symbolic reasoners over knowledge?
\newblock In \emph{Proceedings of the 24th Conference on Computational Natural
  Language Learning}, pages 552--564.

\bibitem[{Kassner and Sch{\"u}tze(2020)}]{kassner-schutze-2020-negated}
Kassner, Nora and Hinrich Sch{\"u}tze. 2020.
\newblock Negated and misprimed probes for pretrained language models: Birds
  can talk, but cannot fly.
\newblock In \emph{Proceedings of the 58th Annual Meeting of the Association
  for Computational Linguistics}, pages 7811--7818.

\bibitem[{Katz, Geva, and Berant(2022)}]{katz-etal-2022-inferring}
Katz, Uri, Mor Geva, and Jonathan Berant. 2022.
\newblock Inferring implicit relations in complex questions with language
  models.
\newblock In \emph{Findings of the Association for Computational Linguistics:
  EMNLP 2022}, pages 2548--2566.

\bibitem[{Kauf et~al.(2022)Kauf, Ivanova, Rambelli, Chersoni, She, Chowdhury,
  Fedorenko, and Lenci}]{kauf-etal-2022-event}
Kauf, Carina, Anna~A. Ivanova, Giulia Rambelli, Emmanuele Chersoni, Jingyuan~S.
  She, Zawad Chowdhury, Evelina Fedorenko, and Alessandro Lenci. 2022.
\newblock Event knowledge in large language models: The gap between the
  impossible and the unlikely.
\newblock \emph{ArXiv}, arXiv:2212.01488.

\bibitem[{Kavumba et~al.(2020)Kavumba, Inoue, Heinzerling, Singh, Reisert, and
  Inui}]{kavumba-etal-2020-balanced}
Kavumba, Pride, Naoya Inoue, Benjamin Heinzerling, Keshav Singh, Paul Reisert,
  and Kentarou Inui. 2020.
\newblock Balanced {COPA}: Countering superficial cues in causal reasoning.
\newblock \emph{Association for Natural Language Processing}, pages 1105--1108.

\bibitem[{Kavumba, Takahashi, and Oda(2022)}]{kavumba-etal-2022-are}
Kavumba, Pride, Ryo Takahashi, and Yusuke Oda. 2022.
\newblock Are prompt-based models clueless?
\newblock In \emph{Proceedings of the 60th Annual Meeting of the Association
  for Computational Linguistics (Volume 1: Long Papers)}, pages 2333--2352.

\bibitem[{Kementchedjhieva, Anderson, and
  S{\o}gaard(2021)}]{kementchedjhieva-etal-2021-john}
Kementchedjhieva, Yova, Mark Anderson, and Anders S{\o}gaard. 2021.
\newblock John praised {M}ary because {\_}he{\_}? implicit causality bias and
  its interaction with explicit cues in {LM}s.
\newblock In \emph{Findings of the Association for Computational Linguistics:
  ACL-IJCNLP 2021}, pages 4859--4871.

\bibitem[{Khandelwal et~al.(2020)Khandelwal, Levy, Jurafsky, Zettlemoyer, and
  Lewis}]{khandelwal-etal-2020-generalization}
Khandelwal, Urvashi, Omer Levy, Dan Jurafsky, Luke Zettlemoyer, and Mike Lewis.
  2020.
\newblock Generalization through memorization: {N}earest neighbor language
  models.
\newblock In \emph{International Conference on Learning Representations}.

\bibitem[{Kharitonov, Baroni, and Hupkes(2021)}]{kharitonov-etal-2021-how}
Kharitonov, Eugene, Marco Baroni, and Dieuwke Hupkes. 2021.
\newblock How {BPE} affects memorization in {T}ransformers.
\newblock \emph{ArXiv}, arXiv:2110.02782.

\bibitem[{Kim, Yu, and Ettinger(2022)}]{kim-etal-2022-no}
Kim, Sanghee~J., Lang Yu, and Allyson Ettinger. 2022.
\newblock {``}no, they did not{''}: Dialogue response dynamics in pre-trained
  language models.
\newblock In \emph{Proceedings of the 29th International Conference on
  Computational Linguistics}, pages 863--874, International Committee on
  Computational Linguistics, Gyeongju, Republic of Korea.

\bibitem[{Kirchenbauer et~al.(2023)Kirchenbauer, Geiping, Wen, Katz, Miers, and
  Goldstein}]{kirchenbauer-etal-2023-a}
Kirchenbauer, John, Jonas Geiping, Yuxin Wen, Jonathan Katz, Ian Miers, and Tom
  Goldstein. 2023.
\newblock A watermark for large language models.
\newblock \emph{ArXiv}, arXiv:2301.10226.

\bibitem[{Kirk et~al.(2021)Kirk, Jun, Volpin, Iqbal, Benussi, Dreyer,
  Shtedritski, and Asano}]{kirk-etal-2021-bias}
Kirk, Hannah~Rose, Yennie Jun, Filippo Volpin, Haider Iqbal, Elias Benussi,
  Frederic Dreyer, Aleksandar Shtedritski, and Yuki Asano. 2021.
\newblock Bias out-of-the-box: An empirical analysis of intersectional
  occupational biases in popular generative language models.
\newblock In \emph{Advances in Neural Information Processing Systems},
  volume~34, pages 2611--2624.

\bibitem[{Ko and Li(2020)}]{ko-li-2020-assessing}
Ko, Wei-Jen and Junyi~Jessy Li. 2020.
\newblock Assessing discourse relations in language generation from {GPT}-2.
\newblock In \emph{Proceedings of the 13th International Conference on Natural
  Language Generation}, pages 52--59.

\bibitem[{Kojima et~al.(2022)Kojima, Gu, Reid, Matsuo, and
  Iwasawa}]{kojima-etal-2022-large}
Kojima, Takeshi, Shixiang~Shane Gu, Machel Reid, Yutaka Matsuo, and Yusuke
  Iwasawa. 2022.
\newblock Large language models are zero-shot reasoners.
\newblock In \emph{Advances in Neural Information Processing Systems},
  volume~35, pages 22199--22213.

\bibitem[{Kovaleva et~al.(2019)Kovaleva, Romanov, Rogers, and
  Rumshisky}]{kovaleva-etal-2019-revealing}
Kovaleva, Olga, Alexey Romanov, Anna Rogers, and Anna Rumshisky. 2019.
\newblock Revealing the dark secrets of {BERT}.
\newblock In \emph{Proceedings of the 2019 Conference on Empirical Methods in
  Natural Language Processing and the 9th International Joint Conference on
  Natural Language Processing (EMNLP-IJCNLP)}, pages 4365--4374.

\bibitem[{Krawczyk and Subramanya(2023)}]{krawczyk-subramanya-2023-bard}
Krawczyk, Jack and Amarnag Subramanya. 2023.
\newblock Bard is getting better at logic and reasoning.
\newblock \emph{The Keyword: Google Blog}.

\bibitem[{Kudo(2018)}]{kudo-2018-subword}
Kudo, Taku. 2018.
\newblock Subword regularization: Improving neural network translation models
  with multiple subword candidates.
\newblock In \emph{Proceedings of the 56th Annual Meeting of the Association
  for Computational Linguistics (Volume 1: Long Papers)}, pages 66--75.

\bibitem[{Kurita et~al.(2019)Kurita, Vyas, Pareek, Black, and
  Tsvetkov}]{kurita-etal-2019-measuring}
Kurita, Keita, Nidhi Vyas, Ayush Pareek, Alan~W Black, and Yulia Tsvetkov.
  2019.
\newblock Measuring bias in contextualized word representations.
\newblock In \emph{Proceedings of the First Workshop on Gender Bias in Natural
  Language Processing}, pages 166--172.

\bibitem[{Kwon et~al.(2019)Kwon, Kang, Han, and Choi}]{kwon-etal-2019-why}
Kwon, Sunjae, Cheongwoong Kang, Jiyeon Han, and Jaesik Choi. 2019.
\newblock Why do masked neural language models still need common sense
  knowledge?
\newblock \emph{ArXiv}, arXiv:1911.03024.

\bibitem[{Lakhotia et~al.(2021)Lakhotia, Kharitonov, Hsu, Adi, Polyak, Bolte,
  Nguyen, Copet, Baevski, Mohamed, and Dupoux}]{lakhotia-etal-2021-generative}
Lakhotia, Kushal, Eugene Kharitonov, Wei-Ning Hsu, Yossi Adi, Adam Polyak,
  Benjamin Bolte, Tu-Anh Nguyen, Jade Copet, Alexei Baevski, Abdelrahman
  Mohamed, and Emmanuel Dupoux. 2021.
\newblock On generative spoken language modeling from raw audio.
\newblock \emph{Transactions of the Association for Computational Linguistics},
  9:1336--1354.

\bibitem[{Lakretz et~al.(2022)Lakretz, Desbordes, Hupkes, and
  Dehaene}]{lakretz-etal-2022-can}
Lakretz, Yair, Th{\'e}o Desbordes, Dieuwke Hupkes, and Stanislas Dehaene. 2022.
\newblock Can transformers process recursive nested constructions, like humans?
\newblock In \emph{Proceedings of the 29th International Conference on
  Computational Linguistics}, pages 3226--3232, International Committee on
  Computational Linguistics, Gyeongju, Republic of Korea.

\bibitem[{Lal et~al.(2022)Lal, Tandon, Aggarwal, Liu, Chambers, Mooney, and
  Balasubramanian}]{lal-2022-using}
Lal, Yash~Kumar, Niket Tandon, Tanvi Aggarwal, Horace Liu, Nathanael Chambers,
  Raymond Mooney, and Niranjan Balasubramanian. 2022.
\newblock Using commonsense knowledge to answer why-questions.
\newblock In \emph{Proceedings of the 2022 Conference on Empirical Methods in
  Natural Language Processing}, pages 1204--1219.

\bibitem[{Lampinen(2022)}]{lampinen-2022-can}
Lampinen, Andrew. 2022.
\newblock Can language models handle recursively nested grammatical structures?
  a case study on comparing models and humans.
\newblock \emph{ArXiv}, arXiv:2210.15303.

\bibitem[{Lampinen et~al.(2022)Lampinen, Dasgupta, Chan, Mathewson, Tessler,
  Creswell, McClelland, Wang, and Hill}]{lampinen-etal-2022-can}
Lampinen, Andrew, Ishita Dasgupta, Stephanie Chan, Kory Mathewson, Mh~Tessler,
  Antonia Creswell, James McClelland, Jane Wang, and Felix Hill. 2022.
\newblock Can language models learn from explanations in context?
\newblock In \emph{Findings of the Association for Computational Linguistics:
  EMNLP 2022}, pages 537--563.

\bibitem[{Lasri, Lenci, and Poibeau(2022{\natexlab{a}})}]{lasri-etal-2022-does}
Lasri, Karim, Alessandro Lenci, and Thierry Poibeau. 2022{\natexlab{a}}.
\newblock Does {BERT} really agree ? fine-grained analysis of lexical
  dependence on a syntactic task.
\newblock In \emph{Findings of the Association for Computational Linguistics:
  ACL 2022}, pages 2309--2315.

\bibitem[{Lasri, Lenci, and Poibeau(2022{\natexlab{b}})}]{lasri-etal-2022-word}
Lasri, Karim, Alessandro Lenci, and Thierry Poibeau. 2022{\natexlab{b}}.
\newblock Word order matters when you increase masking.
\newblock In \emph{Proceedings of the 2022 Conference on Empirical Methods in
  Natural Language Processing}, pages 1808--1815.

\bibitem[{Lasri et~al.(2022)Lasri, Seminck, Lenci, and
  Poibeau}]{lasri-etal-2022-subject}
Lasri, Karim, Olga Seminck, Alessandro Lenci, and Thierry Poibeau. 2022.
\newblock Subject verb agreement error patterns in meaningless sentences:
  Humans vs. {BERT}.
\newblock In \emph{Proceedings of the 29th International Conference on
  Computational Linguistics}, pages 37--43, International Committee on
  Computational Linguistics, Gyeongju, Republic of Korea.

\bibitem[{Lee(2023)}]{lee-2023-what}
Lee, Angie. 2023.
\newblock What are large language models used for?
\newblock \emph{{NVIDIA} Blog}.

\bibitem[{Lee et~al.(2023)Lee, Le, Chen, and Lee}]{lee-etal-2022-do}
Lee, Jooyoung, Thai Le, Jinghui Chen, and Dongwon Lee. 2023.
\newblock Do language models plagiarize?
\newblock In \emph{The ACM Web Conference}, pages 3637--3647.

\bibitem[{Lee et~al.(2022)Lee, Ippolito, Nystrom, Zhang, Eck, Callison-Burch,
  and Carlini}]{lee-etal-2021-deduplicating}
Lee, Katherine, Daphne Ippolito, Andrew Nystrom, Chiyuan Zhang, Douglas Eck,
  Chris Callison-Burch, and Nicholas Carlini. 2022.
\newblock Deduplicating training data makes language models better.
\newblock In \emph{Proceedings of the 60th Annual Meeting of the Association
  for Computational Linguistics (Volume 1: Long Papers)}, pages 8424--8445.

\bibitem[{Lee et~al.(2021)Lee, Bang, Madotto, and Fung}]{lee-etal-2021-towards}
Lee, Nayeon, Yejin Bang, Andrea Madotto, and Pascale Fung. 2021.
\newblock Towards few-shot fact-checking via perplexity.
\newblock In \emph{Proceedings of the 2021 Conference of the North American
  Chapter of the Association for Computational Linguistics: Human Language
  Technologies}, pages 1971--1981.

\bibitem[{Lee and Schuster(2022)}]{lee-schuster-2022-can}
Lee, Soo-Hwan and Sebastian Schuster. 2022.
\newblock Can language models capture syntactic associations without surface
  cues? a case study of reflexive anaphor licensing in {E}nglish control
  constructions.
\newblock In \emph{Proceedings of the Society for Computation in Linguistics
  2022}, pages 206--211.

\bibitem[{Lees et~al.(2022)Lees, Tran, Tay, Sorensen, Gupta, Metzler, and
  Vasserman}]{lees-etal-2022-new}
Lees, Alyssa, Vinh~Q. Tran, Yi~Tay, Jeffrey Sorensen, Jai Gupta, Donald
  Metzler, and Lucy Vasserman. 2022.
\newblock A new generation of {Perspective} {API}: Efficient multilingual
  character-level {T}ransformers.
\newblock In \emph{Proceedings of the 28th ACM SIGKDD Conference on Knowledge
  Discovery and Data Mining}, page 3197–3207.

\bibitem[{Lehman et~al.(2021)Lehman, Jain, Pichotta, Goldberg, and
  Wallace}]{lehman-etal-2021-does}
Lehman, Eric, Sarthak Jain, Karl Pichotta, Yoav Goldberg, and Byron Wallace.
  2021.
\newblock Does {BERT} pretrained on clinical notes reveal sensitive data?
\newblock In \emph{Proceedings of the 2021 Conference of the North American
  Chapter of the Association for Computational Linguistics: Human Language
  Technologies}, pages 946--959.

\bibitem[{Lester, Al-Rfou, and Constant(2021)}]{lester-etal-2021-the}
Lester, Brian, Rami Al-Rfou, and Noah Constant. 2021.
\newblock The power of scale for parameter-efficient prompt tuning.
\newblock In \emph{Proceedings of the 2021 Conference on Empirical Methods in
  Natural Language Processing}, pages 3045--3059.

\bibitem[{Levy et~al.(2022)Levy, Allaway, Subbiah, Chilton, Patton, McKeown,
  and Wang}]{levy-etal-2022-safetext}
Levy, Sharon, Emily Allaway, Melanie Subbiah, Lydia Chilton, Desmond Patton,
  Kathleen McKeown, and William~Yang Wang. 2022.
\newblock {S}afe{T}ext: A benchmark for exploring physical safety in language
  models.
\newblock In \emph{Proceedings of the 2022 Conference on Empirical Methods in
  Natural Language Processing}, pages 2407--2421.

\bibitem[{Levy, Saxon, and Wang(2021)}]{levy-etal-2021-investigating}
Levy, Sharon, Michael Saxon, and William~Yang Wang. 2021.
\newblock Investigating memorization of conspiracy theories in text generation.
\newblock In \emph{Findings of the Association for Computational Linguistics:
  ACL-IJCNLP 2021}, pages 4718--4729.

\bibitem[{Li, Yu, and Ettinger(2022)}]{li-2022-counterfactual}
Li, Jiaxuan, Lang Yu, and Allyson Ettinger. 2022.
\newblock Counterfactual reasoning: Do language models need world knowledge for
  causal inference?
\newblock In \emph{Workshop on Neuro Causal and Symbolic AI (nCSI)}.

\bibitem[{Li et~al.(2021)Li, Tang, Zhao, and Wen}]{li-etal-2021-pretrained}
Li, Junyi, Tianyi Tang, Wayne~Xin Zhao, and Ji-Rong Wen. 2021.
\newblock Pretrained language models for text generation: {A} survey.
\newblock In \emph{International Joint Conference on Artificial Intelligence},
  pages 4492--4499.

\bibitem[{Li et~al.(2022{\natexlab{a}})Li, Kuncoro, Hoffmann,
  de~Masson~d{'}Autume, Blunsom, and Nematzadeh}]{li-etal-2021-a}
Li, Xiang~Lorraine, Adhiguna Kuncoro, Jordan Hoffmann, Cyprien
  de~Masson~d{'}Autume, Phil Blunsom, and Aida Nematzadeh. 2022{\natexlab{a}}.
\newblock A systematic investigation of commonsense knowledge in large language
  models.
\newblock In \emph{Proceedings of the 2022 Conference on Empirical Methods in
  Natural Language Processing}, pages 11838--11855.

\bibitem[{Li et~al.(2022{\natexlab{b}})Li, Li, Liu, Bing, and
  Joty}]{li-etal-2022-is}
Li, Xingxuan, Yutong Li, Linlin Liu, Lidong Bing, and Shafiq Joty.
  2022{\natexlab{b}}.
\newblock Is {GPT-3} a psychopath? evaluating large language models from a
  psychological perspective.
\newblock \emph{ArXiv}, arXiv:2212.10529.

\bibitem[{Lieber et~al.(2021)Lieber, Sharir, Lenz, and
  Shoham}]{lieber-etal-2021-jurassic}
Lieber, Opher, Or~Sharir, Barak Lenz, and Yoav Shoham. 2021.
\newblock Jurassic-1: Technical details and evaluation.
\newblock \emph{White Paper. AI21 Labs}.

\bibitem[{Lin et~al.(2020)Lin, Lee, Khanna, and Ren}]{lin-etal-2020-birds}
Lin, Bill~Yuchen, Seyeon Lee, Rahul Khanna, and Xiang Ren. 2020.
\newblock {B}irds have four legs?! {N}umer{S}ense: {P}robing {N}umerical
  {C}ommonsense {K}nowledge of {P}re-{T}rained {L}anguage {M}odels.
\newblock In \emph{Proceedings of the 2020 Conference on Empirical Methods in
  Natural Language Processing (EMNLP)}, pages 6862--6868.

\bibitem[{Lin, Hilton, and Evans(2022)}]{lin-etal-2022-truthfulqa}
Lin, Stephanie, Jacob Hilton, and Owain Evans. 2022.
\newblock {T}ruthful{QA}: Measuring how models mimic human falsehoods.
\newblock In \emph{Proceedings of the 60th Annual Meeting of the Association
  for Computational Linguistics (Volume 1: Long Papers)}, pages 3214--3252.

\bibitem[{Liu et~al.(2022{\natexlab{a}})Liu, Cui, Zheng, and
  Neubig}]{liu-etal-2022-testing}
Liu, Emmy, Chenxuan Cui, Kenneth Zheng, and Graham Neubig. 2022{\natexlab{a}}.
\newblock Testing the ability of language models to interpret figurative
  language.
\newblock In \emph{Proceedings of the 2022 Conference of the North American
  Chapter of the Association for Computational Linguistics: Human Language
  Technologies}, pages 4437--4452.

\bibitem[{Liu et~al.(2022{\natexlab{b}})Liu, Eisenschlos, Cole, and
  Collier}]{liu-etal-2022-do}
Liu, Fangyu, Julian Eisenschlos, Jeremy Cole, and Nigel Collier.
  2022{\natexlab{b}}.
\newblock Do ever larger octopi still amplify reporting biases? evidence from
  judgments of typical colour.
\newblock In \emph{Proceedings of the 2nd Conference of the Asia-Pacific
  Chapter of the Association for Computational Linguistics and the 12th
  International Joint Conference on Natural Language Processing (Volume 2:
  Short Papers)}, pages 210--220.

\bibitem[{Liu et~al.(2022{\natexlab{c}})Liu, Jia, Wei, Xu, and
  Vosoughi}]{liu-etal-2022-quantifying}
Liu, Ruibo, Chenyan Jia, Jason Wei, Guangxuan Xu, and Soroush Vosoughi.
  2022{\natexlab{c}}.
\newblock Quantifying and alleviating political bias in language models.
\newblock \emph{Artificial Intelligence}, 304:103654.

\bibitem[{Liu et~al.(2019)Liu, Ott, Goyal, Du, Joshi, Chen, Levy, Lewis,
  Zettlemoyer, and Stoyanov}]{liu-etal-2019-roberta}
Liu, Yinhan, Myle Ott, Naman Goyal, Jingfei Du, Mandar Joshi, Danqi Chen, Omer
  Levy, Mike Lewis, Luke Zettlemoyer, and Veselin Stoyanov. 2019.
\newblock {RoBERTa}: A robustly optimized {BERT} pretraining approach.
\newblock \emph{ArXiv}, arXiv:1907.11692.

\bibitem[{Liu et~al.(2021)Liu, Wang, Kasai, Hajishirzi, and
  Smith}]{liu-etal-2021-probing}
Liu, Zeyu, Yizhong Wang, Jungo Kasai, Hannaneh Hajishirzi, and Noah~A. Smith.
  2021.
\newblock Probing across time: What does {R}o{BERT}a know and when?
\newblock In \emph{Findings of the Association for Computational Linguistics:
  EMNLP 2021}, pages 820--842.

\bibitem[{Magee et~al.(2021)Magee, Ghahremanlou, Soldatic, and
  Robertson}]{magee-etal-2021-intersectional}
Magee, Liam, Lida Ghahremanlou, Karen Soldatic, and Shanthi Robertson. 2021.
\newblock Intersectional bias in causal language models.
\newblock \emph{ArXiv}, arXiv:2107.07691.

\bibitem[{Mahowald(2023)}]{mahowald-2023-a}
Mahowald, Kyle. 2023.
\newblock A discerning several thousand judgments: {GPT-3} rates the article +
  adjective + numeral + noun construction.
\newblock In \emph{Proceedings of the 17th Conference of the European Chapter
  of the Association for Computational Linguistics}, pages 265--273.

\bibitem[{Mahowald et~al.(2022)Mahowald, Diachek, Gibson, Fedorenko, and
  Futrell}]{mahowald-etal-2022-grammatical}
Mahowald, Kyle, Evgeniia Diachek, Edward Gibson, Evelina Fedorenko, and Richard
  Futrell. 2022.
\newblock Experimentally measuring the redundancy of grammatical cues in
  transitive clauses.
\newblock \emph{ArXiv}, arXiv:2201.12911.

\bibitem[{Malkin et~al.(2021)Malkin, Lanka, Goel, and
  Jojic}]{malkin-etal-2021-studying}
Malkin, Nikolay, Sameera Lanka, Pranav Goel, and Nebojsa Jojic. 2021.
\newblock Studying word order through iterative shuffling.
\newblock In \emph{Proceedings of the 2021 Conference on Empirical Methods in
  Natural Language Processing}, pages 10351--10366.

\bibitem[{Mallen et~al.(2022)Mallen, Asai, Zhong, Das, Hajishirzi, and
  Khashabi}]{mallen-etal-2022-when}
Mallen, Alex, Akari Asai, Victor Zhong, Rajarshi Das, Hannaneh Hajishirzi, and
  Daniel Khashabi. 2022.
\newblock When not to trust language models: Investigating effectiveness and
  limitations of parametric and non-parametric memories.
\newblock \emph{ArXiv}, arXiv:2212.10511.

\bibitem[{Marr(2010)}]{marr-2010-vision}
Marr, David. 2010.
\newblock \emph{{Vision: A Computational Investigation into the Human
  Representation and Processing of Visual Information}}.
\newblock The MIT Press.

\bibitem[{Masis and Anderson(2021)}]{masis-anderson-2021-prosper}
Masis, Tessa and Carolyn Anderson. 2021.
\newblock {P}ro{SP}er: Probing human and neural network language model
  understanding of spatial perspective.
\newblock In \emph{Proceedings of the Fourth BlackboxNLP Workshop on Analyzing
  and Interpreting Neural Networks for NLP}, pages 95--135.

\bibitem[{Massarelli et~al.(2020)Massarelli, Petroni, Piktus, Ott,
  Rockt{"a}schel, Plachouras, Silvestri, and Riedel}]{massarelli-etal-2019-how}
Massarelli, Luca, Fabio Petroni, Aleksandra Piktus, Myle Ott, Tim
  Rockt{"a}schel, Vassilis Plachouras, Fabrizio Silvestri, and Sebastian
  Riedel. 2020.
\newblock How decoding strategies affect the verifiability of generated text.
\newblock In \emph{Findings of the Association for Computational Linguistics:
  EMNLP 2020}, pages 223--235.

\bibitem[{Mattern et~al.(2022)Mattern, Jin, Sachan, Mihalcea, and
  Sch\"{o}lkopf}]{mattern-etal-2022-understanding}
Mattern, Justus, Zhijing Jin, Mrinmaya Sachan, Rada Mihalcea, and
  B.~Sch\"{o}lkopf. 2022.
\newblock Understanding stereotypes in language models: Towards robust
  measurement and zero-shot debiasing.
\newblock \emph{ArXiv}, arXiv:2212.10678.

\bibitem[{McCoy et~al.(2021)McCoy, Smolensky, Linzen, Gao, and
  Celikyilmaz}]{mccoy-etal-2021-how}
McCoy, R.~Thomas, Paul Smolensky, Tal Linzen, Jianfeng Gao, and Asli
  Celikyilmaz. 2021.
\newblock How much do language models copy from their training data? evaluating
  linguistic novelty in text generation using {RAVEN}.
\newblock \emph{ArXiv}, arXiv:2111.09509.

\bibitem[{McCoy, Frank, and Linzen(2018)}]{mccoy-etal-2018-revisiting}
McCoy, Thomas, Robert Frank, and Tal Linzen. 2018.
\newblock Revisiting the poverty of the stimulus: Hierarchical generalization
  without a hierarchical bias in recurrent neural networks.
\newblock In \emph{Proceedings of the Annual Meeting of the Cognitive Science
  Society}, volume~40, pages 2096--2101.

\bibitem[{McCoy, Pavlick, and Linzen(2019)}]{mccoy-etal-2019-right}
McCoy, Tom, Ellie Pavlick, and Tal Linzen. 2019.
\newblock Right for the wrong reasons: Diagnosing syntactic heuristics in
  natural language inference.
\newblock In \emph{Proceedings of the 57th Annual Meeting of the Association
  for Computational Linguistics}, pages 3428--3448.

\bibitem[{Meade, Poole-Dayan, and Reddy(2022)}]{meade-etal-2021-an}
Meade, Nicholas, Elinor Poole-Dayan, and Siva Reddy. 2022.
\newblock An empirical survey of the effectiveness of debiasing techniques for
  pre-trained language models.
\newblock In \emph{Proceedings of the 60th Annual Meeting of the Association
  for Computational Linguistics (Volume 1: Long Papers)}, pages 1878--1898.

\bibitem[{Mehrabi et~al.(2022)Mehrabi, Beirami, Morstatter, and
  Galstyan}]{mehrabi-etal-2022-robust}
Mehrabi, Ninareh, Ahmad Beirami, Fred Morstatter, and Aram Galstyan. 2022.
\newblock Robust conversational agents against imperceptible toxicity triggers.
\newblock In \emph{Proceedings of the 2022 Conference of the North American
  Chapter of the Association for Computational Linguistics: Human Language
  Technologies}, pages 2831--2847.

\bibitem[{Meister and Cotterell(2021)}]{meister-cotterell-2021-language}
Meister, Clara and Ryan Cotterell. 2021.
\newblock Language model evaluation beyond perplexity.
\newblock In \emph{Proceedings of the 59th Annual Meeting of the Association
  for Computational Linguistics and the 11th International Joint Conference on
  Natural Language Processing (Volume 1: Long Papers)}, pages 5328--5339.

\bibitem[{Meng et~al.(2022)Meng, Bau, Andonian, and
  Belinkov}]{meng-etal-2022-locating}
Meng, Kevin, David Bau, Alex~J Andonian, and Yonatan Belinkov. 2022.
\newblock Locating and editing factual associations in {GPT}.
\newblock In \emph{Advances in Neural Information Processing Systems},
  volume~35, pages 17359--17372.

\bibitem[{Miaschi et~al.(2020)Miaschi, Brunato, Dell{'}Orletta, and
  Venturi}]{miaschi-etal-2020-linguistic}
Miaschi, Alessio, Dominique Brunato, Felice Dell{'}Orletta, and Giulia Venturi.
  2020.
\newblock Linguistic profiling of a neural language model.
\newblock In \emph{Proceedings of the 28th International Conference on
  Computational Linguistics}, pages 745--756, International Committee on
  Computational Linguistics, Barcelona, Spain (Online).

\bibitem[{Michaelov and
  Bergen(2022{\natexlab{a}})}]{michaelov-bergen-2022-collateral}
Michaelov, James and Benjamin Bergen. 2022{\natexlab{a}}.
\newblock Collateral facilitation in humans and language models.
\newblock In \emph{Proceedings of the 26th Conference on Computational Natural
  Language Learning (CoNLL)}, pages 13--26.

\bibitem[{Michaelov and
  Bergen(2022{\natexlab{b}})}]{michaelov-bergen-2022-rarely}
Michaelov, James and Benjamin Bergen. 2022{\natexlab{b}}.
\newblock {`}rarely{'} a problem? language models exhibit inverse scaling in
  their predictions following {`}few{'}-type quantifiers.
\newblock \emph{ArXiv}, arXiv:2212.08700.

\bibitem[{Min et~al.(2022)Min, Lyu, Holtzman, Artetxe, Lewis, Hajishirzi, and
  Zettlemoyer}]{min-etal-2022-rethinking}
Min, Sewon, Xinxi Lyu, Ari Holtzman, Mikel Artetxe, Mike Lewis, Hannaneh
  Hajishirzi, and Luke Zettlemoyer. 2022.
\newblock Rethinking the role of demonstrations: What makes in-context learning
  work?
\newblock In \emph{Proceedings of the 2022 Conference on Empirical Methods in
  Natural Language Processing}, pages 11048--11064.

\bibitem[{Miotto, Rossberg, and Kleinberg(2022)}]{miotto-etal-2022-who}
Miotto, Maril{\`u}, Nicola Rossberg, and Bennett Kleinberg. 2022.
\newblock Who is {GPT}-3? an exploration of personality, values and
  demographics.
\newblock In \emph{Proceedings of the Fifth Workshop on Natural Language
  Processing and Computational Social Science (NLP+CSS)}, pages 218--227.

\bibitem[{Misra(2022)}]{misra-2022-minicons}
Misra, Kanishka. 2022.
\newblock {minicons}: Enabling flexible behavioral and representational
  analyses of {T}ransformer language models.
\newblock \emph{ArXiv}, arXiv:2203.13112.

\bibitem[{Misra, Ettinger, and Rayz(2020)}]{misra-etal-2020-exploring}
Misra, Kanishka, Allyson Ettinger, and Julia Rayz. 2020.
\newblock Exploring {BERT}{'}s sensitivity to lexical cues using tests from
  semantic priming.
\newblock In \emph{Findings of the Association for Computational Linguistics:
  EMNLP 2020}, pages 4625--4635.

\bibitem[{Misra, Ettinger, and Rayz(2021)}]{misra-etal-2021-do}
Misra, Kanishka, Allyson Ettinger, and Julia~Taylor Rayz. 2021.
\newblock Do language models learn typicality judgments from text?
\newblock In \emph{Proceedings of the Annual Meeting of the Cognitive Science
  Society}, volume~43, pages 216--222.

\bibitem[{Misra, Rayz, and Ettinger(2023)}]{misra-etal-2022-comps}
Misra, Kanishka, Julia~Taylor Rayz, and Allyson Ettinger. 2023.
\newblock {COMPS}: Conceptual minimal pair sentences for testing property
  knowledge and inheritance in pre-trained language models.
\newblock In \emph{Proceedings of the 17th Conference of the European Chapter
  of the Association for Computational Linguistics}, pages 2928--2949.

\bibitem[{Mitchell and Krakauer(2022)}]{mitchell-krakauer-2022-the}
Mitchell, Melanie and David Krakauer. 2022.
\newblock The debate over understanding in {AI}'s large language models.
\newblock \emph{ArXiv}, arXiv:2210.13966.

\bibitem[{Monroe and Potts(2015)}]{monroe-potts-2015-learning}
Monroe, Will and Christopher Potts. 2015.
\newblock Learning in the rational speech acts model.
\newblock \emph{ArXiv}, arXiv:1510.06807.

\bibitem[{Mosbach et~al.(2020)Mosbach, Khokhlova, Hedderich, and
  Klakow}]{mosbach-etal-2020-on}
Mosbach, Marius, Anna Khokhlova, Michael~A. Hedderich, and Dietrich Klakow.
  2020.
\newblock {O}n the interplay between fine-tuning and sentence-level probing for
  linguistic knowledge in pre-trained {T}ransformers.
\newblock In \emph{Findings of the Association for Computational Linguistics:
  EMNLP 2020}, pages 2502--2516.

\bibitem[{Nadeem, Bethke, and Reddy(2021)}]{nadeem-etal-2020-stereoset}
Nadeem, Moin, Anna Bethke, and Siva Reddy. 2021.
\newblock {S}tereo{S}et: Measuring stereotypical bias in pretrained language
  models.
\newblock In \emph{Proceedings of the 59th Annual Meeting of the Association
  for Computational Linguistics and the 11th International Joint Conference on
  Natural Language Processing (Volume 1: Long Papers)}, pages 5356--5371.

\bibitem[{Nangia et~al.(2020)Nangia, Vania, Bhalerao, and
  Bowman}]{nangia-etal-2020-crows}
Nangia, Nikita, Clara Vania, Rasika Bhalerao, and Samuel~R. Bowman. 2020.
\newblock {C}row{S}-pairs: A challenge dataset for measuring social biases in
  masked language models.
\newblock In \emph{Proceedings of the 2020 Conference on Empirical Methods in
  Natural Language Processing (EMNLP)}, pages 1953--1967.

\bibitem[{Nayak(2019)}]{nayak-2019-understanding}
Nayak, Pandu. 2019.
\newblock Understanding searches better than ever before.
\newblock \emph{The Keyword: Google Blog}.

\bibitem[{Newman et~al.(2021)Newman, Ang, Gong, and
  Hewitt}]{newman-etal-2021-refining}
Newman, Benjamin, Kai-Siang Ang, Julia Gong, and John Hewitt. 2021.
\newblock Refining targeted syntactic evaluation of language models.
\newblock In \emph{Proceedings of the 2021 Conference of the North American
  Chapter of the Association for Computational Linguistics: Human Language
  Technologies}, pages 3710--3723.

\bibitem[{Nozza, Bianchi, and Hovy(2021)}]{nozza-etal-2021-honest}
Nozza, Debora, Federico Bianchi, and Dirk Hovy. 2021.
\newblock {HONEST}: Measuring hurtful sentence completion in language models.
\newblock In \emph{Proceedings of the 2021 Conference of the North American
  Chapter of the Association for Computational Linguistics: Human Language
  Technologies}, pages 2398--2406.

\bibitem[{Nozza et~al.(2022)Nozza, Bianchi, Lauscher, and
  Hovy}]{nozza-etal-2022-measuring}
Nozza, Debora, Federico Bianchi, Anne Lauscher, and Dirk Hovy. 2022.
\newblock Measuring harmful sentence completion in language models for
  {LGBTQIA}+ individuals.
\newblock In \emph{Proceedings of the Second Workshop on Language Technology
  for Equality, Diversity and Inclusion}, pages 26--34.

\bibitem[{O{'}Connor and Andreas(2021)}]{oconnor-andreas-2021-what}
O{'}Connor, Joe and Jacob Andreas. 2021.
\newblock What context features can transformer language models use?
\newblock In \emph{Proceedings of the 59th Annual Meeting of the Association
  for Computational Linguistics and the 11th International Joint Conference on
  Natural Language Processing (Volume 1: Long Papers)}, pages 851--864.

\bibitem[{Olsson et~al.(2022)Olsson, Elhage, Nanda, Joseph, DasSarma, Henighan,
  Mann, Askell, Bai, Chen, Conerly, Drain, Ganguli, Hatfield-Dodds, Hernandez,
  Johnston, Jones, Kernion, Lovitt, Ndousse, Amodei, Brown, Clark, Kaplan,
  McCandlish, and Olah}]{olsson-etal-2022-in}
Olsson, Catherine, Nelson Elhage, Neel Nanda, Nicholas Joseph, Nova DasSarma,
  Tom Henighan, Benjamin Mann, Amanda Askell, Yuntao Bai, Anna Chen, Tom
  Conerly, Dawn Drain, Deep Ganguli, Zac Hatfield-Dodds, Danny Hernandez, Scott
  Johnston, Andy Jones, John Kernion, Liane Lovitt, Kamal Ndousse, Dario
  Amodei, Tom~B. Brown, Jack Clark, Jared Kaplan, Sam McCandlish, and Chris
  Olah. 2022.
\newblock In-context learning and induction heads.
\newblock \emph{ArXiv}, arXiv:2209.11895.

\bibitem[{OpenAI(2022)}]{openai-2021-chatgpt}
OpenAI. 2022.
\newblock {ChatGPT}: Optimizing language models for dialogue.
\newblock \emph{{OpenAI} Blog}.

\bibitem[{OpenAI(2023{\natexlab{a}})}]{openai-2023-gpt4}
OpenAI. 2023{\natexlab{a}}.
\newblock {GPT-4} technical report.
\newblock \emph{OpenAI}.

\bibitem[{OpenAI(2023{\natexlab{b}})}]{openai-2023-model}
OpenAI. 2023{\natexlab{b}}.
\newblock Model index for researchers.
\newblock \emph{OpenAI}.

\bibitem[{Ousidhoum et~al.(2021)Ousidhoum, Zhao, Fang, Song, and
  Yeung}]{ousidhoum-etal-2021-probing}
Ousidhoum, Nedjma, Xinran Zhao, Tianqing Fang, Yangqiu Song, and Dit-Yan Yeung.
  2021.
\newblock Probing toxic content in large pre-trained language models.
\newblock In \emph{Proceedings of the 59th Annual Meeting of the Association
  for Computational Linguistics and the 11th International Joint Conference on
  Natural Language Processing (Volume 1: Long Papers)}, pages 4262--4274.

\bibitem[{Ouyang et~al.(2022)Ouyang, Wu, Jiang, Almeida, Wainwright, Mishkin,
  Zhang, Agarwal, Slama, Gray, Schulman, Hilton, Kelton, Miller, Simens,
  Askell, Welinder, Christiano, Leike, and Lowe}]{ouyang-etal-2022-training}
Ouyang, Long, Jeffrey Wu, Xu~Jiang, Diogo Almeida, Carroll Wainwright, Pamela
  Mishkin, Chong Zhang, Sandhini Agarwal, Katarina Slama, Alex Gray, John
  Schulman, Jacob Hilton, Fraser Kelton, Luke Miller, Maddie Simens, Amanda
  Askell, Peter Welinder, Paul Christiano, Jan Leike, and Ryan Lowe. 2022.
\newblock Training language models to follow instructions with human feedback.
\newblock In \emph{Advances in Neural Information Processing Systems},
  volume~35, pages 27730--27744.

\bibitem[{Paik et~al.(2021)Paik, Aroca-Ouellette, Roncone, and
  Kann}]{paik-etal-2021-the}
Paik, Cory, St{\'e}phane Aroca-Ouellette, Alessandro Roncone, and Katharina
  Kann. 2021.
\newblock {T}he {W}orld of an {O}ctopus: {H}ow {R}eporting {B}ias {I}nfluences
  a {L}anguage {M}odel{'}s {P}erception of {C}olor.
\newblock In \emph{Proceedings of the 2021 Conference on Empirical Methods in
  Natural Language Processing}, pages 823--835.

\bibitem[{Pandia, Cong, and Ettinger(2021)}]{pandia-etal-2021-pragmatic}
Pandia, Lalchand, Yan Cong, and Allyson Ettinger. 2021.
\newblock Pragmatic competence of pre-trained language models through the lens
  of discourse connectives.
\newblock In \emph{Proceedings of the 25th Conference on Computational Natural
  Language Learning}, pages 367--379.

\bibitem[{Pandia and Ettinger(2021)}]{pandia-ettinger-2021-sorting}
Pandia, Lalchand and Allyson Ettinger. 2021.
\newblock Sorting through the noise: Testing robustness of information
  processing in pre-trained language models.
\newblock In \emph{Proceedings of the 2021 Conference on Empirical Methods in
  Natural Language Processing}, pages 1583--1596.

\bibitem[{Pandit and Hou(2021)}]{pandit-hou-2021-probing}
Pandit, Onkar and Yufang Hou. 2021.
\newblock Probing for bridging inference in transformer language models.
\newblock In \emph{Proceedings of the 2021 Conference of the North American
  Chapter of the Association for Computational Linguistics: Human Language
  Technologies}, pages 4153--4163.

\bibitem[{Park, Park, and Song(2021)}]{park-etal-2021-deep}
Park, Kwonsik, Myung-Kwan Park, and Sanghoun Song. 2021.
\newblock Deep learning can contrast the minimal pairs of syntactic data.
\newblock \emph{Linguistic Research}, 38(2):395--424.

\bibitem[{Patel and Pavlick(2021)}]{patel-pavlick-2021-was}
Patel, Roma and Ellie Pavlick. 2021.
\newblock {``}was it {``}stated{''} or was it {``}claimed{''}?: How linguistic
  bias affects generative language models.
\newblock In \emph{Proceedings of the 2021 Conference on Empirical Methods in
  Natural Language Processing}, pages 10080--10095.

\bibitem[{Pedinotti et~al.(2021{\natexlab{a}})Pedinotti, Di~Palma, Cerini, and
  Lenci}]{pedinotti-etal-2021-a}
Pedinotti, Paolo, Eliana Di~Palma, Ludovica Cerini, and Alessandro Lenci.
  2021{\natexlab{a}}.
\newblock A howling success or a working sea? testing what {BERT} knows about
  metaphors.
\newblock In \emph{Proceedings of the Fourth BlackboxNLP Workshop on Analyzing
  and Interpreting Neural Networks for NLP}, pages 192--204.

\bibitem[{Pedinotti et~al.(2021{\natexlab{b}})Pedinotti, Rambelli, Chersoni,
  Santus, Lenci, and Blache}]{pedinotti-etal-2021-did}
Pedinotti, Paolo, Giulia Rambelli, Emmanuele Chersoni, Enrico Santus,
  Alessandro Lenci, and Philippe Blache. 2021{\natexlab{b}}.
\newblock Did the cat drink the coffee? challenging transformers with
  generalized event knowledge.
\newblock In \emph{Proceedings of *SEM 2021: The Tenth Joint Conference on
  Lexical and Computational Semantics}, pages 1--11.

\bibitem[{Peng et~al.(2022)Peng, Wang, Hu, Jin, Hou, Li, Liu, and
  Liu}]{peng-etal-2022-copen}
Peng, Hao, Xiaozhi Wang, Shengding Hu, Hailong Jin, Lei Hou, Juanzi Li, Zhiyuan
  Liu, and Qun Liu. 2022.
\newblock {COPEN}: Probing conceptual knowledge in pre-trained language models.
\newblock In \emph{Proceedings of the 2022 Conference on Empirical Methods in
  Natural Language Processing}, pages 5015--5035.

\bibitem[{Penha and Hauff(2020)}]{penha-hauff-2020-what}
Penha, Gustavo and Claudia Hauff. 2020.
\newblock What does {BERT} know about books, movies and music? probing {BERT}
  for conversational recommendation.
\newblock In \emph{Proceedings of the 14th ACM Conference on Recommender
  Systems}, page 388–397, Association for Computing Machinery, New York, NY,
  USA.

\bibitem[{Perez et~al.(2022{\natexlab{a}})Perez, Huang, Song, Cai, Ring,
  Aslanides, Glaese, McAleese, and Irving}]{perez-etal-2022-red}
Perez, Ethan, Saffron Huang, Francis Song, Trevor Cai, Roman Ring, John
  Aslanides, Amelia Glaese, Nat McAleese, and Geoffrey Irving.
  2022{\natexlab{a}}.
\newblock Red teaming language models with language models.
\newblock In \emph{Proceedings of the 2022 Conference on Empirical Methods in
  Natural Language Processing}, pages 3419--3448.

\bibitem[{Perez et~al.(2022{\natexlab{b}})Perez, Ringer, Lukosiute, Nguyen,
  Chen, Heiner, Pettit, Olsson, Kundu, Kadavath, Jones, Chen, Mann, Israel,
  Seethor, McKinnon, Olah, Yan, Amodei, Amodei, Drain, Li, Tran-Johnson,
  Khundadze, Kernion, Landis, Kerr, Mueller, Hyun, Landau, Ndousse, Goldberg,
  Lovitt, Lucas, Sellitto, Zhang, Kingsland, Elhage, Joseph, Mercado, DasSarma,
  Rausch, Larson, McCandlish, Johnston, Kravec, Showk, Lanham, Telleen-Lawton,
  Brown, Henighan, Hume, Bai, Hatfield-Dodds, Clark, Bowman, Askell, Grosse,
  Hernandez, Ganguli, Hubinger, Schiefer, and
  Kaplan}]{perez-etal-2022-discovering}
Perez, Ethan, Sam Ringer, Kamile Lukosiute, Karina Nguyen, Edwin Chen, Scott
  Heiner, Craig Pettit, Catherine Olsson, Sandipan Kundu, Saurav Kadavath, Andy
  Jones, Anna Chen, Benjamin Mann, Brian Israel, Bryan Seethor, Cameron
  McKinnon, Christopher Olah, Daisong Yan, Daniela Amodei, Dario Amodei, Dawn
  Drain, Dustin Li, Eli Tran-Johnson, Guro Khundadze, John Kernion, James
  Landis, Jamie Kerr, Jared Mueller, Jeeyoon Hyun, Joshua Landau, Kamal
  Ndousse, Landon Goldberg, Liane Lovitt, Martin Lucas, Michael Sellitto,
  Miranda Zhang, Neerav Kingsland, Nelson Elhage, Nicholas Joseph, Noem{\'i}
  Mercado, Nova DasSarma, Oliver Rausch, Robin Larson, Sam McCandlish, Scott
  Johnston, Shauna Kravec, Sheer~El Showk, Tamera Lanham, Timothy
  Telleen-Lawton, Tom Brown, Tom Henighan, Tristan Hume, Yuntao Bai, Zac
  Hatfield-Dodds, Jack Clark, Sam Bowman, Amanda Askell, Roger Grosse, Danny
  Hernandez, Deep Ganguli, Evan Hubinger, Nicholas Schiefer, and Jared Kaplan.
  2022{\natexlab{b}}.
\newblock Discovering language model behaviors with model-written evaluations.
\newblock \emph{ArXiv}, arXiv:2212.09251.

\bibitem[{P{\'e}rez-Mayos, Ballesteros, and Wanner(2021)}]{mayos-etal-2021-how}
P{\'e}rez-Mayos, Laura, Miguel Ballesteros, and Leo Wanner. 2021.
\newblock How much pretraining data do language models need to learn syntax?
\newblock In \emph{Proceedings of the 2021 Conference on Empirical Methods in
  Natural Language Processing}, pages 1571--1582.

\bibitem[{Petroni et~al.(2019)Petroni, Rockt{\"a}schel, Riedel, Lewis, Bakhtin,
  Wu, and Miller}]{petroni-etal-2019-language}
Petroni, Fabio, Tim Rockt{\"a}schel, Sebastian Riedel, Patrick Lewis, Anton
  Bakhtin, Yuxiang Wu, and Alexander Miller. 2019.
\newblock Language models as knowledge bases?
\newblock In \emph{Proceedings of the 2019 Conference on Empirical Methods in
  Natural Language Processing and the 9th International Joint Conference on
  Natural Language Processing (EMNLP-IJCNLP)}, pages 2463--2473.

\bibitem[{Petty and Frank(2021)}]{petty-frank-2021-transformers}
Petty, Jackson and Robert Frank. 2021.
\newblock Transformers generalize linearly.
\newblock \emph{ArXiv}, arXiv:2109.12036.

\bibitem[{Piantadosi and
  Fedorenko(2017)}]{piantadosi-fedorenko-2017-infinitely}
Piantadosi, Steven and Evelina Fedorenko. 2017.
\newblock Infinitely productive language can arise from chance under
  communicative pressure.
\newblock \emph{Journal of Language Evolution}, 2(2):141--147.

\bibitem[{Podkorytov, Bis, and Liu(2021)}]{podkorytov-etal-2021-how}
Podkorytov, Maksim, Daniel Bis, and Xiuwen Liu. 2021.
\newblock How can the [{MASK}] know? the sources and limitations of knowledge
  in {BERT}.
\newblock In \emph{IEEE International Joint Conference on Neural Networks},
  pages 1--8.

\bibitem[{Poerner, Waltinger, and Sch\"{u}tze(2019)}]{poerner-etal-2019-bert}
Poerner, Nina, Ulli Waltinger, and Hinrich Sch\"{u}tze. 2019.
\newblock {BERT} is not a knowledge base (yet): Factual knowledge vs.
  name-based reasoning in unsupervised {QA}.
\newblock \emph{ArXiv}, arXiv:1911.03681.

\bibitem[{Porada, Sordoni, and Cheung(2022)}]{porada-etal-2021-does}
Porada, Ian, Alessandro Sordoni, and Jackie Cheung. 2022.
\newblock Does pre-training induce systematic inference? how masked language
  models acquire commonsense knowledge.
\newblock In \emph{Proceedings of the 2022 Conference of the North American
  Chapter of the Association for Computational Linguistics: Human Language
  Technologies}, pages 4550--4557.

\bibitem[{Prefors, Regier, and Tenenbaum(2006)}]{prefors-etal-2006-poverty}
Prefors, Amy, Terry Regier, and Joshua Tenenbaum. 2006.
\newblock Poverty of the stimulus? a rational approach.
\newblock In \emph{Proceedings of the Annual Meeting of the Cognitive Science
  Society}, volume~28, pages 663--668.

\bibitem[{Press, Smith, and Lewis(2022)}]{press-etal-2021-train}
Press, Ofir, Noah Smith, and Mike Lewis. 2022.
\newblock Train short, test long: Attention with linear biases enables input
  length extrapolation.
\newblock In \emph{International Conference on Learning Representations}.

\bibitem[{Press et~al.(2022)Press, Zhang, Min, Schmidt, Smith, and
  Lewis}]{press-etal-2022-measuring}
Press, Ofir, Muru Zhang, Sewon Min, Ludwig Schmidt, Noah~A. Smith, and Mike
  Lewis. 2022.
\newblock Measuring and narrowing the compositionality gap in language models.
\newblock \emph{ArXiv}, arXiv:2210.03350.

\bibitem[{Qin et~al.(2021)Qin, Gupta, Upadhyay, He, Choi, and
  Faruqui}]{qin-etal-2021-timedial}
Qin, Lianhui, Aditya Gupta, Shyam Upadhyay, Luheng He, Yejin Choi, and Manaal
  Faruqui. 2021.
\newblock {TIMEDIAL}: Temporal commonsense reasoning in dialog.
\newblock In \emph{Proceedings of the 59th Annual Meeting of the Association
  for Computational Linguistics and the 11th International Joint Conference on
  Natural Language Processing (Volume 1: Long Papers)}, pages 7066--7076.

\bibitem[{Qiu et~al.(2022)Qiu, Shaw, Pasupat, Shi, Herzig, Pitler, Sha, and
  Toutanova}]{qiu-etal-2022-evaluating}
Qiu, Linlu, Peter Shaw, Panupong Pasupat, Tianze Shi, Jonathan Herzig, Emily
  Pitler, Fei Sha, and Kristina Toutanova. 2022.
\newblock Evaluating the impact of model scale for compositional generalization
  in semantic parsing.
\newblock In \emph{Proceedings of the 2022 Conference on Empirical Methods in
  Natural Language Processing}, pages 9157--9179.

\bibitem[{Radford et~al.(2022)Radford, Kim, Xu, Brockman, McLeavey, and
  Sutskever}]{radford-etal-2022-robust}
Radford, Alec, Jong~Wook Kim, Tao Xu, Greg Brockman, Christine McLeavey, and
  Ilya Sutskever. 2022.
\newblock Robust speech recognition via large-scale weak supervision.
\newblock \emph{ArXiv}, arXiv:2212.04356.

\bibitem[{Radford et~al.(2018)Radford, Narasimhan, Salimans, and
  Sutskever}]{radford-etal-2018-improving}
Radford, Alec, Karthik Narasimhan, Tim Salimans, and Ilya Sutskever. 2018.
\newblock Improving language understanding by generative pre-training.
\newblock \emph{OpenAI}.

\bibitem[{Radford et~al.(2019)Radford, Wu, Child, Luan, Amodei, and
  Sutskever}]{radford-etal-2019-language}
Radford, Alec, Jeff Wu, Rewon Child, David Luan, Dario Amodei, and Ilya
  Sutskever. 2019.
\newblock Language models are unsupervised multitask learners.
\newblock \emph{OpenAI}.

\bibitem[{Rae et~al.(2021)Rae, Borgeaud, Cai, Millican, Hoffmann, Song,
  Aslanides, Henderson, Ring, Young, Rutherford, Hennigan, Menick, Cassirer,
  Powell, van~den Driessche, Hendricks, Rauh, Huang, Glaese, Welbl, Dathathri,
  Huang, Uesato, Mellor, Higgins, Creswell, McAleese, Wu, Elsen, Jayakumar,
  Buchatskaya, Budden, Sutherland, Simonyan, Paganini, Sifre, Martens, Li,
  Kuncoro, Nematzadeh, Gribovskaya, Donato, Lazaridou, Mensch, Lespiau,
  Tsimpoukelli, Grigorev, Fritz, Sottiaux, Pajarskas, Pohlen, Gong, Toyama,
  de~Masson~d'Autume, Li, Terzi, Mikulik, Babuschkin, Clark, de~Las~Casas, Guy,
  Jones, Bradbury, Johnson, Hechtman, Weidinger, Gabriel, Isaac, Lockhart,
  Osindero, Rimell, Dyer, Vinyals, Ayoub, Stanway, Bennett, Hassabis,
  Kavukcuoglu, and Irving}]{rae-etal-2021-scaling}
Rae, Jack~W., Sebastian Borgeaud, Trevor Cai, Katie Millican, Jordan Hoffmann,
  Francis Song, John Aslanides, Sarah Henderson, Roman Ring, Susannah Young,
  Eliza Rutherford, Tom Hennigan, Jacob Menick, Albin Cassirer, Richard Powell,
  George van~den Driessche, Lisa~Anne Hendricks, Maribeth Rauh, Po-Sen Huang,
  Amelia Glaese, Johannes Welbl, Sumanth Dathathri, Saffron Huang, Jonathan
  Uesato, John F.~J. Mellor, Irina Higgins, Antonia Creswell, Nathan McAleese,
  Amy Wu, Erich Elsen, Siddhant~M. Jayakumar, Elena Buchatskaya, David Budden,
  Esme Sutherland, Karen Simonyan, Michela Paganini, L.~Sifre, Lena Martens,
  Xiang~Lorraine Li, Adhiguna Kuncoro, Aida Nematzadeh, Elena Gribovskaya,
  Domenic Donato, Angeliki Lazaridou, Arthur Mensch, Jean-Baptiste Lespiau,
  Maria Tsimpoukelli, N.~K. Grigorev, Doug Fritz, Thibault Sottiaux, Mantas
  Pajarskas, Tobias Pohlen, Zhitao Gong, Daniel Toyama, Cyprien
  de~Masson~d'Autume, Yujia Li, Tayfun Terzi, Vladimir Mikulik, Igor
  Babuschkin, Aidan Clark, Diego de~Las~Casas, Aurelia Guy, Chris Jones, James
  Bradbury, Matthew~G. Johnson, Blake~A. Hechtman, Laura Weidinger, Iason
  Gabriel, William~S. Isaac, Edward Lockhart, Simon Osindero, Laura Rimell,
  Chris Dyer, Oriol Vinyals, Kareem~W. Ayoub, Jeff Stanway, L.~L. Bennett,
  Demis Hassabis, Koray Kavukcuoglu, and Geoffrey Irving. 2021.
\newblock Scaling language models: Methods, analysis \& insights from training
  gopher.
\newblock \emph{ArXiv}, arXiv:2112.11446.

\bibitem[{Raffel et~al.(2020)Raffel, Shazeer, Roberts, Lee, Narang, Matena,
  Zhou, Li, and Liu}]{raffel-etal-2020-exploring}
Raffel, Colin, Noam Shazeer, Adam Roberts, Katherine Lee, Sharan Narang,
  Michael Matena, Yanqi Zhou, Wei Li, and Peter~J. Liu. 2020.
\newblock Exploring the limits of transfer learning with a unified text-to-text
  {T}ransformer.
\newblock \emph{Journal of Machine Learning Research}, 21(140):5485--5551.

\bibitem[{Raj, Rosati, and Majumdar(2022)}]{raj-etal-2022-measuring}
Raj, Harsha, Domenic Rosati, and Subhabrata Majumdar. 2022.
\newblock Measuring reliability of large language models through semantic
  consistency.
\newblock In \emph{NeurIPS ML Safety Workshop}.

\bibitem[{Ravichander et~al.(2020)Ravichander, Hovy, Suleman, Trischler, and
  Cheung}]{ravichander-etal-2020-on}
Ravichander, Abhilasha, Eduard Hovy, Kaheer Suleman, Adam Trischler, and Jackie
  Chi~Kit Cheung. 2020.
\newblock On the systematicity of probing contextualized word representations:
  The case of hypernymy in {BERT}.
\newblock In \emph{Proceedings of the Ninth Joint Conference on Lexical and
  Computational Semantics}, pages 88--102.

\bibitem[{Razeghi et~al.(2022)Razeghi, Logan~IV, Gardner, and
  Singh}]{razeghi-etal-2022-impact}
Razeghi, Yasaman, Robert Logan~IV, Matt Gardner, and Sameer Singh. 2022.
\newblock Impact of pretraining term frequencies on few-shot numerical
  reasoning.
\newblock In \emph{Findings of the Association for Computational Linguistics:
  EMNLP 2022}, pages 840--854.

\bibitem[{Reif et~al.(2022)Reif, Ippolito, Yuan, Coenen, Callison-Burch, and
  Wei}]{reif-etal-2021-a}
Reif, Emily, Daphne Ippolito, Ann Yuan, Andy Coenen, Chris Callison-Burch, and
  Jason Wei. 2022.
\newblock A recipe for arbitrary text style transfer with large language
  models.
\newblock In \emph{Proceedings of the 60th Annual Meeting of the Association
  for Computational Linguistics (Volume 2: Short Papers)}, pages 837--848.

\bibitem[{Reif et~al.(2019)Reif, Yuan, Wattenberg, Viegas, Coenen, Pearce, and
  Kim}]{coenen-etal-2019-visualizing}
Reif, Emily, Ann Yuan, Martin Wattenberg, Fernanda~B Viegas, Andy Coenen, Adam
  Pearce, and Been Kim. 2019.
\newblock Visualizing and measuring the geometry of {BERT}.
\newblock In \emph{Advances in Neural Information Processing Systems},
  volume~32, pages 8594--8603.

\bibitem[{Rogers, Kovaleva, and Rumshisky(2020)}]{rogers-etal-2020-a}
Rogers, Anna, Olga Kovaleva, and Anna Rumshisky. 2020.
\newblock A primer in {BERT}ology: What we know about how {BERT} works.
\newblock \emph{Transactions of the Association for Computational Linguistics},
  8:842--866.

\bibitem[{Romero and Razniewski(2022)}]{romero-razniewski-2022-do}
Romero, Julien and Simon Razniewski. 2022.
\newblock Do children texts hold the key to commonsense knowledge?
\newblock In \emph{Proceedings of the 2022 Conference on Empirical Methods in
  Natural Language Processing}, pages 10954--10959.

\bibitem[{Ruis et~al.(2022)Ruis, Khan, Biderman, Hooker, Rocktaschel, and
  Grefenstette}]{ruis-etal-2022-large}
Ruis, Laura, Akbir Khan, Stella~Rose Biderman, Sara Hooker, Tim Rocktaschel,
  and Edward Grefenstette. 2022.
\newblock Large language models are not zero-shot communicators.
\newblock \emph{ArXiv}, arXiv:2210.14986.

\bibitem[{Ryu and Lewis(2021)}]{ryu-lewis-2021-accounting}
Ryu, Soo~Hyun and Richard Lewis. 2021.
\newblock Accounting for agreement phenomena in sentence comprehension with
  transformer language models: Effects of similarity-based interference on
  surprisal and attention.
\newblock In \emph{Proceedings of the Workshop on Cognitive Modeling and
  Computational Linguistics}, pages 61--71.

\bibitem[{Sahu et~al.(2022)Sahu, Cogswell, Gong, and
  Divakaran}]{sahu-etal-2022-unpacking}
Sahu, Pritish, Michael Cogswell, Yunye Gong, and Ajay Divakaran. 2022.
\newblock Unpacking large language models with conceptual consistency.
\newblock \emph{ArXiv}, arXiv:2209.15093.

\bibitem[{Sancheti and Rudinger(2022)}]{sancheti-rudinger-2021-what}
Sancheti, Abhilasha and Rachel Rudinger. 2022.
\newblock What do large language models learn about scripts?
\newblock In \emph{Proceedings of the 11th Joint Conference on Lexical and
  Computational Semantics}, pages 1--11.

\bibitem[{Sanh et~al.(2019)Sanh, Debut, Chaumond, and
  Wolf}]{sanh-etal-2019-distilbert}
Sanh, Victor, Lysandre Debut, Julien Chaumond, and Thomas Wolf. 2019.
\newblock {DistilBERT}, a distilled version of {BERT}: Smaller, faster, cheaper
  and lighter.
\newblock In \emph{Workshop on Energy Efficient Machine Learning and Cognitive
  Computing}.

\bibitem[{Sap et~al.(2022)Sap, Le~Bras, Fried, and Choi}]{sap-etal-2022-neural}
Sap, Maarten, Ronan Le~Bras, Daniel Fried, and Yejin Choi. 2022.
\newblock Neural theory-of-mind? on the limits of social intelligence in large
  {LM}s.
\newblock In \emph{Proceedings of the 2022 Conference on Empirical Methods in
  Natural Language Processing}, pages 3762--3780.

\bibitem[{Saparov and He(2023)}]{saparov-he-2022-language}
Saparov, Abulhair and He~He. 2023.
\newblock Language models are greedy reasoners: A systematic formal analysis of
  chain-of-thought.
\newblock In \emph{International Conference on Learning Representations}.

\bibitem[{Schick et~al.(2023)Schick, Dwivedi-Yu, Dess{\`i}, Raileanu, Lomeli,
  Zettlemoyer, Cancedda, and Scialom}]{schick-etal-2023-toolformer}
Schick, Timo, Jane Dwivedi-Yu, Roberto Dess{\`i}, Roberta Raileanu, Maria
  Lomeli, Luke Zettlemoyer, Nicola Cancedda, and Thomas Scialom. 2023.
\newblock Toolformer: {L}anguage models can teach themselves to use tools.
\newblock \emph{ArXiv}, arXiv:2302.04761.

\bibitem[{van Schijndel, Mueller, and
  Linzen(2019)}]{schijndel-etal-2019-quantity}
van Schijndel, Marten, Aaron Mueller, and Tal Linzen. 2019.
\newblock Quantity doesn{'}t buy quality syntax with neural language models.
\newblock In \emph{Proceedings of the 2019 Conference on Empirical Methods in
  Natural Language Processing and the 9th International Joint Conference on
  Natural Language Processing (EMNLP-IJCNLP)}, pages 5831--5837.

\bibitem[{Schuster and Linzen(2022)}]{schuster-linzen-2022-when}
Schuster, Sebastian and Tal Linzen. 2022.
\newblock When a sentence does not introduce a discourse entity,
  transformer-based models still sometimes refer to it.
\newblock In \emph{Proceedings of the 2022 Conference of the North American
  Chapter of the Association for Computational Linguistics: Human Language
  Technologies}, pages 969--982.

\bibitem[{Senel and Sch{\"u}tze(2021)}]{senel-schutze-2021-does}
Senel, Lutfi~Kerem and Hinrich Sch{\"u}tze. 2021.
\newblock Does she wink or does she nod? a challenging benchmark for evaluating
  word understanding of language models.
\newblock In \emph{Proceedings of the 16th Conference of the European Chapter
  of the Association for Computational Linguistics: Main Volume}, pages
  532--538.

\bibitem[{Sennrich, Haddow, and Birch(2016)}]{sennrich-etal-2016-neural}
Sennrich, Rico, Barry Haddow, and Alexandra Birch. 2016.
\newblock Neural machine translation of rare words with subword units.
\newblock In \emph{Proceedings of the 54th Annual Meeting of the Association
  for Computational Linguistics (Volume 1: Long Papers)}, pages 1715--1725.

\bibitem[{Serrano and Smith(2019)}]{serrano-smith-2019-is}
Serrano, Sofia and Noah~A. Smith. 2019.
\newblock Is attention interpretable?
\newblock In \emph{Proceedings of the 57th Annual Meeting of the Association
  for Computational Linguistics}, pages 2931--2951.

\bibitem[{Seshadri, Pezeshkpour, and
  Singh(2022)}]{seshadri-etal-2022-quantifying}
Seshadri, Preethi, Pouya Pezeshkpour, and Sameer Singh. 2022.
\newblock Quantifying social biases using templates is unreliable.
\newblock In \emph{Workshop on Trustworthy and Socially Responsible Machine
  Learning}.

\bibitem[{Shaikh et~al.(2022)Shaikh, Zhang, Held, Bernstein, and
  Yang}]{shaikh-etal-2022-on}
Shaikh, Omar, Hongxin Zhang, William~B. Held, Michael Bernstein, and Diyi Yang.
  2022.
\newblock On second thought, let's not think step by step! {B}ias and toxicity
  in zero-shot reasoning.
\newblock \emph{ArXiv}, arXiv:2212.08061.

\bibitem[{Shardlow and
  Przybyla(2022)}]{shardlow-przybyla-2022-deanthropomorphising}
Shardlow, Matthew and Piotr Przybyla. 2022.
\newblock Deanthropomorphising {NLP}: Can a language model be conscious?
\newblock \emph{ArXiv}, arXiv:2211.11483.

\bibitem[{Shaw, Uszkoreit, and Vaswani(2018)}]{shaw-etal-2018-self}
Shaw, Peter, Jakob Uszkoreit, and Ashish Vaswani. 2018.
\newblock Self-attention with relative position representations.
\newblock In \emph{Proceedings of the 2018 Conference of the North {A}merican
  Chapter of the Association for Computational Linguistics: Human Language
  Technologies, Volume 2 (Short Papers)}, pages 464--468.

\bibitem[{Sheng et~al.(2021{\natexlab{a}})Sheng, Chang, Natarajan, and
  Peng}]{sheng-etal-2021-nice}
Sheng, Emily, Kai-Wei Chang, Prem Natarajan, and Nanyun Peng.
  2021{\natexlab{a}}.
\newblock {``}nice try, kiddo{''}: Investigating ad hominems in dialogue
  responses.
\newblock In \emph{Proceedings of the 2021 Conference of the North American
  Chapter of the Association for Computational Linguistics: Human Language
  Technologies}, pages 750--767.

\bibitem[{Sheng et~al.(2021{\natexlab{b}})Sheng, Chang, Natarajan, and
  Peng}]{sheng-etal-2021-societal}
Sheng, Emily, Kai-Wei Chang, Prem Natarajan, and Nanyun Peng.
  2021{\natexlab{b}}.
\newblock Societal biases in language generation: Progress and challenges.
\newblock In \emph{Proceedings of the 59th Annual Meeting of the Association
  for Computational Linguistics and the 11th International Joint Conference on
  Natural Language Processing (Volume 1: Long Papers)}, pages 4275--4293.

\bibitem[{Sheng et~al.(2019)Sheng, Chang, Natarajan, and
  Peng}]{sheng-etal-2019-woman}
Sheng, Emily, Kai-Wei Chang, Premkumar Natarajan, and Nanyun Peng. 2019.
\newblock The woman worked as a babysitter: On biases in language generation.
\newblock In \emph{Proceedings of the 2019 Conference on Empirical Methods in
  Natural Language Processing and the 9th International Joint Conference on
  Natural Language Processing (EMNLP-IJCNLP)}, pages 3407--3412.

\bibitem[{Shi et~al.(2023)Shi, Chen, Misra, Scales, Dohan, Chi, Sch{\"a}rli,
  and Zhou}]{shi-etal-2023-large}
Shi, Freda, Xinyun Chen, Kanishka Misra, Nathan Scales, David Dohan, Ed~Chi,
  Nathanael Sch{\"a}rli, and Denny Zhou. 2023.
\newblock Large language models can be easily distracted by irrelevant context.
\newblock \emph{ArXiv}, arXiv:2302.00093.

\bibitem[{Shi and Wolff(2021)}]{haohan-wolff-2021-what}
Shi, Haohan and Phillip Wolff. 2021.
\newblock What {T}ransformers might know about the physical world: {T5} and the
  origins of knowledge.
\newblock In \emph{Proceedings of the Annual Meeting of the Cognitive Science
  Society}, volume~43, pages 2218--2224.

\bibitem[{Shwartz and Choi(2020)}]{shwartz-choi-2020-do}
Shwartz, Vered and Yejin Choi. 2020.
\newblock Do neural language models overcome reporting bias?
\newblock In \emph{Proceedings of the 28th International Conference on
  Computational Linguistics}, pages 6863--6870, International Committee on
  Computational Linguistics, Barcelona, Spain (Online).

\bibitem[{Shwartz, Rudinger, and Tafjord(2020)}]{shwartz-etal-2020-you}
Shwartz, Vered, Rachel Rudinger, and Oyvind Tafjord. 2020.
\newblock {``}you are grounded!{''}: Latent name artifacts in pre-trained
  language models.
\newblock In \emph{Proceedings of the 2020 Conference on Empirical Methods in
  Natural Language Processing (EMNLP)}, pages 6850--6861.

\bibitem[{Sikos et~al.(2021)Sikos, Venhuizen, Drenhaus, and
  Crocker}]{sikos-etal-2021-reevaluating}
Sikos, Les, Noortje Venhuizen, Heiner Drenhaus, and Matthew Crocker. 2021.
\newblock Reevaluating pragmatic reasoning in language games.
\newblock \emph{PLOS One}, 16(3):1--33.

\bibitem[{Silva, Tambwekar, and Gombolay(2021)}]{silva-etal-2021-towards}
Silva, Andrew, Pradyumna Tambwekar, and Matthew Gombolay. 2021.
\newblock Towards a comprehensive understanding and accurate evaluation of
  societal biases in pre-trained transformers.
\newblock In \emph{Proceedings of the 2021 Conference of the North American
  Chapter of the Association for Computational Linguistics: Human Language
  Technologies}, pages 2383--2389.

\bibitem[{Simmons(2022)}]{simmons-2022-moral}
Simmons, Gabriel. 2022.
\newblock Moral mimicry: Large language models produce moral rationalizations
  tailored to political identity.
\newblock \emph{ArXiv}, arXiv:2209.12106.

\bibitem[{Sinclair et~al.(2022)Sinclair, Jumelet, Zuidema, and
  Fern{\'a}ndez}]{sinclair-etal-2022-structural}
Sinclair, Arabella, Jaap Jumelet, Willem Zuidema, and Raquel Fern{\'a}ndez.
  2022.
\newblock Structural persistence in language models: Priming as a window into
  abstract language representations.
\newblock \emph{Transactions of the Association for Computational Linguistics},
  10:1031--1050.

\bibitem[{Sinha et~al.(2022{\natexlab{a}})Sinha, Gauthier, Mueller, Misra,
  Fuentes, Levy, and Williams}]{sinha-etal-2022-language}
Sinha, Koustuv, Jon Gauthier, Aaron Mueller, Kanishka Misra, Keren Fuentes,
  Roger Levy, and Adina Williams. 2022{\natexlab{a}}.
\newblock Language model acceptability judgements are not always robust to
  context.
\newblock \emph{ArXiv}, arXiv:2212.08979.

\bibitem[{Sinha et~al.(2021)Sinha, Jia, Hupkes, Pineau, Williams, and
  Kiela}]{sinha-etal-2021-masked}
Sinha, Koustuv, Robin Jia, Dieuwke Hupkes, Joelle Pineau, Adina Williams, and
  Douwe Kiela. 2021.
\newblock Masked language modeling and the distributional hypothesis: Order
  word matters pre-training for little.
\newblock In \emph{Proceedings of the 2021 Conference on Empirical Methods in
  Natural Language Processing}, pages 2888--2913.

\bibitem[{Sinha et~al.(2022{\natexlab{b}})Sinha, Kazemnejad, Reddy, Pineau,
  Hupkes, and Williams}]{sinha-etal-2022-the}
Sinha, Koustuv, Amirhossein Kazemnejad, Siva Reddy, Joelle Pineau, Dieuwke
  Hupkes, and Adina Williams. 2022{\natexlab{b}}.
\newblock The curious case of absolute position embeddings.
\newblock In \emph{Findings of the Association for Computational Linguistics:
  EMNLP 2022}, pages 4449--4472.

\bibitem[{Smith et~al.(2022{\natexlab{a}})Smith, Hall, Kambadur, Presani, and
  Williams}]{smith-etal-2022-im}
Smith, Eric~Michael, Melissa Hall, Melanie Kambadur, Eleonora Presani, and
  Adina Williams. 2022{\natexlab{a}}.
\newblock {``}{I}{'}m sorry to hear that{''}: Finding new biases in language
  models with a holistic descriptor dataset.
\newblock In \emph{Proceedings of the 2022 Conference on Empirical Methods in
  Natural Language Processing}, pages 9180--9211.

\bibitem[{Smith et~al.(2022{\natexlab{b}})Smith, Patwary, Norick, LeGresley,
  Rajbhandari, Casper, Liu, Prabhumoye, Zerveas, Korthikanti, Zhang, Child,
  Aminabadi, Bernauer, Song, Shoeybi, He, Houston, Tiwary, and
  Catanzaro}]{smith-etal-2022-using}
Smith, Shaden, Mostofa Patwary, Brandon Norick, Patrick LeGresley, Samyam
  Rajbhandari, Jared Casper, Zhun Liu, Shrimai Prabhumoye, George Zerveas,
  Vijay~Anand Korthikanti, Elton Zhang, Rewon Child, Reza~Yazdani Aminabadi,
  Julie Bernauer, Xia Song, Mohammad Shoeybi, Yuxiong He, Michael Houston,
  Saurabh Tiwary, and Bryan Catanzaro. 2022{\natexlab{b}}.
\newblock Using {DeepSpeed} and {Megatron} to train {Megatron-Turing} {NLG}
  {530B}, a large-scale generative language model.
\newblock \emph{ArXiv}, arXiv:2201.11990.

\bibitem[{Spitale, Biller-Andorno, and Germani(2023)}]{spitale-etal-2023-ai}
Spitale, Giovanni, Nikola Biller-Andorno, and Federico Germani. 2023.
\newblock {AI} model {GPT-3} (dis)informs us better than humans.
\newblock \emph{ArXiv}, arXiv:2301.11924.

\bibitem[{Srivastava et~al.(2022)Srivastava, Rastogi, Rao, Shoeb, Abid, Fisch,
  Brown, Santoro, Gupta, Garriga-Alonso et~al.}]{srivastava-etal-2022-beyond}
Srivastava, Aarohi, Abhinav Rastogi, Abhishek Rao, Abu Awal~Md Shoeb, Abubakar
  Abid, Adam Fisch, Adam~R Brown, Adam Santoro, Aditya Gupta, Adri{\`a}
  Garriga-Alonso, et~al. 2022.
\newblock Beyond the imitation game: Quantifying and extrapolating the
  capabilities of language models.
\newblock \emph{ArXiv}, arXiv:2206.04615.

\bibitem[{Stammbach, Antoniak, and Ash(2022)}]{stammbach-etal-2022-heroes}
Stammbach, Dominik, Maria Antoniak, and Elliott Ash. 2022.
\newblock Heroes, villains, and victims, and {GPT}-3: Automated extraction of
  character roles without training data.
\newblock In \emph{Proceedings of the 4th Workshop of Narrative Understanding
  (WNU2022)}, pages 47--56.

\bibitem[{Stevenson et~al.(2022)Stevenson, Smal, Baas, Grasman, and van~der
  Maas}]{stevenson-etal-2022-putting}
Stevenson, Claire, Iris Smal, Matthijs Baas, Raoul Grasman, and Han van~der
  Maas. 2022.
\newblock Putting {GPT-3}'s creativity to the (alternative uses) test.
\newblock In \emph{International Conference on Computational Creativity}, pages
  164--168.

\bibitem[{Stolfo et~al.(2022)Stolfo, Jin, Shridhar, Sch\"{o}lkopf, and
  Sachan}]{stolfo-etal-2022-a}
Stolfo, Alessandro, Zhijing Jin, Kumar Shridhar, Bernhard Sch\"{o}lkopf, and
  Mrinmaya Sachan. 2022.
\newblock A causal framework to quantify the robustness of mathematical
  reasoning with language models.
\newblock \emph{ArXiv}, arXiv:2210.12023.

\bibitem[{Strubell, Ganesh, and McCallum(2019)}]{strubell-etal-2019-energy}
Strubell, Emma, Ananya Ganesh, and Andrew McCallum. 2019.
\newblock Energy and policy considerations for deep learning in {NLP}.
\newblock In \emph{Proceedings of the 57th Annual Meeting of the Association
  for Computational Linguistics}, pages 3645--3650.

\bibitem[{Su et~al.(2021)Su, Lu, Pan, Murtadha, Wen, and
  Liu}]{su-etal-2021-roformer}
Su, Jianlin, Yu~Lu, Shengfeng Pan, Ahmed Murtadha, Bo~Wen, and Yunfeng Liu.
  2021.
\newblock {RoFormer}: Enhanced {Transformer} with rotary position embedding.
\newblock \emph{ArXiv}, arXiv:2104.09864.

\bibitem[{Summers-Stay, Bonial, and Voss(2021)}]{stay-etal-2021-what}
Summers-Stay, Douglas, Claire Bonial, and Clare Voss. 2021.
\newblock What can a generative language model answer about a passage?
\newblock In \emph{Proceedings of the 3rd Workshop on Machine Reading for
  Question Answering}, pages 73--81.

\bibitem[{Suzgun et~al.(2022)Suzgun, Scales, Scharli, Gehrmann, Tay, Chung,
  Chowdhery, Le, Chi, Zhou, and Wei}]{suzgun-etal-2022-challenging}
Suzgun, Mirac, Nathan Scales, Nathanael Scharli, Sebastian Gehrmann, Yi~Tay,
  Hyung~Won Chung, Aakanksha Chowdhery, Quoc~V. Le, Ed~Chi, Denny Zhou, and
  Jason Wei. 2022.
\newblock Challenging {BIG-Bench} tasks and whether chain-of-thought can solve
  them.
\newblock \emph{ArXiv}, arXiv:2210.09261.

\bibitem[{Swamy, Romanou, and Jaggi(2021)}]{swamy-etal-2021-interpreting}
Swamy, Vinitra, Angelika Romanou, and Martin Jaggi. 2021.
\newblock Interpreting language models through knowledge graph extraction.
\newblock In \emph{Workshop on eXplainable AI Approaches for Debugging and
  Diagnosis.}

\bibitem[{Tal, Magar, and Schwartz(2022)}]{tal-etal-2022-fewer}
Tal, Yarden, Inbal Magar, and Roy Schwartz. 2022.
\newblock Fewer errors, but more stereotypes? the effect of model size on
  gender bias.
\newblock In \emph{Proceedings of the 4th Workshop on Gender Bias in Natural
  Language Processing (GeBNLP)}, pages 112--120.

\bibitem[{Tamborrino et~al.(2020)Tamborrino, Pellican{\`o}, Pannier, Voitot,
  and Naudin}]{tamborrino-etal-2020-pre}
Tamborrino, Alexandre, Nicola Pellican{\`o}, Baptiste Pannier, Pascal Voitot,
  and Louise Naudin. 2020.
\newblock Pre-training is (almost) all you need: An application to commonsense
  reasoning.
\newblock In \emph{Proceedings of the 58th Annual Meeting of the Association
  for Computational Linguistics}, pages 3878--3887.

\bibitem[{Tang and Jiang(2022)}]{tang-jiang-2022-gender}
Tang, Kenan and Hanchun Jiang. 2022.
\newblock Gender biases unexpectedly fluctuate in the pre-training stage of
  masked language models.
\newblock \emph{ArXiv}, arXiv:2211.14639.

\bibitem[{Tay et~al.(2022{\natexlab{a}})Tay, Dehghani, Abnar, Chung, Fedus,
  Rao, Narang, Tran, Yogatama, and Metzler}]{tay-etal-2022-scaling}
Tay, Yi, Mostafa Dehghani, Samira Abnar, Hyung~Won Chung, W.~Fedus, Jinfeng
  Rao, Sharan Narang, Vinh Tran, Dani Yogatama, and Donald Metzler.
  2022{\natexlab{a}}.
\newblock Scaling laws vs model architectures: How does inductive bias
  influence scaling?
\newblock \emph{ArXiv}, arXiv:2207.10551.

\bibitem[{Tay et~al.(2022{\natexlab{b}})Tay, Dehghani, Rao, Fedus, Abnar,
  Chung, Narang, Yogatama, Vaswani, and Metzler}]{tay-etal-2021-scale}
Tay, Yi, Mostafa Dehghani, Jinfeng Rao, William Fedus, Samira Abnar, Hyung~Won
  Chung, Sharan Narang, Dani Yogatama, Ashish Vaswani, and Donald Metzler.
  2022{\natexlab{b}}.
\newblock Scale efficiently: Insights from pre-training and fine-tuning
  {Transformers}.
\newblock In \emph{International Conference on Learning Representations}.

\bibitem[{Tejada, Scholtes, and Spanakis(2021)}]{tejada-etal-2021-a}
Tejada, Giorgia Nidia~Carranza, Johannes Scholtes, and Gerasimos Spanakis.
  2021.
\newblock A study of {BERT}’s processing of negations to determine sentiment.
\newblock In \emph{Benelux Conference on Artificial Intelligence and the
  Belgian Dutch Conference on Machine Learning}, pages 47--59.

\bibitem[{Tenney, Das, and Pavlick(2019)}]{tenney-etal-2019-bert}
Tenney, Ian, Dipanjan Das, and Ellie Pavlick. 2019.
\newblock {BERT} rediscovers the classical {NLP} pipeline.
\newblock In \emph{Proceedings of the 57th Annual Meeting of the Association
  for Computational Linguistics}, pages 4593--4601.

\bibitem[{Thewsey(2021)}]{thewsey-2021-bring}
Thewsey, Alex. 2021.
\newblock Bring structure to diverse documents with {Amazon} {Textract} and
  transformer-based models on {Amazon} {SageMaker}.
\newblock \emph{{AWS} Machine Learning Blog}.

\bibitem[{Thoppilan et~al.(2022)Thoppilan, Freitas, Hall, Shazeer,
  Kulshreshtha, Cheng, Jin, Bos, Baker, Du, Li, Lee, Zheng, Ghafouri, Menegali,
  Huang, Krikun, Lepikhin, Qin, Chen, Xu, Chen, Roberts, Bosma, Zhou, Chang,
  Krivokon, Rusch, Pickett, Meier-Hellstern, Morris, Doshi, Santos, Duke,
  S{\o}raker, Zevenbergen, Prabhakaran, D{\'i}az, Hutchinson, Olson, Molina,
  Hoffman-John, Lee, Aroyo, Rajakumar, Butryna, Lamm, Kuzmina, Fenton, Cohen,
  Bernstein, Kurzweil, Aguera-Arcas, Cui, Croak, hsin Chi, and
  Le}]{thoppilan-etal-2022-lamda}
Thoppilan, Romal, Daniel~De Freitas, Jamie Hall, Noam~M. Shazeer, Apoorv
  Kulshreshtha, Heng-Tze Cheng, Alicia Jin, Taylor Bos, Leslie Baker, Yu~Du,
  Yaguang Li, Hongrae Lee, Huaixiu~Steven Zheng, Amin Ghafouri, Marcelo
  Menegali, Yanping Huang, Maxim Krikun, Dmitry Lepikhin, James Qin, Dehao
  Chen, Yuanzhong Xu, Zhifeng Chen, Adam Roberts, Maarten Bosma, Yanqi Zhou,
  Chung-Ching Chang, I.~A. Krivokon, Willard~James Rusch, Marc Pickett,
  Kathleen~S. Meier-Hellstern, Meredith~Ringel Morris, Tulsee Doshi,
  Renelito~Delos Santos, Toju Duke, Johnny~Hartz S{\o}raker, Ben Zevenbergen,
  Vinodkumar Prabhakaran, Mark D{\'i}az, Ben Hutchinson, Kristen Olson,
  Alejandra Molina, Erin Hoffman-John, Josh Lee, Lora Aroyo, Ravindran
  Rajakumar, Alena Butryna, Matthew Lamm, V.~O. Kuzmina, Joseph Fenton, Aaron
  Cohen, Rachel Bernstein, Ray Kurzweil, Blaise Aguera-Arcas, Claire Cui,
  Marian~Rogers Croak, Ed~Huai hsin Chi, and Quoc Le. 2022.
\newblock {LaMDA}: Language models for dialog applications.
\newblock \emph{ArXiv}, arXiv:2201.08239.

\bibitem[{Tirumala et~al.(2022)Tirumala, Markosyan, Zettlemoyer, and
  Aghajanyan}]{tirumala-etal-2022-memorization}
Tirumala, Kushal, Aram~H. Markosyan, Luke Zettlemoyer, and Armen Aghajanyan.
  2022.
\newblock Memorization without overfitting: Analyzing the training dynamics of
  large language models.
\newblock In \emph{Advances in Neural Information Processing Systems},
  volume~35, pages 38274--38290.

\bibitem[{Touileb(2022)}]{touileb-2022-exploring}
Touileb, Samia. 2022.
\newblock Exploring the effects of negation and grammatical tense on bias
  probes.
\newblock In \emph{Proceedings of the 2nd Conference of the Asia-Pacific
  Chapter of the Association for Computational Linguistics and the 12th
  International Joint Conference on Natural Language Processing (Volume 2:
  Short Papers)}, pages 423--429.

\bibitem[{Traylor, Feiman, and Pavlick(2021)}]{traylor-etal-2021-and}
Traylor, Aaron, Roman Feiman, and Ellie Pavlick. 2021.
\newblock {AND} does not mean {OR}: Using formal languages to study language
  models{'} representations.
\newblock In \emph{Proceedings of the 59th Annual Meeting of the Association
  for Computational Linguistics and the 11th International Joint Conference on
  Natural Language Processing (Volume 2: Short Papers)}, pages 158--167.

\bibitem[{Trott(2023)}]{trott-2023-in}
Trott, Sean. 2023.
\newblock In cautious defense of {LLM}-ology.
\newblock \emph{Blog Post}.

\bibitem[{Trott et~al.(2023)Trott, Jones, Chang, Michaelov, and
  Bergen}]{trott-etal-2022-do}
Trott, Sean, Cameron~J. Jones, Tyler~A. Chang, James Michaelov, and Benjamin
  Bergen. 2023.
\newblock Do large language models know what humans know?
\newblock \emph{Cognitive Science}, 47(7):e13309.

\bibitem[{Truong et~al.(2022)Truong, Otmakhova, Baldwin, Cohn, Lau, and
  Verspoor}]{truong-etal-2022-not}
Truong, Thinh~Hung, Yulia Otmakhova, Timothy Baldwin, Trevor Cohn, Jey~Han Lau,
  and Karin Verspoor. 2022.
\newblock Not another negation benchmark: The {N}a{N}-{NLI} test suite for
  sub-clausal negation.
\newblock In \emph{Proceedings of the 2nd Conference of the Asia-Pacific
  Chapter of the Association for Computational Linguistics and the 12th
  International Joint Conference on Natural Language Processing (Volume 1: Long
  Papers)}, pages 883--894.

\bibitem[{Tuckute et~al.(2022)Tuckute, Sathe, Wang, Yoder, Shain, and
  Fedorenko}]{tuckute-etal-2022-sentspace}
Tuckute, Greta, Aalok Sathe, Mingye Wang, Harley Yoder, Cory Shain, and Evelina
  Fedorenko. 2022.
\newblock {S}ent{S}pace: Large-scale benchmarking and evaluation of text using
  cognitively motivated lexical, syntactic, and semantic features.
\newblock In \emph{Proceedings of the 2022 Conference of the North American
  Chapter of the Association for Computational Linguistics: Human Language
  Technologies: System Demonstrations}, pages 99--113.

\bibitem[{Upadhye, Bergen, and Kehler(2020)}]{upadhye-etal-2020-predicting}
Upadhye, Shiva, Leon Bergen, and Andrew Kehler. 2020.
\newblock Predicting reference: What do language models learn about discourse
  models?
\newblock In \emph{Proceedings of the 2020 Conference on Empirical Methods in
  Natural Language Processing (EMNLP)}, pages 977--982.

\bibitem[{Ushio et~al.(2021)Ushio, Espinosa~Anke, Schockaert, and
  Camacho-Collados}]{ushio-etal-2021-bert}
Ushio, Asahi, Luis Espinosa~Anke, Steven Schockaert, and Jose Camacho-Collados.
  2021.
\newblock {BERT} is to {NLP} what {A}lex{N}et is to {CV}: Can pre-trained
  language models identify analogies?
\newblock In \emph{Proceedings of the 59th Annual Meeting of the Association
  for Computational Linguistics and the 11th International Joint Conference on
  Natural Language Processing (Volume 1: Long Papers)}, pages 3609--3624.

\bibitem[{Valmeekam et~al.(2022)Valmeekam, Olmo, Sreedharan, and
  Kambhampati}]{valmeekam-etal-2022-large}
Valmeekam, Karthik, Alberto Olmo, Sarath Sreedharan, and Subbarao Kambhampati.
  2022.
\newblock Large language models still can't plan (a benchmark for {LLM}s on
  planning and reasoning about change).
\newblock In \emph{Foundation Models for Decision Making Workshop}.

\bibitem[{de~Vassimon~Manela et~al.(2021)de~Vassimon~Manela, Errington, Fisher,
  van Breugel, and Minervini}]{manela-etal-2021-stereotype}
de~Vassimon~Manela, Daniel, David Errington, Thomas Fisher, Boris van Breugel,
  and Pasquale Minervini. 2021.
\newblock Stereotype and skew: Quantifying gender bias in pre-trained and
  fine-tuned language models.
\newblock In \emph{Proceedings of the 16th Conference of the European Chapter
  of the Association for Computational Linguistics: Main Volume}, pages
  2232--2242.

\bibitem[{Vaswani et~al.(2017)Vaswani, Shazeer, Parmar, Uszkoreit, Jones,
  Gomez, Kaiser, and Polosukhin}]{vaswani-etal-2017-attention}
Vaswani, Ashish, Noam Shazeer, Niki Parmar, Jakob Uszkoreit, Llion Jones, Aidan
  Gomez, {\L}ukasz Kaiser, and Illia Polosukhin. 2017.
\newblock Attention is all you need.
\newblock In \emph{Advances in Neural Information Processing Systems},
  volume~30, pages 5998--6008.

\bibitem[{Vig and Belinkov(2019)}]{vig-belinkov-2019-analyzing}
Vig, Jesse and Yonatan Belinkov. 2019.
\newblock Analyzing the structure of attention in a transformer language model.
\newblock In \emph{Proceedings of the 2019 ACL Workshop BlackboxNLP: Analyzing
  and Interpreting Neural Networks for NLP}, pages 63--76.

\bibitem[{Vig et~al.(2020)Vig, Gehrmann, Belinkov, Qian, Nevo, Singer, and
  Shieber}]{vig-etal-2020-causal}
Vig, Jesse, Sebastian Gehrmann, Yonatan Belinkov, Sharon Qian, Daniel Nevo,
  Yaron Singer, and Stuart Shieber. 2020.
\newblock Investigating gender bias in language models using causal mediation
  analysis.
\newblock In \emph{Advances in Neural Information Processing Systems},
  volume~33, pages 12388--12401.

\bibitem[{Wahle et~al.(2022)Wahle, Ruas, Kirstein, and
  Gipp}]{wahle-etal-2022-how}
Wahle, Jan~Philip, Terry Ruas, Frederic Kirstein, and Bela Gipp. 2022.
\newblock How large language models are transforming machine-paraphrase
  plagiarism.
\newblock In \emph{Proceedings of the 2022 Conference on Empirical Methods in
  Natural Language Processing}, pages 952--963.

\bibitem[{Wallace et~al.(2019{\natexlab{a}})Wallace, Feng, Kandpal, Gardner,
  and Singh}]{wallace-etal-2019-universal}
Wallace, Eric, Shi Feng, Nikhil Kandpal, Matt Gardner, and Sameer Singh.
  2019{\natexlab{a}}.
\newblock Universal adversarial triggers for attacking and analyzing {NLP}.
\newblock In \emph{Proceedings of the 2019 Conference on Empirical Methods in
  Natural Language Processing and the 9th International Joint Conference on
  Natural Language Processing (EMNLP-IJCNLP)}, pages 2153--2162.

\bibitem[{Wallace et~al.(2019{\natexlab{b}})Wallace, Wang, Li, Singh, and
  Gardner}]{wallace-etal-2019-do}
Wallace, Eric, Yizhong Wang, Sujian Li, Sameer Singh, and Matt Gardner.
  2019{\natexlab{b}}.
\newblock Do {NLP} models know numbers? probing numeracy in embeddings.
\newblock In \emph{Proceedings of the 2019 Conference on Empirical Methods in
  Natural Language Processing and the 9th International Joint Conference on
  Natural Language Processing (EMNLP-IJCNLP)}, pages 5307--5315.

\bibitem[{Wang et~al.(2019)Wang, Pruksachatkun, Nangia, Singh, Michael, Hill,
  Levy, and Bowman}]{wang-etal-2019-superglue}
Wang, Alex, Yada Pruksachatkun, Nikita Nangia, Amanpreet Singh, Julian Michael,
  Felix Hill, Omer Levy, and Samuel Bowman. 2019.
\newblock {SuperGLUE}: A stickier benchmark for general-purpose language
  understanding systems.
\newblock In \emph{Advances in Neural Information Processing Systems},
  volume~32, pages 3266--3280.

\bibitem[{Wang et~al.(2021{\natexlab{a}})Wang, Shang, Lioma, Jiang, Yang, Liu,
  and Simonsen}]{wang-etal-2021-position}
Wang, Benyou, Lifeng Shang, Christina Lioma, Xin Jiang, Hao Yang, Qun Liu, and
  Jakob~Grue Simonsen. 2021{\natexlab{a}}.
\newblock On position embeddings in {BERT}.
\newblock In \emph{International Conference on Learning Representations}.

\bibitem[{Wang et~al.(2022{\natexlab{a}})Wang, Min, Deng, Shen, Wu,
  Zettlemoyer, and Sun}]{wang-etal-2022-towards}
Wang, Boshi, Sewon Min, Xiang Deng, Jiaming Shen, You Wu, Luke Zettlemoyer, and
  Huan Sun. 2022{\natexlab{a}}.
\newblock Towards understanding chain-of-thought prompting: An empirical study
  of what matters.
\newblock \emph{ArXiv}, arXiv:2212.10001.

\bibitem[{Wang et~al.(2021{\natexlab{b}})Wang, Zheng, Niu, and
  Zhang}]{wang-etal-2021-exploring}
Wang, Cunxiang, Boyuan Zheng, Yuchen Niu, and Yue Zhang. 2021{\natexlab{b}}.
\newblock Exploring generalization ability of pretrained language models on
  arithmetic and logical reasoning.
\newblock In \emph{Natural Language Processing and Chinese Computing}, pages
  758--769, Springer International Publishing.

\bibitem[{Wang et~al.(2022{\natexlab{b}})Wang, Sridhar, Yang, and
  Wang}]{wang-etal-2021-identifying}
Wang, Tianlu, Rohit Sridhar, Diyi Yang, and Xuezhi Wang. 2022{\natexlab{b}}.
\newblock Identifying and mitigating spurious correlations for improving
  robustness in {NLP} models.
\newblock In \emph{Findings of the Association for Computational Linguistics:
  NAACL 2022}, pages 1719--1729.

\bibitem[{Warstadt et~al.(2019)Warstadt, Cao, Grosu, Peng, Blix, Nie, Alsop,
  Bordia, Liu, Parrish, Wang, Phang, Mohananey, Htut, Jeretic, and
  Bowman}]{warstadt-etal-2019-investigating}
Warstadt, Alex, Yu~Cao, Ioana Grosu, Wei Peng, Hagen Blix, Yining Nie, Anna
  Alsop, Shikha Bordia, Haokun Liu, Alicia Parrish, Sheng-Fu Wang, Jason Phang,
  Anhad Mohananey, Phu~Mon Htut, Paloma Jeretic, and Samuel~R. Bowman. 2019.
\newblock Investigating {BERT}{'}s knowledge of language: Five analysis methods
  with {NPI}s.
\newblock In \emph{Proceedings of the 2019 Conference on Empirical Methods in
  Natural Language Processing and the 9th International Joint Conference on
  Natural Language Processing (EMNLP-IJCNLP)}, pages 2877--2887.

\bibitem[{Warstadt et~al.(2020)Warstadt, Parrish, Liu, Mohananey, Peng, Wang,
  and Bowman}]{warstadt-etal-2020-blimp}
Warstadt, Alex, Alicia Parrish, Haokun Liu, Anhad Mohananey, Wei Peng, Sheng-Fu
  Wang, and Samuel~R. Bowman. 2020.
\newblock {BL}i{MP}: The benchmark of linguistic minimal pairs for {E}nglish.
\newblock \emph{Transactions of the Association for Computational Linguistics},
  8:377--392.

\bibitem[{Webb, Holyoak, and Lu(2022)}]{webb-etal-2022-emergent}
Webb, Taylor, Keith Holyoak, and Hongjing Lu. 2022.
\newblock Emergent analogical reasoning in large language models.
\newblock \emph{ArXiv}, arXiv:2212.09196.

\bibitem[{Wei et~al.(2022{\natexlab{a}})Wei, Bosma, Zhao, Guu, Yu, Lester, Du,
  Dai, and Le}]{wei-etal-2021-finetuned}
Wei, Jason, Maarten Bosma, Vincent Zhao, Kelvin Guu, Adams~Wei Yu, Brian
  Lester, Nan Du, Andrew~M. Dai, and Quoc~V. Le. 2022{\natexlab{a}}.
\newblock Finetuned language models are zero-shot learners.
\newblock In \emph{International Conference on Learning Representations}.

\bibitem[{Wei et~al.(2021)Wei, Garrette, Linzen, and
  Pavlick}]{wei-etal-2021-frequency}
Wei, Jason, Dan Garrette, Tal Linzen, and Ellie Pavlick. 2021.
\newblock Frequency effects on syntactic rule learning in transformers.
\newblock In \emph{Proceedings of the 2021 Conference on Empirical Methods in
  Natural Language Processing}, pages 932--948.

\bibitem[{Wei et~al.(2023)Wei, Kim, Tay, and Le}]{wei-etal-2022-inverse}
Wei, Jason, Najoung Kim, Yi~Tay, and Quoc~V. Le. 2023.
\newblock Inverse scaling can become {U}-shaped.
\newblock \emph{ArXiv}, arXiv:2211.02011.

\bibitem[{Wei et~al.(2022{\natexlab{b}})Wei, Tay, Bommasani, Raffel, Zoph,
  Borgeaud, Yogatama, Bosma, Zhou, Metzler, Chi, Hashimoto, Vinyals, Liang,
  Dean, and Fedus}]{wei-etal-2022-emergent}
Wei, Jason, Yi~Tay, Rishi Bommasani, Colin Raffel, Barret Zoph, Sebastian
  Borgeaud, Dani Yogatama, Maarten Bosma, Denny Zhou, Donald Metzler, Ed~H.
  Chi, Tatsunori Hashimoto, Oriol Vinyals, Percy Liang, Jeff Dean, and William
  Fedus. 2022{\natexlab{b}}.
\newblock Emergent abilities of large language models.
\newblock \emph{Transactions on Machine Learning Research}.

\bibitem[{Wei et~al.(2022{\natexlab{c}})Wei, Wang, Schuurmans, Bosma, Ichter,
  Xia, Chi, Le, and Zhou}]{wei-etal-2022-chain}
Wei, Jason, Xuezhi Wang, Dale Schuurmans, Maarten Bosma, Brian Ichter, Fei Xia,
  Ed~Chi, Quoc Le, and Denny Zhou. 2022{\natexlab{c}}.
\newblock Chain of thought prompting elicits reasoning in large language
  models.
\newblock In \emph{Advances in Neural Information Processing Systems},
  volume~35, pages 24824--24837.

\bibitem[{Weidinger et~al.(2021)Weidinger, Mellor, Rauh, Griffin, Uesato,
  Huang, Cheng, Glaese, Balle, Kasirzadeh, Kenton, Brown, Hawkins, Stepleton,
  Biles, Birhane, Haas, Rimell, Hendricks, Isaac, Legassick, Irving, and
  Gabriel}]{weidinger-etal-2021-ethical}
Weidinger, Laura, John F.~J. Mellor, Maribeth Rauh, Conor Griffin, Jonathan
  Uesato, Po-Sen Huang, Myra Cheng, Mia Glaese, Borja Balle, Atoosa Kasirzadeh,
  Zachary Kenton, Sande~Minnich Brown, William~T. Hawkins, Tom Stepleton,
  Courtney Biles, Abeba Birhane, Julia Haas, Laura Rimell, Lisa~Anne Hendricks,
  William~S. Isaac, Sean Legassick, Geoffrey Irving, and Iason Gabriel. 2021.
\newblock Ethical and social risks of harm from language models.
\newblock \emph{ArXiv}, arXiv:2112.04359.

\bibitem[{Weidinger et~al.(2022)Weidinger, Uesato, Rauh, Griffin, Huang,
  Mellor, Glaese, Cheng, Balle, Kasirzadeh, Biles, Brown, Kenton, Hawkins,
  Stepleton, Birhane, Hendricks, Rimell, Isaac, Haas, Legassick, Irving, and
  Gabriel}]{weidinger-etal-2022-taxonomy}
Weidinger, Laura, Jonathan Uesato, Maribeth Rauh, Conor Griffin, Po-Sen Huang,
  John Mellor, Amelia Glaese, Myra Cheng, Borja Balle, Atoosa Kasirzadeh,
  Courtney Biles, Sasha Brown, Zac Kenton, Will Hawkins, Tom Stepleton, Abeba
  Birhane, Lisa~Anne Hendricks, Laura Rimell, William Isaac, Julia Haas, Sean
  Legassick, Geoffrey Irving, and Iason Gabriel. 2022.
\newblock Taxonomy of risks posed by language models.
\newblock In \emph{Proceedings of the ACM Conference on Fairness,
  Accountability, and Transparency}, page 214–229, Association for Computing
  Machinery, New York, NY, USA.

\bibitem[{Weir, Poliak, and Durme(2020)}]{weir-etal-2020-probing}
Weir, Nathaniel, Adam Poliak, and Benjamin~Van Durme. 2020.
\newblock Probing neural language models for human tacit assumptions.
\newblock In \emph{Annual Meeting of the Cognitive Science Society}, volume~42,
  pages 377--383.

\bibitem[{Weissweiler et~al.(2022)Weissweiler, Hofmann, K{\"o}ksal, and
  Sch{\"u}tze}]{weissweiler-etal-2022-the}
Weissweiler, Leonie, Valentin Hofmann, Abdullatif K{\"o}ksal, and Hinrich
  Sch{\"u}tze. 2022.
\newblock The better your syntax, the better your semantics? probing pretrained
  language models for the {E}nglish comparative correlative.
\newblock In \emph{Proceedings of the 2022 Conference on Empirical Methods in
  Natural Language Processing}, pages 10859--10882.

\bibitem[{Welbl et~al.(2021)Welbl, Glaese, Uesato, Dathathri, Mellor,
  Hendricks, Anderson, Kohli, Coppin, and
  Huang}]{welbl-etal-2021-challenges-detoxifying}
Welbl, Johannes, Amelia Glaese, Jonathan Uesato, Sumanth Dathathri, John
  Mellor, Lisa~Anne Hendricks, Kirsty Anderson, Pushmeet Kohli, Ben Coppin, and
  Po-Sen Huang. 2021.
\newblock Challenges in detoxifying language models.
\newblock In \emph{Findings of the Association for Computational Linguistics:
  EMNLP 2021}, pages 2447--2469.

\bibitem[{Wettig et~al.(2023)Wettig, Gao, Zhong, and
  Chen}]{wettig-etal-2022-should}
Wettig, Alexander, Tianyu Gao, Zexuan Zhong, and Danqi Chen. 2023.
\newblock Should you mask 15\% in masked language modeling?
\newblock In \emph{Proceedings of the 17th Conference of the European Chapter
  of the Association for Computational Linguistics}, pages 2985--3000.

\bibitem[{White and Cotterell(2021)}]{white-cotterell-2021-examining}
White, Jennifer~C. and Ryan Cotterell. 2021.
\newblock Examining the inductive bias of neural language models with
  artificial languages.
\newblock In \emph{Proceedings of the 59th Annual Meeting of the Association
  for Computational Linguistics and the 11th International Joint Conference on
  Natural Language Processing (Volume 1: Long Papers)}, pages 454--463.

\bibitem[{Wilcox, Futrell, and Levy(2022)}]{wilcox-etal-2022-using}
Wilcox, Ethan~Gotlieb, Richard Futrell, and Roger Levy. 2022.
\newblock {Using Computational Models to Test Syntactic Learnability}.
\newblock \emph{Linguistic Inquiry}, pages 1--88.

\bibitem[{Williams, Nangia, and Bowman(2018)}]{williams-etal-2018-a-broad}
Williams, Adina, Nikita Nangia, and Samuel Bowman. 2018.
\newblock A broad-coverage challenge corpus for sentence understanding through
  inference.
\newblock In \emph{Proceedings of the 2018 Conference of the North American
  Chapter of the Association for Computational Linguistics: Human Language
  Technologies, Volume 1 (Long Papers)}, pages 1112--1122.

\bibitem[{Wu and Dredze(2020)}]{wu-dredze-2020-languages}
Wu, Shijie and Mark Dredze. 2020.
\newblock Are all languages created equal in multilingual {BERT}?
\newblock In \emph{Proceedings of the 5th Workshop on Representation Learning
  for NLP}, pages 120--130.

\bibitem[{Xia et~al.(2022)Xia, Artetxe, Zhou, Lin, Pasunuru, Chen, Zettlemoyer,
  and Stoyanov}]{xia-etal-2022-training}
Xia, Mengzhou, Mikel Artetxe, Chunting Zhou, Xi~Victoria Lin, Ramakanth
  Pasunuru, Danqi Chen, Luke Zettlemoyer, and Ves Stoyanov. 2022.
\newblock Training trajectories of language models across scales.
\newblock \emph{ArXiv}, arXiv:2212.09803.

\bibitem[{Yu et~al.(2020)Yu, Sie, Tedeschi, and Bergen}]{yu-etal-2020-word}
Yu, Charles, Ryan Sie, Nicolas Tedeschi, and Leon Bergen. 2020.
\newblock Word frequency does not predict grammatical knowledge in language
  models.
\newblock In \emph{Proceedings of the 2020 Conference on Empirical Methods in
  Natural Language Processing (EMNLP)}, pages 4040--4054.

\bibitem[{Zafrir et~al.(2019)Zafrir, Boudoukh, Izsak, and
  Wasserblat}]{zafrir-etal-2019-q8bert}
Zafrir, Ofir, Guy Boudoukh, Peter Izsak, and Moshe Wasserblat. 2019.
\newblock {Q8BERT}: Quantized 8bit {BERT}.
\newblock In \emph{Fifth Workshop on Energy Efficient Machine Learning and
  Cognitive Computing}, pages 36--39, IEEE.

\bibitem[{Zellers et~al.(2021)Zellers, Holtzman, Clark, Qin, Farhadi, and
  Choi}]{zellers-etal-2020-evaluating}
Zellers, Rowan, Ari Holtzman, Elizabeth Clark, Lianhui Qin, Ali Farhadi, and
  Yejin Choi. 2021.
\newblock {T}uring{A}dvice: A generative and dynamic evaluation of language
  use.
\newblock In \emph{Proceedings of the 2021 Conference of the North American
  Chapter of the Association for Computational Linguistics: Human Language
  Technologies}, pages 4856--4880.

\bibitem[{Zhang et~al.(2023{\natexlab{a}})Zhang, Song, Li, Zhou, and
  Song}]{zhang-etal-2023-survey}
Zhang, Hanqing, Haolin Song, Shaoyu Li, Ming Zhou, and Dawei Song.
  2023{\natexlab{a}}.
\newblock A survey of controllable text generation using {T}ransformer-based
  pre-trained language models.
\newblock \emph{ArXiv}, arXiv:2201.05337.

\bibitem[{Zhang et~al.(2023{\natexlab{b}})Zhang, Xu, Yang, Zhou, You, Arora,
  and Callison-Burch}]{zhang-etal-2023-causal}
Zhang, Li, Hai Xu, Yue Yang, Shuyan Zhou, Weiqiu You, Manni Arora, and Chris
  Callison-Burch. 2023{\natexlab{b}}.
\newblock Causal reasoning of entities and events in procedural texts.
\newblock In \emph{Findings of the Association for Computational Linguistics:
  EACL 2023}, pages 415--431.

\bibitem[{Zhang et~al.(2022{\natexlab{a}})Zhang, Wang, Chen, and
  Zhang}]{zhang-etal-2022-probing-gpt3s}
Zhang, Lining, Mengchen Wang, Liben Chen, and Wenxin Zhang. 2022{\natexlab{a}}.
\newblock Probing {GPT}-3{'}s linguistic knowledge on semantic tasks.
\newblock In \emph{Proceedings of the Fifth BlackboxNLP Workshop on Analyzing
  and Interpreting Neural Networks for NLP}, pages 297--304.

\bibitem[{Zhang et~al.(2021{\natexlab{a}})Zhang, Zhang, Zhang, and
  S{\o}gaard}]{zhang-etal-2021-sociolectal}
Zhang, Sheng, Xin Zhang, Weiming Zhang, and Anders S{\o}gaard.
  2021{\natexlab{a}}.
\newblock Sociolectal analysis of pretrained language models.
\newblock In \emph{Proceedings of the 2021 Conference on Empirical Methods in
  Natural Language Processing}, pages 4581--4588.

\bibitem[{Zhang et~al.(2022{\natexlab{b}})Zhang, Roller, Goyal, Artetxe, Chen,
  Chen, Dewan, Diab, Li, Lin, Mihaylov, Ott, Shleifer, Shuster, Simig, Koura,
  Sridhar, Wang, and Zettlemoyer}]{zhang-etal-2022-opt}
Zhang, Susan, Stephen Roller, Naman Goyal, Mikel Artetxe, Moya Chen, Shuohui
  Chen, Christopher Dewan, Mona Diab, Xian Li, Xi~Victoria Lin, Todor Mihaylov,
  Myle Ott, Sam Shleifer, Kurt Shuster, Daniel Simig, Punit~Singh Koura, Anjali
  Sridhar, Tianlu Wang, and Luke Zettlemoyer. 2022{\natexlab{b}}.
\newblock {OPT}: Open pre-trained {Transformer} language models.
\newblock \emph{ArXiv}, arXiv:2205.01068.

\bibitem[{Zhang et~al.(2020)Zhang, Kishore, Wu, Weinberger, and
  Artzi}]{zhang-etal-2019-bertscore}
Zhang, Tianyi, Varsha Kishore, Felix Wu, Kilian~Q. Weinberger, and Yoav Artzi.
  2020.
\newblock {BERTScore}: Evaluating text generation with {BERT}.
\newblock In \emph{International Conference on Learning Representations}.

\bibitem[{Zhang et~al.(2021{\natexlab{b}})Zhang, Warstadt, Li, and
  Bowman}]{zhang-etal-2020-when-do}
Zhang, Yian, Alex Warstadt, Xiaocheng Li, and Samuel~R. Bowman.
  2021{\natexlab{b}}.
\newblock When do you need billions of words of pretraining data?
\newblock In \emph{Proceedings of the 59th Annual Meeting of the Association
  for Computational Linguistics and the 11th International Joint Conference on
  Natural Language Processing (Volume 1: Long Papers)}, pages 1112--1125.

\bibitem[{Zhang et~al.(2019)Zhang, Han, Liu, Jiang, Sun, and
  Liu}]{zhang-etal-2019-ernie}
Zhang, Zhengyan, Xu~Han, Zhiyuan Liu, Xin Jiang, Maosong Sun, and Qun Liu.
  2019.
\newblock {ERNIE}: Enhanced language representation with informative entities.
\newblock In \emph{Proceedings of the 57th Annual Meeting of the Association
  for Computational Linguistics}, pages 1441--1451.

\bibitem[{Zhao et~al.(2021)Zhao, Ngui, Hall~Hartley, and
  Bethard}]{zhao-etal-2021-do}
Zhao, Yiyun, Jian~Gang Ngui, Lucy Hall~Hartley, and Steven Bethard. 2021.
\newblock Do pretrained transformers infer telicity like humans?
\newblock In \emph{Proceedings of the 25th Conference on Computational Natural
  Language Learning}, pages 72--81.

\bibitem[{Zhao, Zhang, and Hopfgartner(2021)}]{zhao-etal-2021-a-comparative}
Zhao, Zhixue, Ziqi Zhang, and Frank Hopfgartner. 2021.
\newblock A comparative study of using pre-trained language models for toxic
  comment classification.
\newblock In \emph{The ACM Web Conference}, page 500–507, Association for
  Computing Machinery, New York, NY, USA.

\bibitem[{Zhou, Ethayarajh, and Jurafsky(2022)}]{zhou-etal-2022-richer}
Zhou, Kaitlyn, Kawin Ethayarajh, and Dan Jurafsky. 2022.
\newblock Richer countries and richer representations.
\newblock In \emph{Findings of the Association for Computational Linguistics:
  ACL 2022}, pages 2074--2085.

\bibitem[{Zhou et~al.(2021)Zhou, Khanna, Lee, Lin, Ho, Pujara, and
  Ren}]{zhou-etal-2020-rica}
Zhou, Pei, Rahul Khanna, Seyeon Lee, Bill~Yuchen Lin, Daniel Ho, Jay Pujara,
  and Xiang Ren. 2021.
\newblock {RICA}: Evaluating robust inference capabilities based on commonsense
  axioms.
\newblock In \emph{Proceedings of the 2021 Conference on Empirical Methods in
  Natural Language Processing}, pages 7560--7579.

\bibitem[{Zong and Krishnamachari(2022)}]{zong-krishnamachari-2022-a}
Zong, Mingyu and Bhaskar Krishnamachari. 2022.
\newblock A survey on {GPT-3}.
\newblock \emph{ArXiv}, arXiv:2212.00857.

\bibitem[{Zwaan(2016)}]{zwaan-2016-lexical}
Zwaan, Rolf~A. 2016.
\newblock Situation models, mental simulations, and abstract concepts in
  discourse comprehension.
\newblock \emph{Psychonomic Bulletin \& Review}, 23:1028--1034.

\end{thebibliography}

\appendix
\appendixsection{Literature Review Process}
\label{app:lit-review}
We identified papers to include in this survey using Semantic Scholar \citep{fricke-2018-semantic}.
From a seed of 271 relevant language model analysis papers (including the majority of the citation list from \citealp{rogers-etal-2020-a}), we extracted all papers that cited any paper in the seed.
This resulted in over 15K papers, last scraped on February 4, 2023.
Anecdotally, the majority of recent language model analysis papers we encountered were included in this list.
We manually filtered by title down to approximately 1500 potentially relevant papers, gradually refining the scope as described in Section\autoref{sec:scope}.
We then further filtered by abstract down to approximately 400 highly relevant papers.

\end{document}